\documentclass[runningheads]{llncs}

\usepackage{graphicx}
\usepackage{breakcites}  
\usepackage{amsmath}
\usepackage{graphicx}

\usepackage{svg}
\usepackage{subcaption}
\usepackage{booktabs}
\usepackage{amsfonts}
\usepackage{multirow}
\usepackage{pifont}

\newcommand{\Nqueries}{N}
\newcommand{\Lone}[1]{\left\lVert #1 \right\rVert_1}

\newcommand{\bbox}[1]{b_{#1}}
\newcommand{\bboxhat}[1]{\hat{b}_{#1}}

\newcommand{\class}[1]{c_{#1}}

\newcommand{\classparam}[2]{g_{#1}[{#2}]}
\newcommand{\classparamset}[1]{g_{#1}}
\newcommand{\score}[2]{p_{#1}[#2]}

\newcommand{\params}[1]{g_{#1}}

\begin{document}

\title{%
 Historical Astronomical Diagrams\\ Decomposition in Geometric Primitives
 }%

\author{Syrine Kalleli\inst{1} \and Scott Trigg\inst{2} \and Ségolène Albouy\inst{2} \and  Mathieu Husson\inst{2} \and
Mathieu Aubry\inst{1} }

\authorrunning{S. Kalleli et al.}

\institute{LIGM, Ecole des Ponts, Univ Gustave Eiffel, CNRS, Marne-la-Vallée, France \and SYRTE, Observatoire de Paris-PSL, CNRS, Paris, France \\
\email{syrine.kalleli@enpc.fr}\\ }
\maketitle              %
\begin{abstract}

Automatically extracting the geometric content from the hundreds of thousands of diagrams drawn in historical manuscripts would enable historians to study the diffusion of astronomical knowledge on a global scale. 
However, state-of-the-art vectorization methods, often designed to tackle modern data, are not adapted to the complexity and diversity of historical astronomical diagrams. Our contribution is thus twofold. First, we introduce a unique dataset of 303 astronomical diagrams from diverse traditions, ranging from the XIIth to the XVIIIth century, annotated with more than 3000 line segments, circles and arcs. Second, we develop a model that builds on DINO-DETR to enable the prediction of multiple geometric primitives. We show that it can be trained solely on synthetic data and accurately predict primitives on our challenging dataset. Our approach widely improves over the LETR baseline, which is restricted to lines, by introducing a meaningful parametrization for multiple primitives, jointly training for detection and parameter refinement, using deformable attention and training on rich synthetic data. Our dataset and code are available at our project webpage: \texttt{\small http://imagine.enpc.fr/\texttildelow kallelis/icdar2024/}.

\end{abstract}

\begin{figure}
    \centering
    \includegraphics[width=\textwidth]{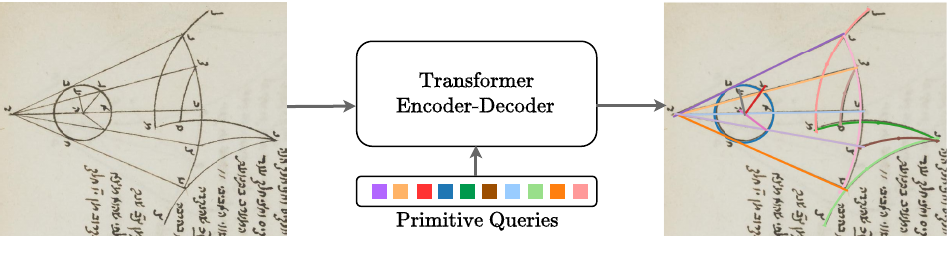}
    \caption{{\bf Goal. } We perform historical astronomical diagram vectorization by predicting simple geometric primitives, such as lines, circles, and arcs, through a transformer encoder-decoder model. Our modified decoder queries, which we refer to as a primitive queries, are associated to different geometric primitives. 
    }
    \label{fig:teaser}
\end{figure}

\section{Introduction}

Astronomical diagrams offer a unique perspective to understand the diffusion of astronomical knowledge over centuries of Afro-Eurasian history. %
Thousands of digitized manuscripts with hundreds of thousands of diagrams are available. 
We aim to provide automatic tools to analyze their content. More precisely, we want to vectorize historical diagram images into a set of parameterized geometric primitives, namely line segments, circles, and arcs, as shown in Figure~\ref{fig:teaser}. 
Such a representation of diagrams would enable effective semantic similarity search and would serve as an important tool for historians to create critical editions of astronomical treatises.

Inferring vector primitives from raster drawings is a long-standing problem in computer vision. Early works address this raster-to-vector task for architectural plans~\cite{Dosch2000ACS} 
and for clean technical drawings~\cite{hilaire2006robust} through a multi-step process for line and arc detection. Several optimization approaches~\cite{bessmeltsev2019vectorization, favreau2016fidelity, BO201614} have been developed and deep learning methods~\cite{liu2017raster, egiazarian2020deep, Mo2021GeneralVS} have shown impressive results on rougher drawings. 
Recently, models based on the DEtection TRansformer (DETR)~\cite{carion2020end} have demonstrated superior performances for line detection~\cite{xu2021line}, floor plan reconstruction~\cite{yue2023connecting}, dashed curve recognition~\cite{liu2022neural}, image vectorization~\cite{chen2023editable}, and primitive parameter inference~\cite{wang2023ppi}. %
However, these methods are constrained to clean input images typically rendered from vector format such as technical diagrams~\cite{seff2021vitruvion, wang2023ppi} or icons and emojis~\cite{reddy2021im2vec, chen2023editable}.

In this work, we %
introduce a challenging dataset of 303 historical astronomical diagrams extracted from manuscripts and Chinese woodblock prints of the Middle Ages and modern period annotated by historian experts. Even state-of-the-art deep models such as LETR~\cite{xu2021line} struggle with such data and most approaches only detect a single kind of primitive.
We thus present a transformer-based vectorization method which, opposite to  %
 prior work~\cite{wang2023ppi, liu2022neural, seff2021vitruvion}, (i) jointly predicts line segments, circles, and arcs and (ii) adopts deformable attention~\cite{zhu2020deformable} in an iterative primitive-refinement scheme, that significantly speeds up training convergence and enables fine-grained primitives localization. Note that our parameter refinement scheme can be seamlessly extended  to support other geometric primitives which could  be beneficial for other vectorization applications.  %
Our model is trained solely on rich synthetic data and we demonstrate that it generalizes to our challenging dataset. Our dataset and code are available at our project webpage: 
\texttt{\small http://imagine.enpc.fr/\texttildelow kallelis/icdar2024/}.

In summary, our two main contributions are:
\begin{itemize}
    \item A diverse dataset of 303 astronomical diagrams
     from Arabic, Latin, Greek, Hebrew, and Chinese manuscripts, ranging from the XIIth to the XVIIIth century, annotated by historians with more than 3000 line segments, circles, and arcs.
    \item A 
     transformer-based model for diagram vectorization that iteratively refines query primitive parameters using deformable attention and can jointly detect lines, circles, and arcs. 
\end{itemize}

\section{Related work}

In this section, we start by reviewing general works on drawing vectorization. As the basis of our approach, we then discuss transformer-based detection, which is the foundation for many recent vectorization techniques~\cite{chen2023editable, wang2023ppi, liu2022neural, liu2022end}, including our method. Finally, we briefly present works related to historical diagrams.

\subsubsection{Drawing vectorization.} 
While our primary focus is the vectorization of line-drawings and sketches, we occasionally refer to works for image vectorization that are particularly relevant~\cite{chen2023editable, xu2021line, reddy2021im2vec}. 

Approaches to vectorize line-drawings can be roughly categorized in two categories: optimization-based and learning-based.
\paragraph{Optimization-based approaches.}
Initial efforts in diagram vectorization~\cite{hilaire2006robust} target technical diagrams composed of straight lines and circular arcs. The approach of~\cite{hilaire2006robust} extracts a 1-pixel wide skeleton and junctions from a clean black-and-white image, fits primitives through local optimization, and refines the output by merging predictions using heuristics. The methods in~\cite{favreau2016fidelity, donati2017accurate} extend this pipeline to extract Bézier curves from input sketches and introduce a global optimization scheme. Many follow-up works~\cite{bessmeltsev2019vectorization, stanko2020integer, Puhachov2021KeypointdrivenLD, Gut2023SingularityFreeFF} shift to frame fields to favor robustness, assigning two directions to each pixel in the image, to guide curve extraction. A fundamental issue of these multi-step processes is that they are prone to error accumulation, which is especially likely with rough sketches. They also often require hyperparameter adjustments, and since optimization happens for each image, they can be time-consuming. Thus, most recent approaches for vectorization rely instead on machine learning, and in particular deep learning.

\paragraph{Learning-based approaches.} 

 A first set of methods use deep learning to replace parts of a multi-step pipeline. For example,~\cite{liu2017raster} outputs vectorized, semantically annotated floor plans from raster images by using a CNN for junction detection and generating primitive proposals which are then post-processed into vector objects. Similarly,~\cite{kim2018semantic} learns a pixel-level segmentation of line-art strokes and then uses Potrace~\cite{Selinger2003PotraceA}, a third-party tracing tool, and~\cite{Mo2021GeneralVS} learns junctions, then reconstructs the topology and vectorizes using least-squares fitting. 
 
 A second  set of methods directly predicts primitive parameters from the input image. Some works predict Bézier or spline-based curves using either RNNs\cite{Gao2019DeepSplineDR, reddy2021im2vec} or Transformers~\cite{liu2022end, liu2022neural, egiazarian2020deep}, which enables modeling complex shapes. 
More closely related to our goal are works that aim at predicting a fixed set of simple geometric primitives. Several methods~\cite{seff2021vitruvion, wang2023ppi, para2021sketchgen, ganin2021computer} target CAD designs~\cite{seff2020sketchgraphs}, and PMN~\cite{alaniz2022abstracting} predicts sketch strokes. LETR~\cite{xu2021line} focuses on line segment detection in natural scenes but was particularly influential for recent vectorization works. It introduces a transformer encoder-decoder model built upon the DEtection TRansformer~\cite{carion2020end} (DETR), which we discuss in the next paragraph. %
The method in~\cite{chen2023editable} adopts DETR to reconstruct images from basic geometric primitives such as rectangles and circles in a self-supervised manner.
Very recently, PPI-Net~\cite{wang2023ppi} proposed a DETR-like model with denoising groups~\cite{li2022dn} for vectorizing simple CAD designs into lines, circles, arcs, and points. However, these models based on DETR are very slow to train and are typically constrained to specific primitive types \cite{xu2021line, wang2023ppi} or very simplistic image domains \cite{chen2023editable, wang2023ppi, seff2021vitruvion, para2021sketchgen, ganin2021computer}. In contrast, we adopt several improvements to the DETR model which speed up training. Aditionally, we propose a method that jointly handles lines, circles, and arcs, but can be easily extended to support more primitive types.

\subsubsection{Transformer-based detectors}
DETR~\cite{carion2020end} casts object detection as an image-to-set problem and employs a transformer encoder-decoder architecture to produce a set of box predictions. The DETR framework is general and has proven to adapt to several tasks but it suffers from slow convergence and poor performance on small objects.
Deformable-DETR~\cite{zhu2020deformable} uses multi-scale features and introduces deformable attention to significantly speed up convergence. Most follow-up works adopt deformable attention and further speed up training by incorporating spatial priors in the queries~\cite{liu2022dab, meng2021conditional, wang2021anchor} or stabilizing Hungarian matching~\cite{li2022dn, zhang2022dino}. The most relevant to our work are (i) Dab-DETR~\cite{liu2022dab}, which dynamically refines anchor boxes instead of object queries, further improving training speed; (ii) DN-DETR~\cite{li2022dn}, which adds a denoising branch to the decoder; (iii) DINO-DETR~\cite{zhang2022dino}, which achieves state-of-the-art results through several innovations, including contrastive denoising. Our method can be seen as formulating primitive detection in such a way that it can be performed with an approach in the spirit of DINO-DETR.

\subsubsection{Historical diagrams.} 
Historical document datasets typically favor textual, tabular, and image content, often overlooking diagrams. A recent exception is the S-VED dataset~\cite{buttner2022cor}, a large heterogeneous collection of illustrations featuring many {printed} astronomical diagrams. Unfortunately, this dataset is not labeled with geometric primitives. There are however works addressing other challenges related to deteriorated historical drawings, including sketches and architectural plans. 
Many works focus on restoration, for example the method in~\cite{sasaki2018learning} focuses on line drawings, while a VQ VAE proposed in~\cite{choi2023restoration} restores aged hand-drawn architectural plans. Closer to our task,~\cite{swaileh2021versailles} presents a CNN-based wall detection method on ancient floor plans and~\cite{chen2021vectorization} designed a three-step optimization process to convert a raster image of a historical map into a set of polygons.

\section{Approach}

\begin{figure}[t]
    \centering
    \includegraphics[width=\textwidth]{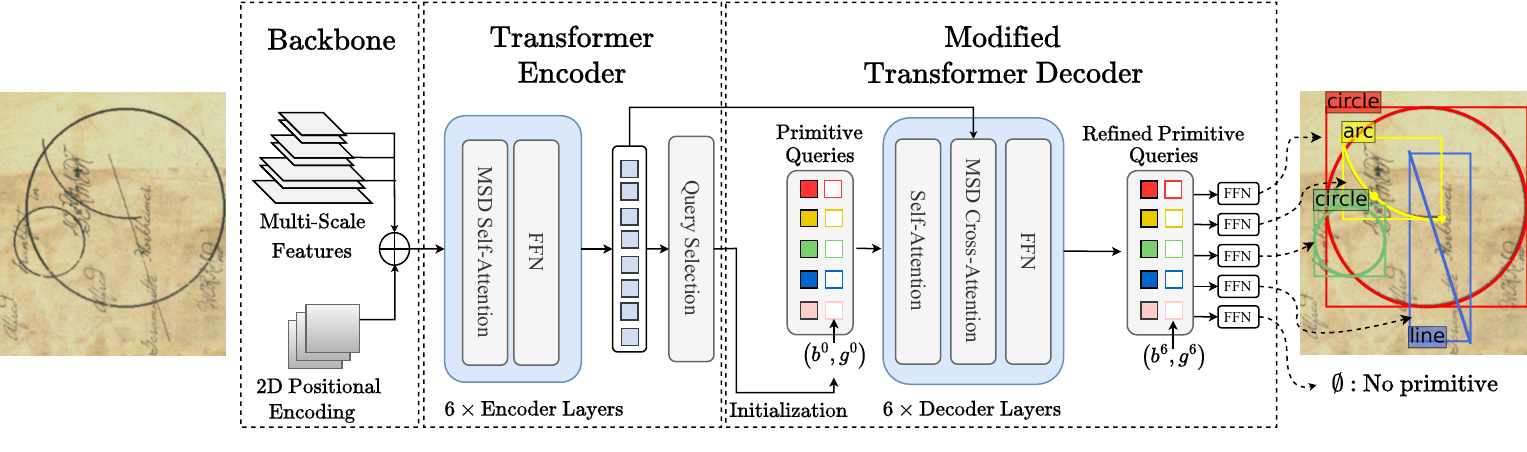}
    \caption{ {\bf Model architecture.} Given an input image, the backbone extracts multi-scale features which are fed to the Transformer encoder along with a positional encoding. The primitive queries, composed of content (filled) and modified positional (empty) queries, go through the Transformer decoder where they probe the enhanced encoder features through deformable cross-attention. Queries are refined layer-by-layer in the decoder, to finally predict the primitive class, bounding box and parameters. %
    }
    \label{fig:pipeline}
\end{figure}

\subsection{Model overview and notations}
Given an input image of a diagram, our goal is to output the set of primitives shapes present in the input together with their parameters. We focus on three primitive classes, Line, Circle, and Arc, but our model could be extended to more classes. We write $\mathcal{C}$ the set of primitive classes, including a specific ``no primitive" class we write $\emptyset$. 
Each primitive class $c \in \mathcal{C}$ is associated to a vector of geometric parameters $\classparam{}{c}$, where the length of the vector $m[c]$ can be different for each class and is by convention $0$ for the ``no primitive" class. %

\subsubsection{Architecture} To address our image-to-set problem, we use the transformer encoder-decoder architecture outlined in Figure~\ref{fig:pipeline}. This is similar to LETR~\cite{xu2021line}, but instead of using a two-stage coarse-to-fine strategy to perform fine-grained detection, we demonstrate how to leverage Multi-Scale Deformable (MSD) attention~\cite{zhu2020deformable} for geometric primitive prediction. In general, we follow the standard end-to-end DETR-like architecture but we adapt several improvements from DETR-variants, notably DINO-DETR~\cite{zhang2022dino} to boost speed and performance. 

A CNN backbone first extracts multi-scale features from the input image which are subsequently concatenated with a positional embedding. A standard transformer encoder then processes these features using 6 encoder blocks based on MSD self-attention. It outputs both an initial set of $\Nqueries$ modified positional queries that will be used as initialization by the decoder and enhanced multi-scale image features which are then used by the transformer decoder as keys and values in cross attention. Our modified transformer decoder concatenates learned {content} queries to the positional queries provided by the encoder (mixed query selection~\cite{zhang2022dino}) and processes them successively in 6 modified decoder blocks using MSD self-attention and MSD cross-attention with the encoder image features. These modified decoder blocks can be seen as iteratively refining the predictions, until the final predictions are made by a last feed-forward network, and are detailed in Section~\ref{sec:refinement}.

\subsubsection{Primitive queries} We index the $\Nqueries$ modified decoder queries, which we refer to as primitive queries, by $n\in {1,...,\Nqueries}$ and for each query, we will predict:
\begin{itemize}
    \item probabilities $p_n$ of the prediction to be associated to each primitive class, including the ``no primitive" class $\emptyset$.
    \item a bounding box $\bbox{n} \in [0,1]^4$ relative to the image size.
    \item parameters $\classparam{n}{c} \in [0,1]^{m[c]}$ relative to the image size for all classes $c$, which are concatenated into $\classparamset{n} \in [0,1]^{d}$, with $d = \sum_{c \in \mathcal{C}} m[c]$. 
\end{itemize}
Note that the bounding box could be predicted from the parameters $\classparam{n}{c_n}$, but having a class-independent bounding box is necessary to correctly define MSD attention. This novel positional query parametrization is critical to our primitive refinement detailed in Section~\ref{sec:refinement} and our losses discussed in Section~\ref{sec:losses}.

\label{sec:architecture}

\begin{figure}[t]
    \centering
    \includegraphics[width=0.8\textwidth]{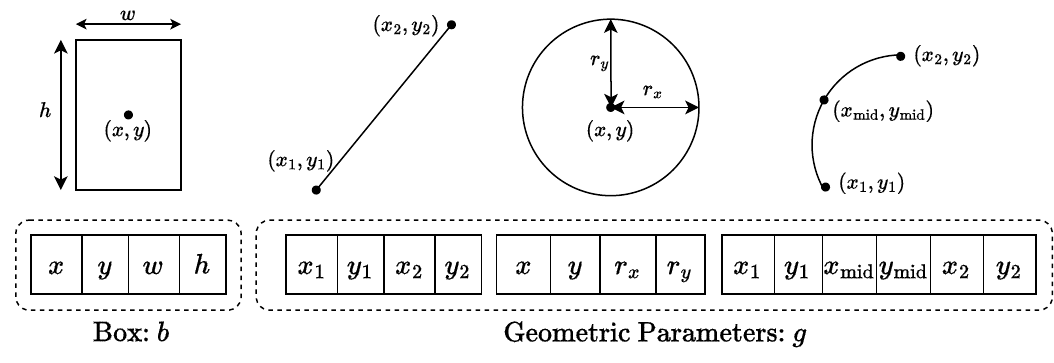}
        \caption{{\bf Modified Positional Queries.} All coordinates are normalized with respect to the image size. The bounding box is defined by its center and size. Lines are defined by endpoints, circles by center and radius (normalized by image width, $r_x$, and by image height, $r_y$), and arcs by endpoints and midpoint.
        }
    \label{fig:parameterization}
\end{figure}

\subsection{Primitive Refinement} 
\label{sec:refinement}

\begin{figure}[h!]
    \begin{minipage}[t]{0.4\textwidth}
    \vspace{-68mm} %
    \normalsize
        \subsubsection{Modified transformer decoder block}
A decoder block along with an example of primitive refinement can be seen in Figure~\ref{fig:decoder_layer}. 
At each decoder layer $l$, the input is a set of primitive queries composed of a content query %
and a modified positional query $(\bbox{}^{l}, \params{}^l)$. At the beginning of the block, they are combined by summing the content query with an encoded positional query, obtained by first concatenating the 2D sinusoidal embedding of all the parameters of the positional query then processing this concatenated vector with a Multi-Layer Perceptron (MLP). Note that this can be done for any number of primitive classes and any length of parametrization for the different classes.
    \end{minipage}
    \hfill
    \begin{minipage}[t]{0.58\textwidth}
        \centering
        \includegraphics[width=\textwidth]{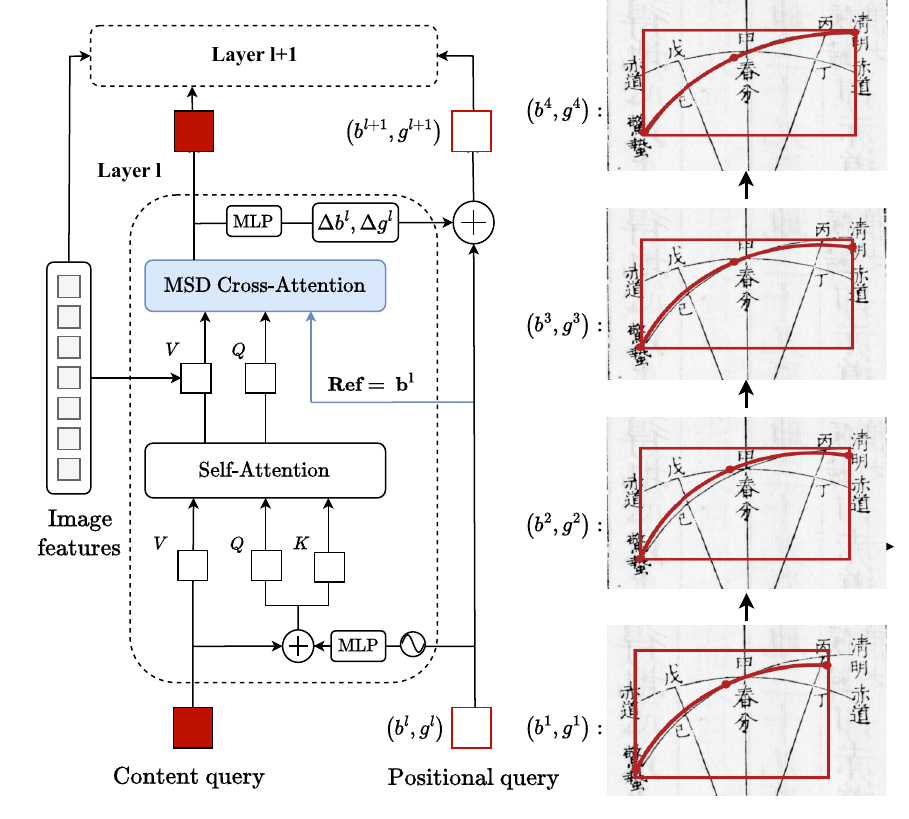}
        \caption{ {\bf Primitive refinement.} Each modified decoder block (left) updates the positional query $(\bbox{}^l, \params{}^l)$, which progressively refines predictions (right).%
        } %
        \label{fig:decoder_layer}
    \end{minipage}%
\end{figure}

Our transformer decoder's core mechanism revolves around refining primitives iteratively, and is inspired by successful object detectors like Deformable-DETR~\cite{zhu2020deformable} and DAB-DETR~\cite{liu2022dab}. These detectors have shown the effectiveness of updating box parameters layer-by-layer. Motivated by this paradigm, our approach refines at each decoder layer $l$ both the primitive bounding box $\bbox{}$ and primitive parameters $\params{}$ for all queries. Each layer receives an initial estimate of $\bbox{}^l$ and $\params{}^l$, and predicts updates $\Delta \bbox{}^{l}$ and $\Delta \params{}^{l}$, which are used to refine the estimates. %
In the following, we first detail the parameterization of the primitive bounding box $\bbox{}$ and parameters $\params{}$, then explain %
how we perform the parameter update in each decoder layer as illustrated in Figure~\ref{fig:decoder_layer}.

\subsubsection{Modified content queries parametrization}
Our parameterization of the decoder positional queries is depicted in Figure~\ref{fig:parameterization}. All values are normalized by the width and height of the image. We use the standard parametrization of the bounding box using its center, width and height. We concatenate these bounding box parameters $b$ to the parameterizations $\params{}{}$ corresponding to the different primitive types.  For line segments, we use the coordinates of the extremities in arbitrary order. For circles, we use the center as well as the radius normalized by the image width and by the image height. For arcs, we use the coordinates of both extremities in arbitrary order and the coordinates of the arc middle point. Using the primitive bounding box alongside the primitive parameters offers two key advantages. Firstly, it enables deformable attention, guiding sampling locations for queries using the bounding box in a way agnostic to the primitive class. Secondly, this approach does not constrain the model to a specific set of primitives or a specific parameterization, making it readily extendable to other geometric primitives, opposite to the joint parameterization used in PPI-Net~\cite{wang2023ppi} which shares the parameterization for several primitives.  %

The queries interact in the self attention module then probe the image features where the predicted box is used as a reference box for deformable attention, making each query focus exclusively on relevant regions of the image. The output of the decoder layer passes through a MLP to generate an update $(\Delta \bbox{}^{l}, \Delta \params{}^{l})$ relative to the previous estimate $(\bbox{}^{l}, \params{}^{l})$. Similarly to~\cite{zhu2020deformable, zhang2022dino}, the refined box and parameters are defined  as: 
\begin{equation}
\bbox{}^{l+1} = \sigma( \Delta \bbox{}^{l} + \sigma^{-1}(\bbox{}^{l})), \quad \params{}^{l+1} = \sigma( \Delta \params{}^{l} + \sigma^{-1}(\params{}^{l}) ) ,\label{eq:refinement_param}
\end{equation}
where $\sigma$ is a sigmoid. We find this refinement strategy particularly suitable for our vectorization task, where precise parameter predictions are crucial and we validate the importance this approach in Section~\ref{sec:expe}.

\subsection{Losses}
\label{sec:losses}

Our model outputs a fixed number of predictions $\Nqueries$ typically much larger than the number of ground-truth primitives. To simplify the notations, we add ``no primitive" elements to the ground truth so that we also have $\Nqueries$ ground truth for each image.
 Given $n\in\{1,...,N\}$, we have $(p_{n}, \bbox{n}, \params{n})$ the predictions associated to query $n$ and we call $(\class{n}^{gt}, \bbox{n}^{gt}, \params{n}^{gt})$ the $n$-th ground truth, with $\class{n}^{gt}$ its class, $\bbox{n}^{gt}$ its bounding box and $\params{n}^{gt}$ its parameters, defined only for class $\class{n}^{gt}$. 

\paragraph{Bipartite Matching} To define our training loss, we establish a one-to-one correspondence between predictions and ground-truth, similar to the approach of DETR~\cite{carion2020end} but considering the predicted primitive parameters. 
We use the Hungarian algorithm to find the permutation ${\pi}$ which minimize the matching cost:
\begin{equation}
    \resizebox{\linewidth}{!}{
    $\displaystyle
    \sum_{n=1}^N   
    \left(
    - 
    \lambda_{\text{cls}} 
    \score{n}{\class{\pi[n]}^{gt}} 
    +  
    \mathbf{1}_{ \{\class{\pi[n]}^{gt} \ne \emptyset \} } 
    \left( 
    \lambda_{\text{box}}
    d_{\text{box}} (b_n, {b}_{\pi[n]}^{gt}) 
    + \lambda_{\text{param}} 
    d_{\class{\pi[n]}^{gt}} (\params{n}[\class{\pi[n]}^{gt}], \params{\pi[n]}^{gt}) \right)
    \right)
    $}, 
\label{eq:match_cost}
\end{equation}
where $\lambda_{\text{cls}}$, $\lambda_{\text{box}}$, and $\lambda_{\text{param}}$ are scalar hyperparameters, $d_{\text{box}}$ is defined by
\begin{equation}
d_{\text{box}} (\bbox{n}, \bbox{\pi[n]}^{gt}) = 
 \Lone{\bbox{n} -  \bbox{\pi[n]}^{gt}} , \label{eq:box_dist} 
\end{equation}
and $d_{\class{}}$ is a distance function tailored to each primitive class. For each primitive class $\class{}$, we define $d_{\class{}}$ as the minimum, over all equivalent parametrizations, of the $L_1$ distance between the parameters normalized by the number of parameters. For instance, in the case of lines, the distance function $d_{\text{line}}$ is computed as the minimum over both permutations of the endpoints of the $L_1$ distance between the predicted and ground-truth endpoints, divided by 4. %
Unlike traditional object detectors, we do not consider an intersection over union for $d_{\text{box}}$ to avoid instability for vertical and horizontal primitives.

\paragraph{Loss} 
Once the permutation $\pi$ between predictions and ground truth is defined, we compute the loss function for each query as a weighted sum of a classification loss $\mathcal{L}_{\text{cls}}$, a box loss $\mathcal{L}_{\text{box}}$ and a parameters loss $\mathcal{L}_{\text{param}} $: 

\begin{equation}
    \mathcal{L}(p_{n}, \bbox{n}, \params{n}) = \lambda_c \cdot \mathcal{L}_{\text{cls}} + \lambda_{\text{box}} \cdot \mathcal{L}_{\text{box}} + \lambda_{\text{param}} \cdot \mathcal{L}_{\text{param}},
\end{equation}
where where $\lambda_{\text{cls}}$, $\lambda_{\text{box}}$, and $\lambda_{\text{param}}$ are scalar hyperparameters as in equation~\ref{eq:match_cost}, we use the focal loss~\cite{lin2017focal} as classification loss $\mathcal{L}_{\text{cls}}$,
\begin{align}
    \mathcal{L}_{\text{box}}
    &= \mathbf{1}_{\{\class{\pi[n]}^{gt} \ne \emptyset \}}   d_{\text{box}} (\bbox{n}, \bboxhat{\sigma(n)}), \ \text{and} \\
    \mathcal{L}_{\text{param}} & =   d_{\class{\pi[n]}^{gt}} (\params{n}[\class{\pi[n]}^{gt}], \params{\pi[n]}^{gt}) ,
\end{align}
where $d_{\text{box}}$ and  $d_{c}$ are the box and parameter distances defined in the previous paragraph.

\paragraph{Contrastive denoising}
Similarly to DINO-DETR~\cite{zhang2022dino}, we also adapt contrastive denoising to speed up convergence and improve performance by training a denoising decoder branch. The architecture of this branch is the same as our decoder but content queries are replaced by class labels, and it is trained in parallel without information sharing. From ground truth primitive parameters and labels, we create positive and negative noisy primitive queries, %
associated to two hyper parameters $\lambda_{pos}$ and $\lambda_{neg}$ controlling the strength of the added noise, more noise being added to the negative group. %
The primitive class can also be noised through label flipping, the class of the primitive being switched to another with probability $p_{\text{flip}}$, which we found to have little impact on our results. %
The denoising branch is trained with the same loss as the rest of the network.

\begin{figure}[t]
    \centering
    \begin{subfigure}[b]{0.70\textwidth}
        \centering
        \includegraphics[width=\textwidth]{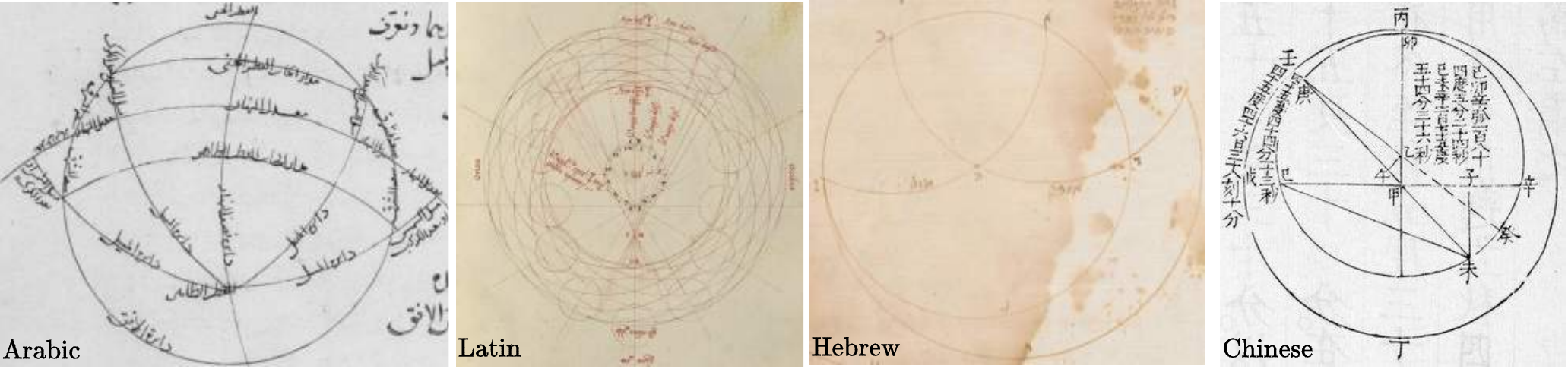}
        \caption{Example diagrams}
        \label{subfig:dataset_sample}
    \end{subfigure}
    \hfill
        \begin{subfigure}[b]{0.29\textwidth}
        \centering
        \includegraphics[width=\textwidth]{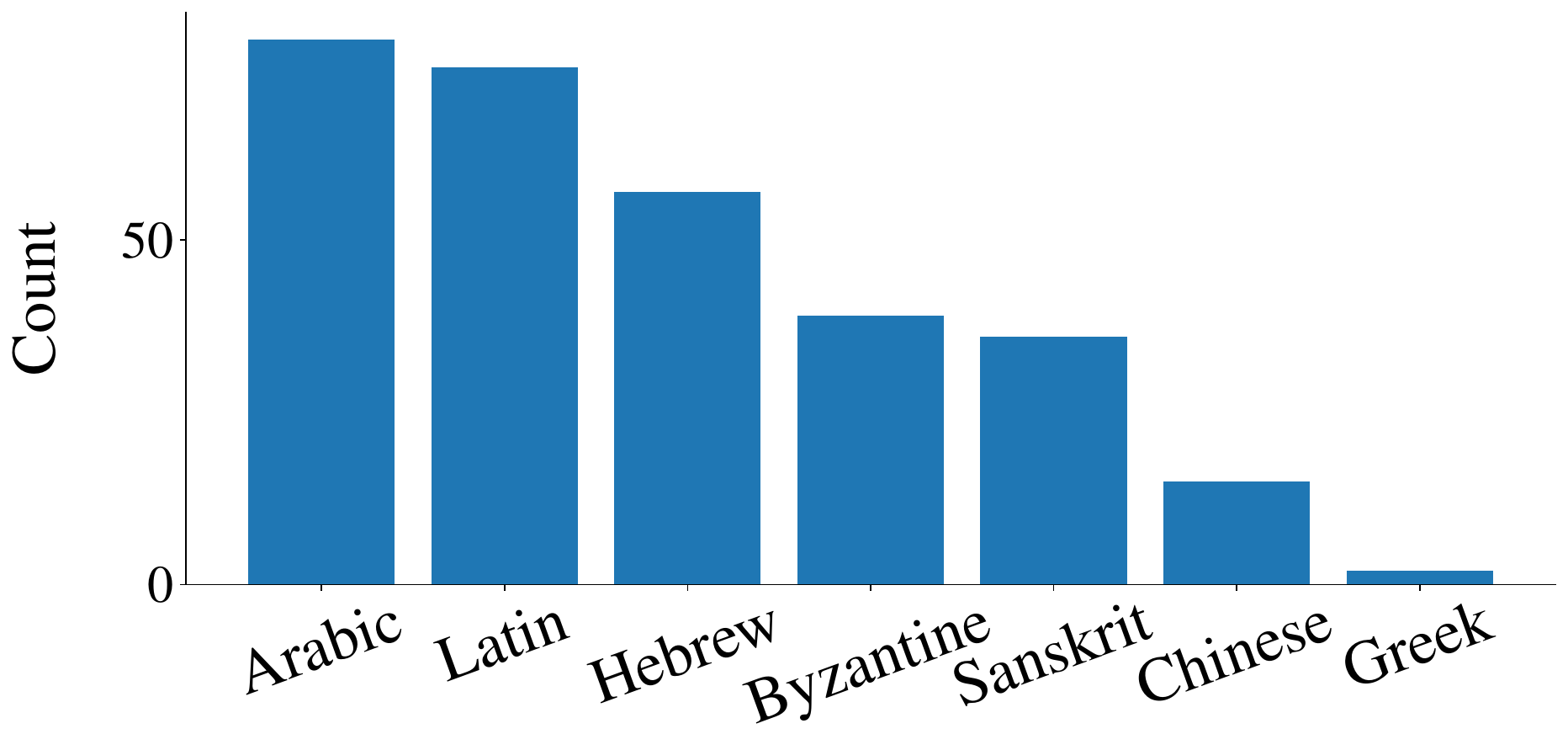}
        \caption{Diagrams' origins}
        \label{subfig:histogram_manuscripts}
    \end{subfigure}
    \vskip\baselineskip
    
      \begin{subfigure}[b]{0.34\textwidth}
        \centering    
        \begin{tabular}{lcc}
            \hline
            & \#Primitives & \#Images \\
            \hline
            Line & 1794 & 261 \\
            Circle & 955 & 263 \\
            Arc & 327 & 84 \\
            \hline
             & \\
        \end{tabular}
        \caption{Primitive statistics}
        \label{subfig:dataset_stats_table}
    \end{subfigure}
    \hfill
          \begin{subfigure}[b]{0.64\textwidth}
        \centering
        \includegraphics[width=\textwidth]{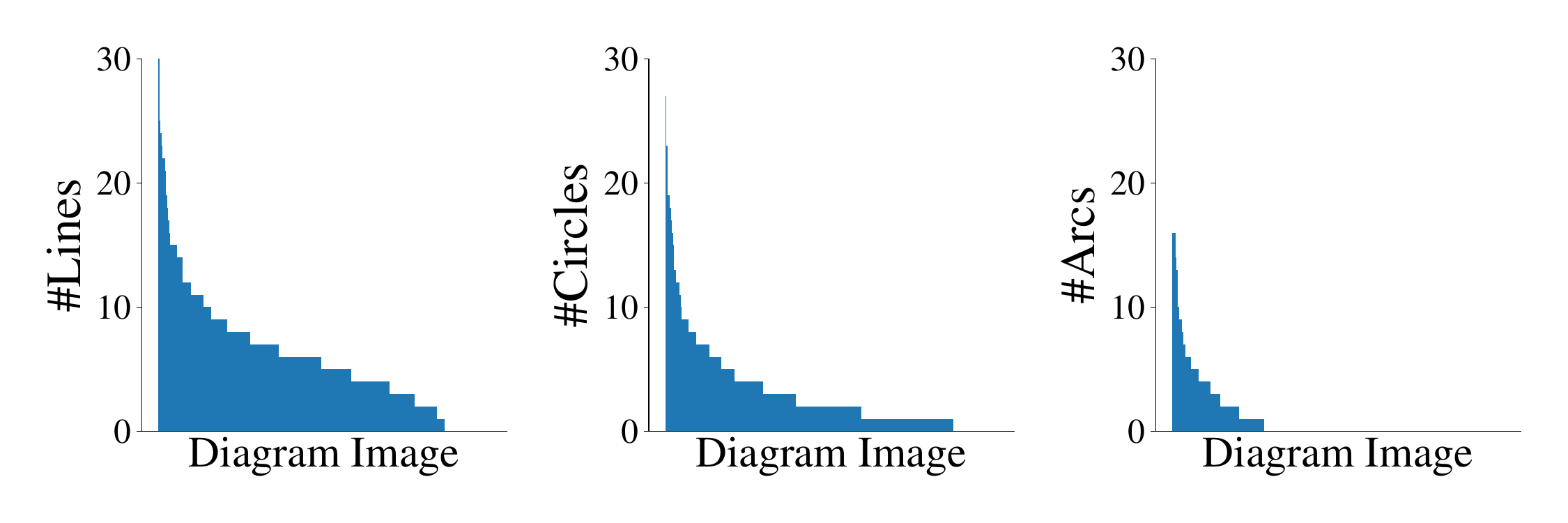}
        \caption{Number of primitives per image}
        \label{subfig:distribution_primitives}
    \end{subfigure}
    \caption{{ \bf Dataset characteristics.} Our diagrams have varying complexity and several challenges are introduced by document deterioration, heavy presence of text, and overlapping primitives.}
    \label{fig:dataset_figure}
\end{figure} 

\subsection{Implementation and training details}
Our feature maps are extracted from stages $C_3$ through $C_5$ of a ResNet50~\cite{he2016deep} pretrained on ImageNet-1k. A fourth feature map is extracted by down-sampling the last stages features as done in~\cite{zhang2022dino}. A $1 \times 1$ convolution reduces the channel dimensions to the hidden dimension of the transformer $256$. We keep the standard $H=8$ heads and $k=4$ neighbors for deformable attention~\cite{zhu2020deformable}. We use a $6$-layer encoder and $6$-layer decoder. Each decoder layer is followed by a shared $3$-layer FFN and each decoder layer outputs its own loss similarly to the training scheme in DETR~\cite{carion2020end}. We use $\Nqueries = 900$. %
The matching cost and loss hyperparameters are set to $\lambda_{\text{cls}} = 2.0 $ and $\lambda_{\text{box}} = \lambda_{\text{param}} = 5.0$. 
We apply scale augmentation by resizing the input images such that the shortest side is at least 480 and at most 800 pixels while the longest is at most 1333. Additionally, we apply instance-aware random crops that do not cut the diagrams, random flips, and random 90 degree rotations. We train for $12$ epochs, defined as 2000 gradient steps, with batches of size 4, generating synthetic data on the fly, and using an initial learning rate of $10^{-5}$ for the backbone and $10^{-4}$ for the rest of the network, where both learning rates are divided by 10 at epoch $11$. %
Our training lasts approximately $12$ hours on an A6000 GPU. Our code will be made available after publication.

\section{Dataset}
\label{sec:dataset}
\paragraph{Diagrams selection and annotation}
Our team of historians has curated a dataset of 303 diagrams so that it spanned diverse relevant traditions for the history of astronomy, namely Arabic, Latin, Hebrew, Byzantine, Sanskrit, Chinese, and Greek sources. Examples of diagrams from different traditions are presented in Figure~\ref{subfig:dataset_sample} and the proportion of diagrams from each tradition is shown in Figure~\ref{subfig:histogram_manuscripts}. 
This dataset includes diagrams from 27 distinct documents from the 12th to the 18th century, most of them manuscripts as well as a few Chinese woodblock prints (which were common in ancient China). This ensures a broad diversity of content, representations, scripts, styles, materials, digitization quality, and conservation state. These diagrams were annotated by the historians with circles, arcs and line segments relevant for their analysis, which resulted in a total of 3076 annotated primitives, summarized in Figure~\ref{subfig:dataset_stats_table}. There is a high variation in the number of primitive per diagram, which can be seen in Figure~\ref{subfig:distribution_primitives}, where we show the number of each type of primitive for the different images. 
This dataset was used solely to evaluate our algorithms, not for training.

\begin{figure}[t]
    \centering
    \begin{tabular}{cccccc}
        \includegraphics[width=0.19\linewidth, height=0.19\linewidth]{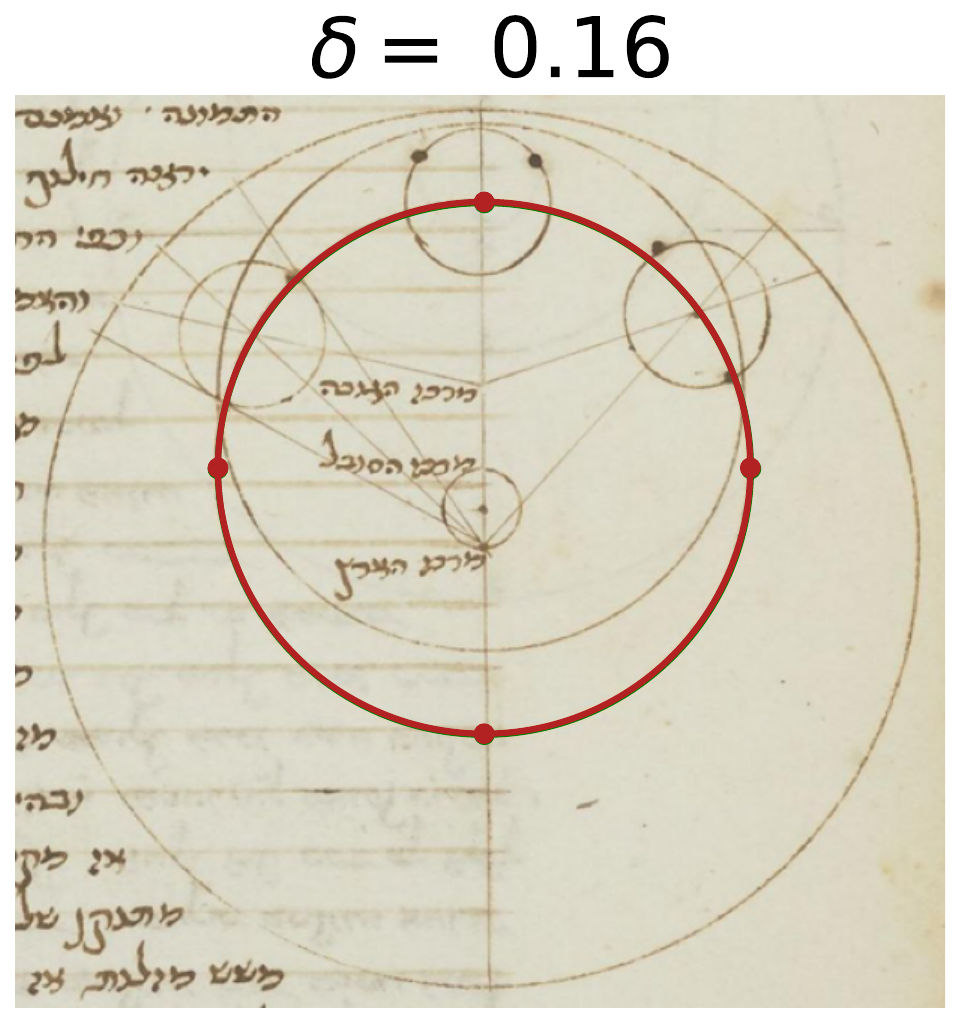} &
        \includegraphics[width=0.19\linewidth, height=0.19\linewidth]{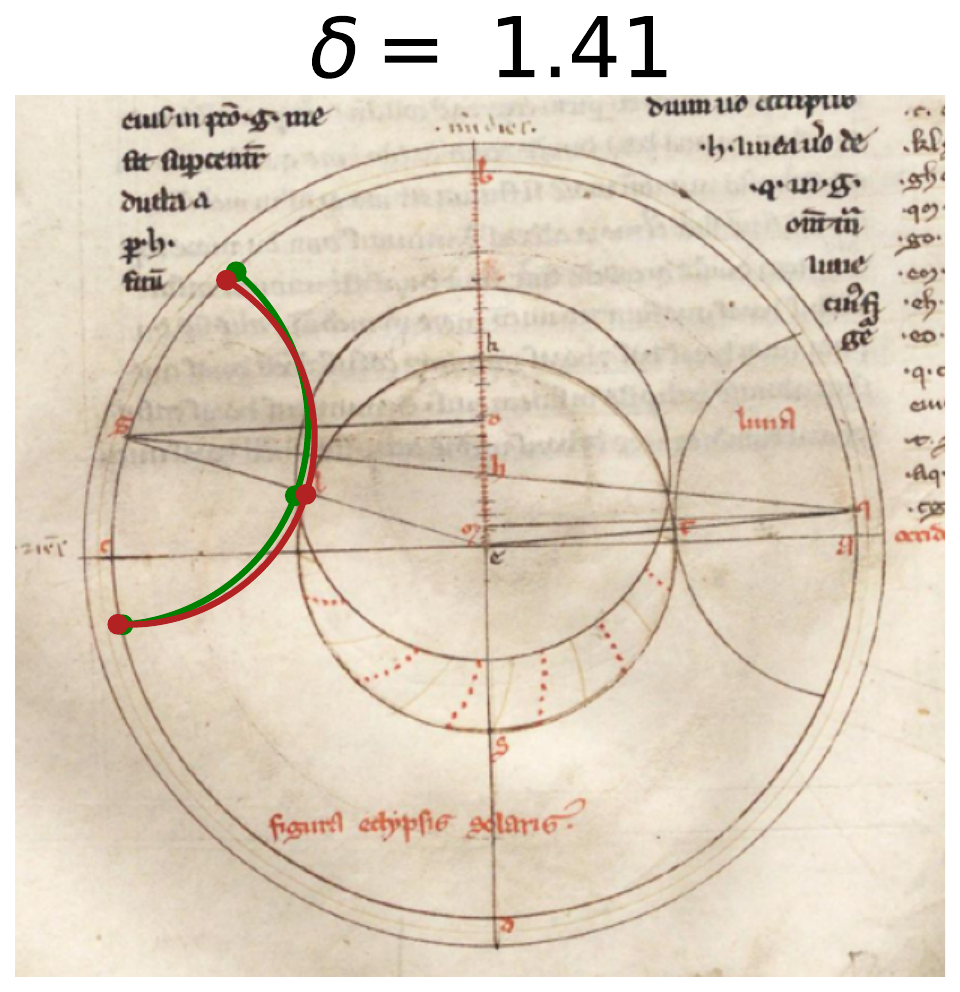} &
        \includegraphics[width=0.19\linewidth, height=0.19\linewidth]{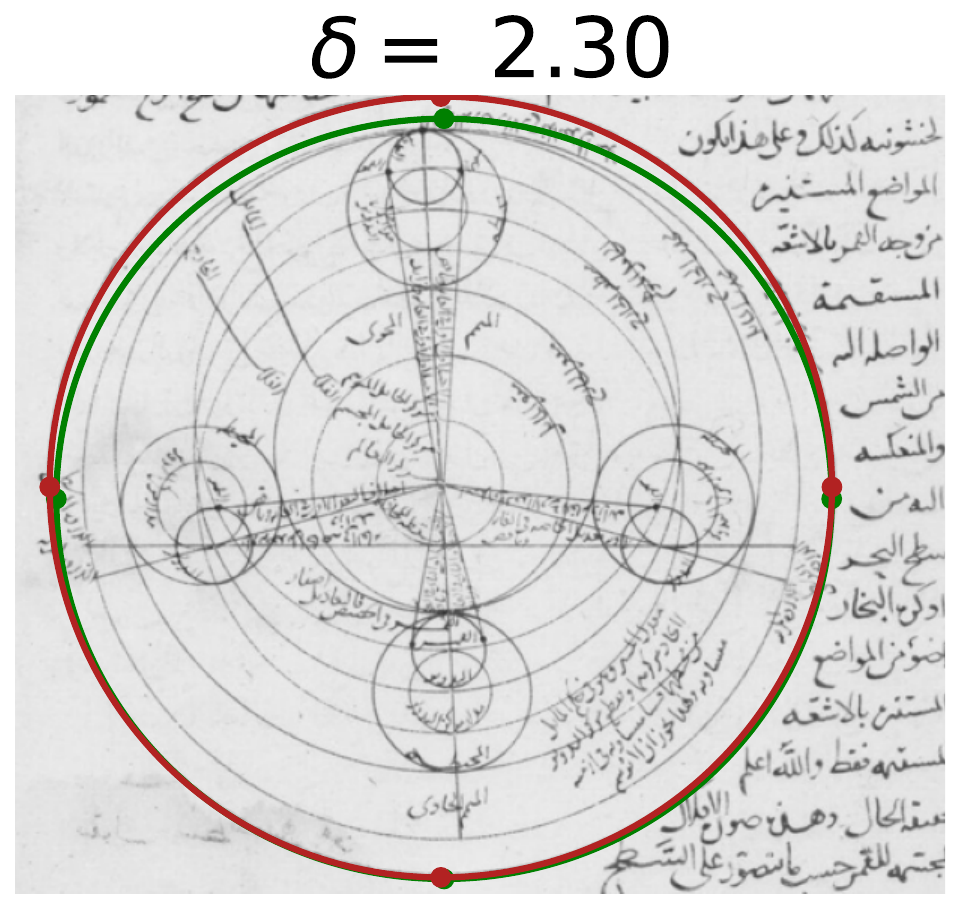} &
        \includegraphics[width=0.19\linewidth, height=0.19\linewidth]{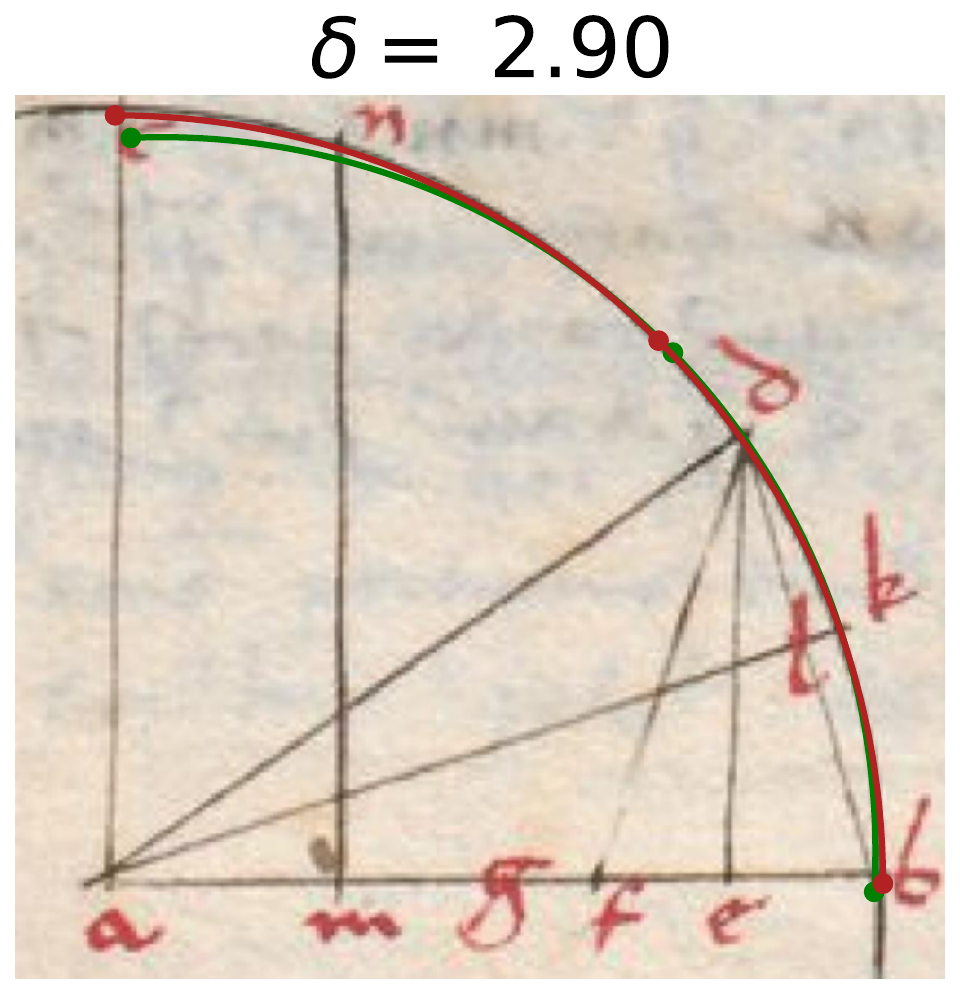} &
        \includegraphics[width=0.19\linewidth, height=0.19\linewidth]{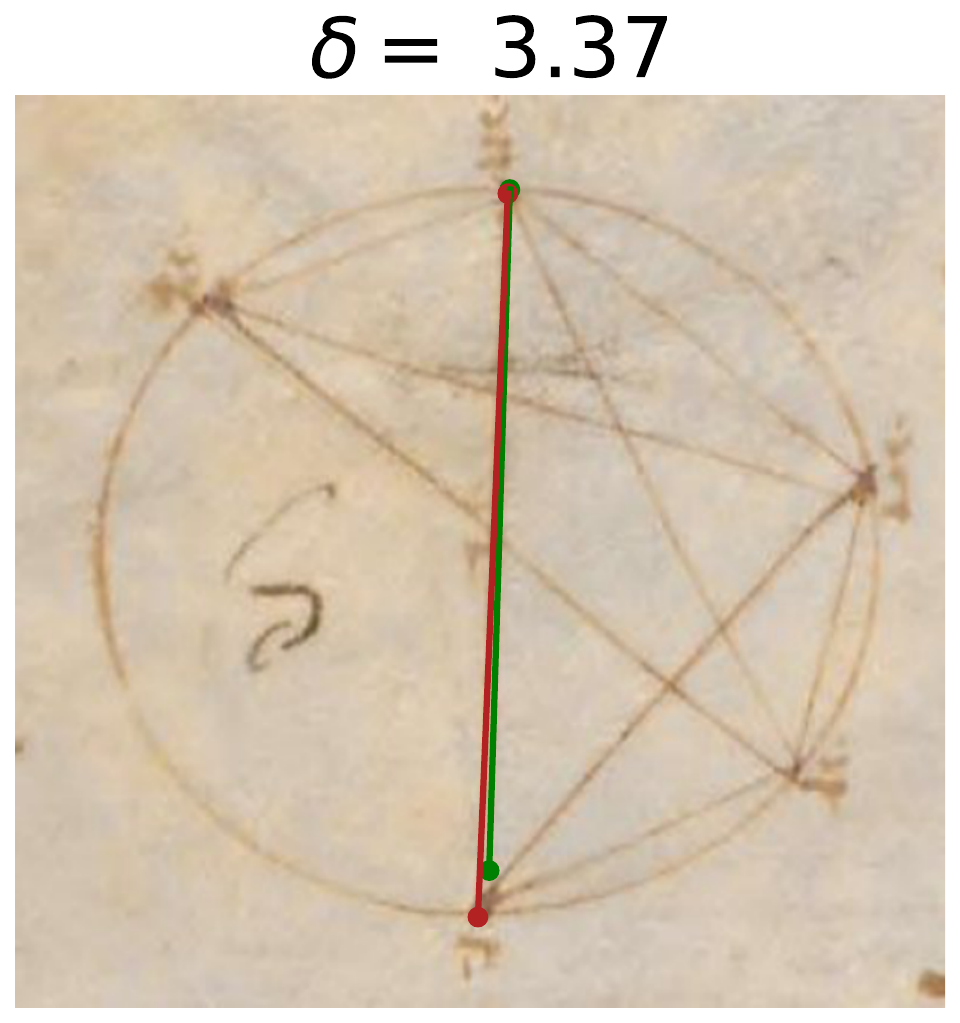} \\

        \includegraphics[width=0.19\linewidth, height=0.19\linewidth]{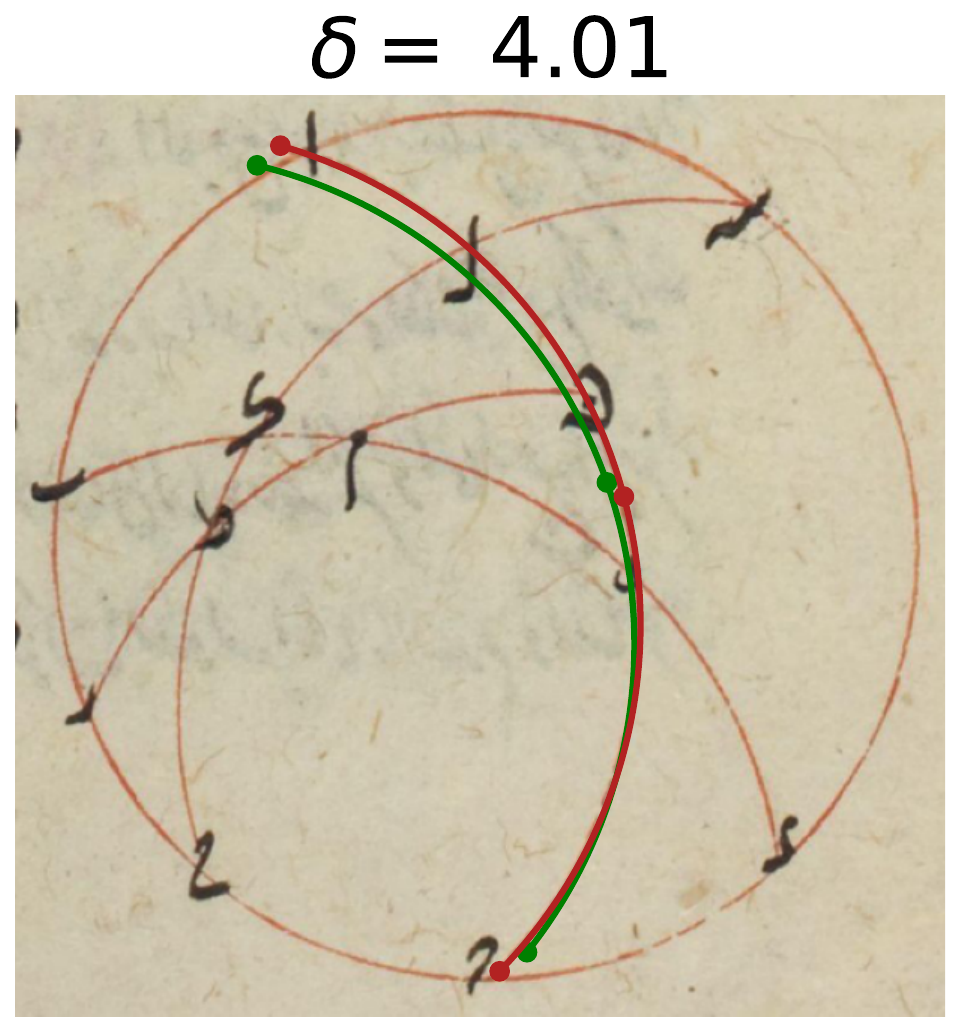} &
        \includegraphics[width=0.19\linewidth, height=0.19\linewidth]{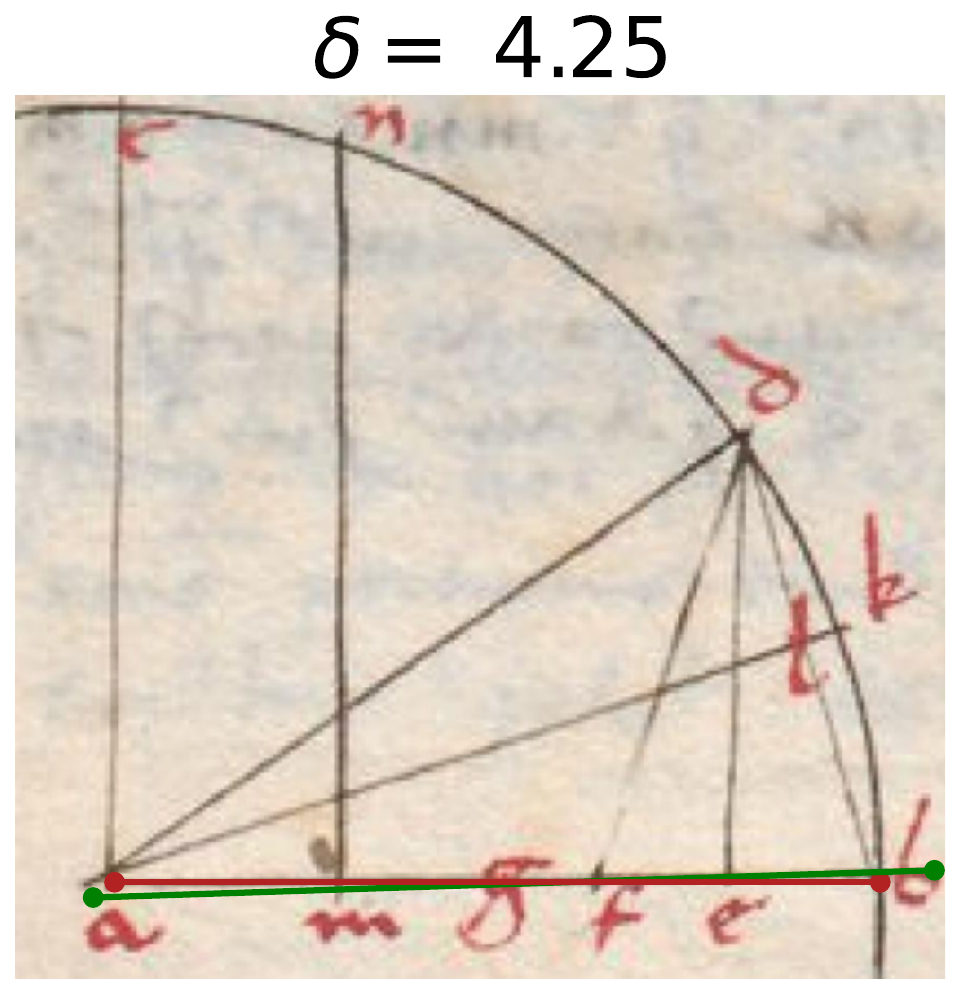} &
        \includegraphics[width=0.19\linewidth, height=0.19\linewidth]{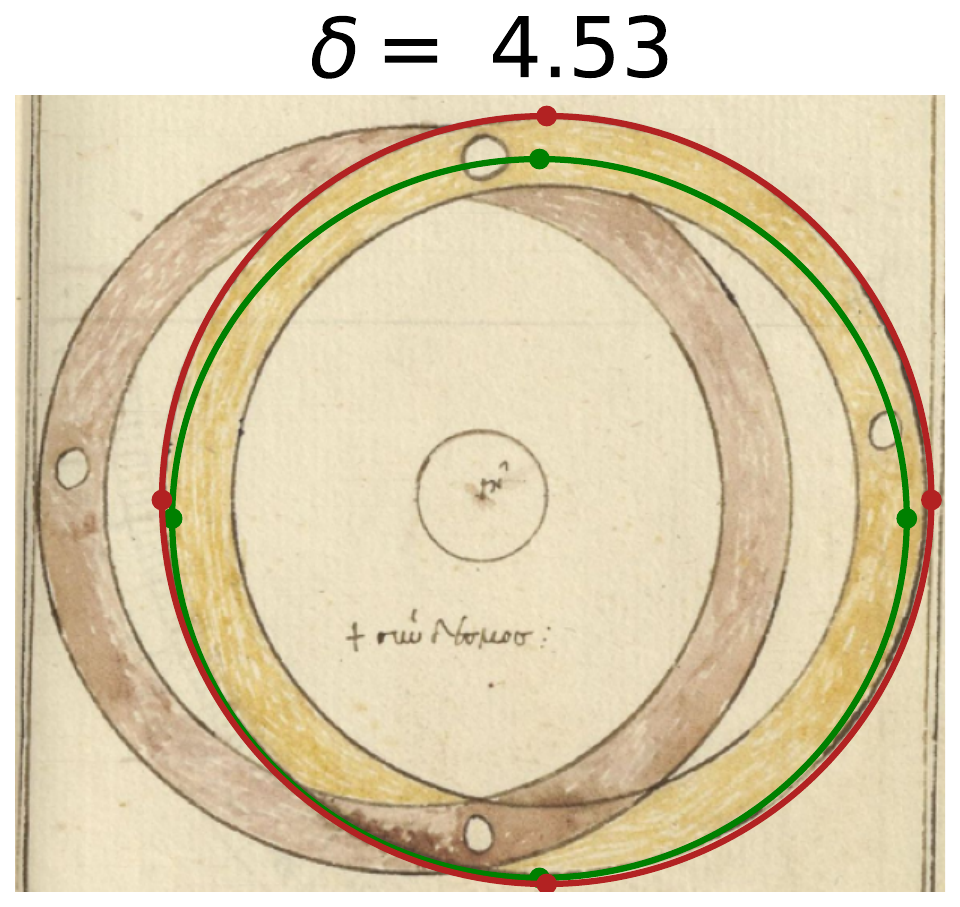} &
        \includegraphics[width=0.19\linewidth, height=0.19\linewidth]{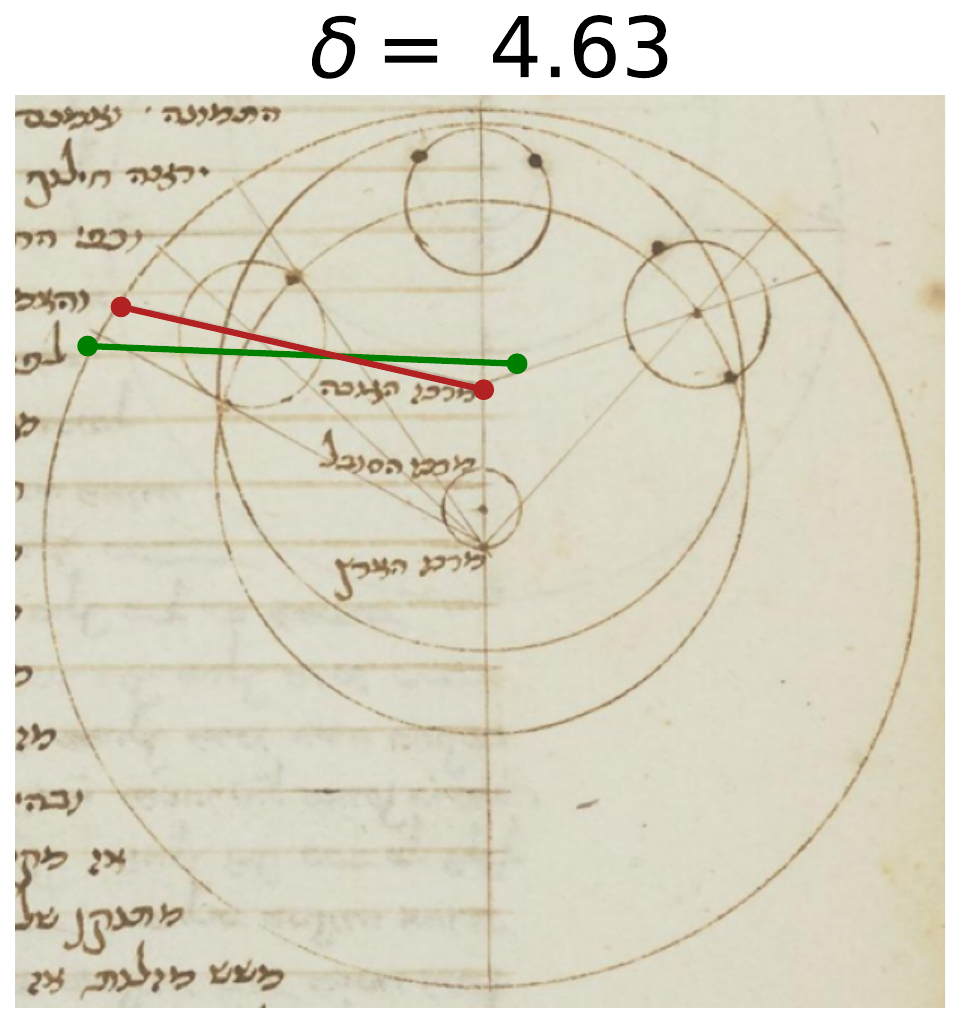} &
        \includegraphics[width=0.19\linewidth, height=0.19\linewidth]{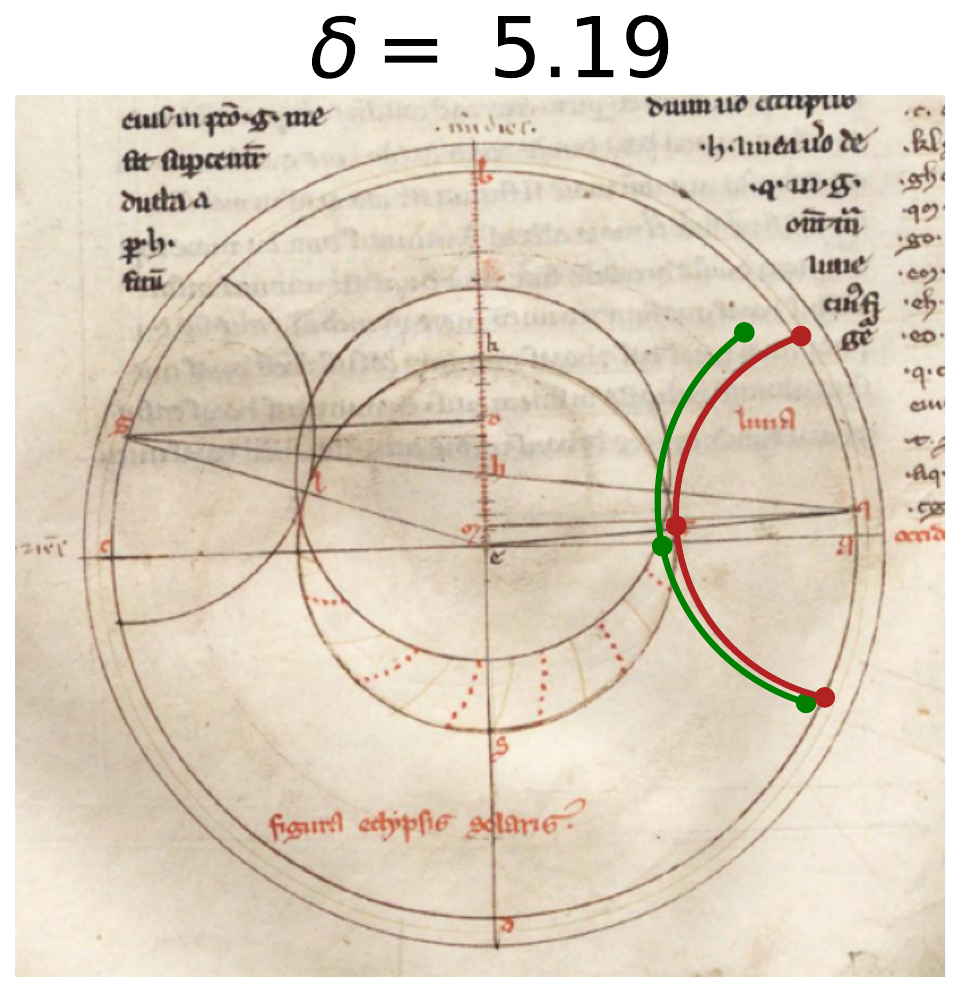}

    \end{tabular}
    \caption{{\bf Primitive distance examples.} We display for each prediction (green) the closest ground-truth (red) and the distance $\delta$ between them. By default, we set the threshold to define valid predictions to $\delta_{\text{max}}=4$, for which examples on the top row are considered true positives, and all examples on the bottom row are false positives.}
    \label{fig:eval_example}
\end{figure}

\paragraph{Evaluation metrics} 
Given a set of primitives, we first match predicted primitives to the closest ground truth. Then, for each ground-truth, we consider the closest predicted primitive a True Positive if its distance $\delta$ is less than a predefined threshold $\delta_{\text{max}}$. All other predictions are considered False Positives. We define the evaluation distance $\delta$ between primitives as the average of the point-wise $L_2$ distance between a set of predefined point depending on the primitive type in the image resized to $128\times128$ pixels: the endpoints for lines, the four cardinal points for circles, and the endpoints and middle point for arcs. %
Unless stated otherwise, we use $\delta_{\text{max}}=4$ as our threshold. Examples illustrating this evaluation distance and true and false positives for each primitive type are visualized in Figure~\ref{fig:eval_example}, giving a sense of the prediction accuracy necessary to be considered a true positive.

\begin{figure}[t]
    \centering
    \includegraphics[width=0.19\linewidth, height=0.24\linewidth]{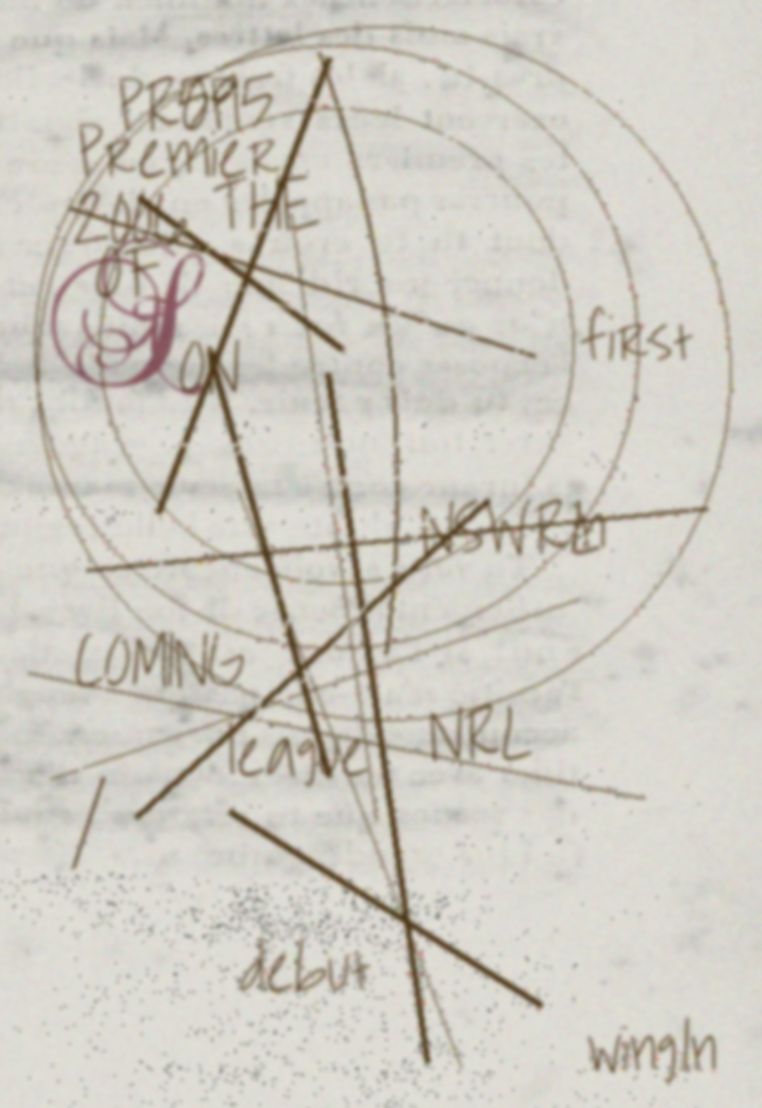}
    \includegraphics[width=0.19\linewidth, height=0.24\linewidth]{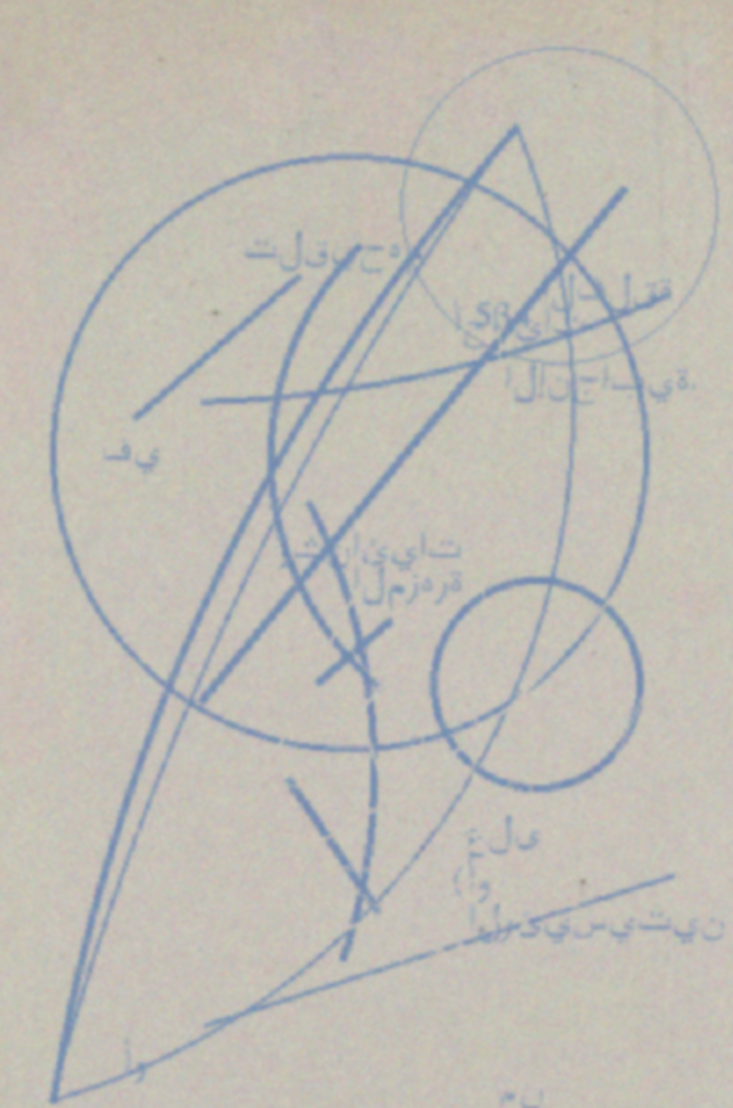}
    \includegraphics[width=0.19\linewidth, height=0.24\linewidth]{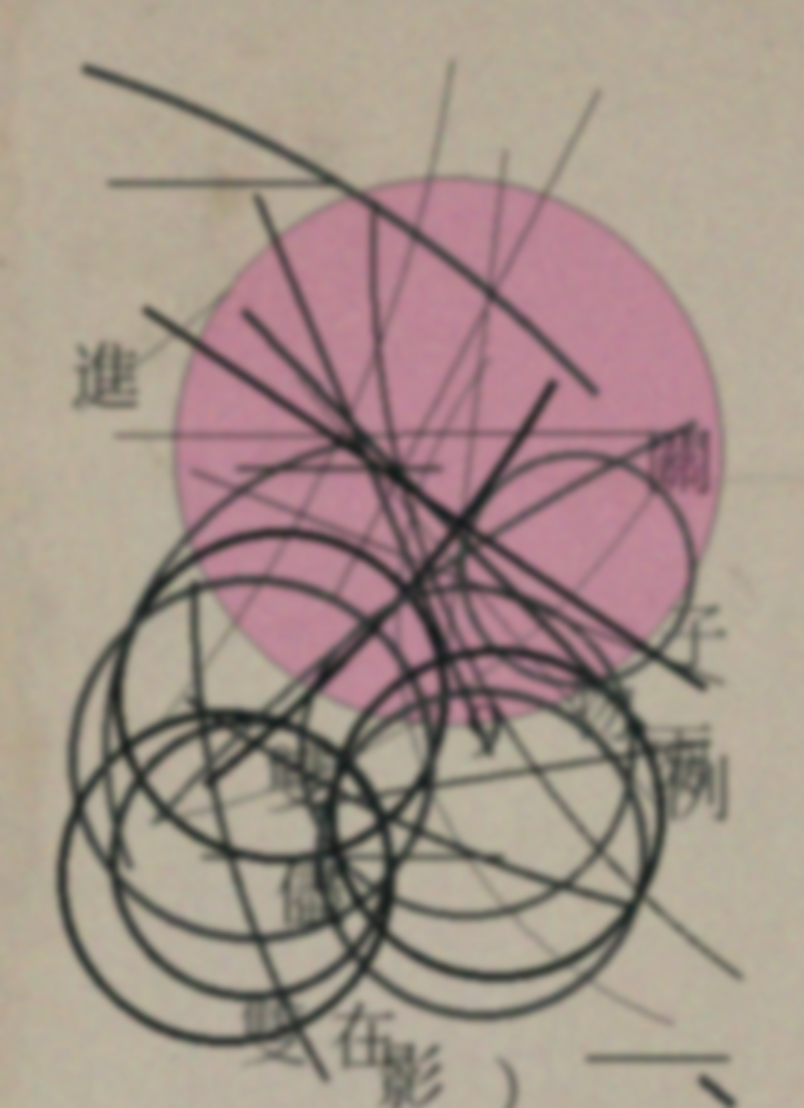}
    \includegraphics[width=0.19\linewidth, height=0.24\linewidth]{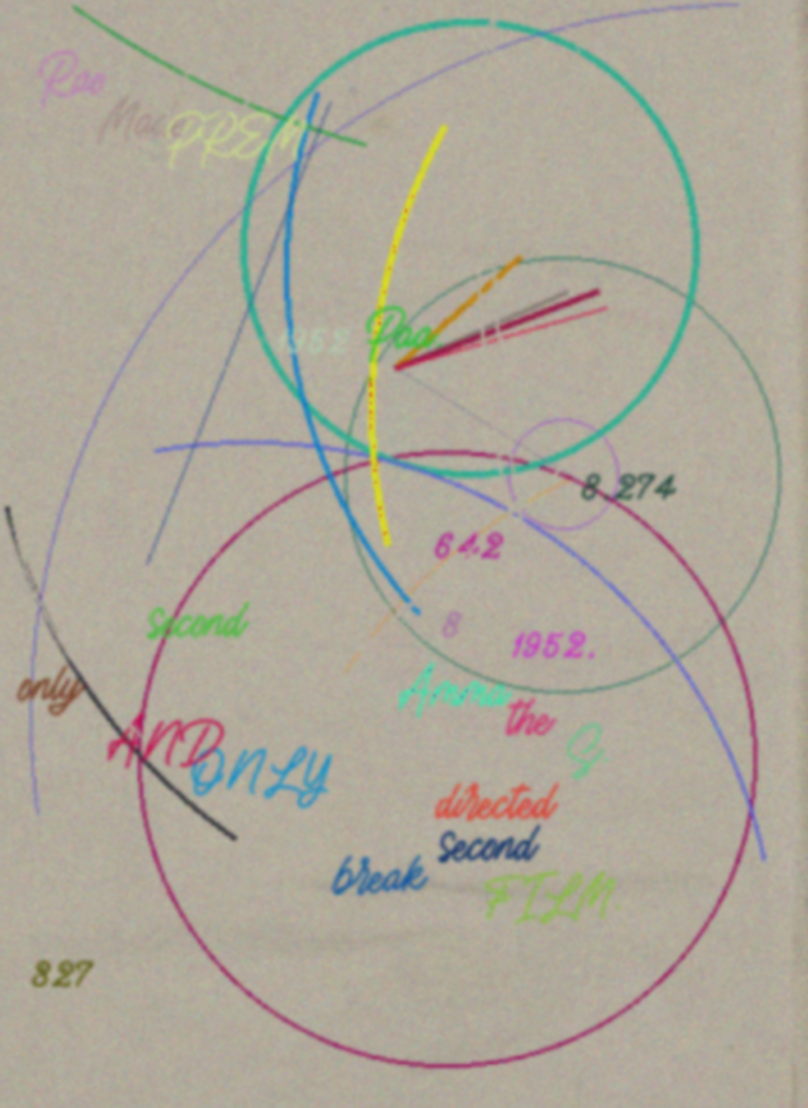}
    \includegraphics[width=0.19\linewidth, height=0.24\linewidth]{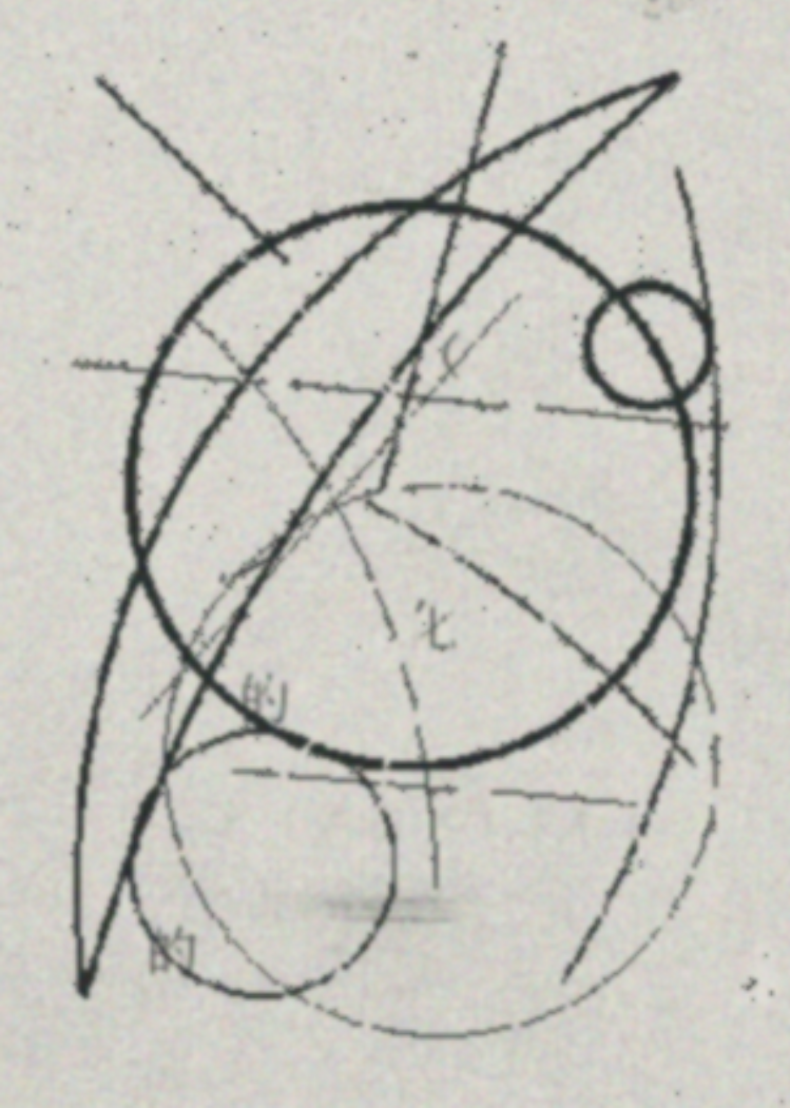}
    
    \caption{{\bf Synthetic dataset examples.} While our synthetic samples are not realistic, they are challenging enough for the network to generalize to real diagrams. }
    \label{fig:synthetic}
\end{figure}

\paragraph{Synthetic training data}
Because annotating geometric primitives in complex examples is very time consuming, we train our models relying solely on dynamically generated synthetic examples. Here, we outline the key elements of this generation process, and we show a sample of generated synthetic data in Figure~\ref{fig:synthetic}. Our data generation code is available on our project webpage.
We leverage the resources of backgrounds, text, and glyph fonts proposed in~\cite{monnier2020docExtractor}. To generate an example, we first randomly sample a background 
randomly and add floating words, random numbers and glyphs in different fonts. %
We then generate a diagram by generating line segments, circles and arcs. Because there is very little chance that randomly generated primitives would generate challenging configurations often present in diagrams, we sample with higher probability specific configurations such as co-centric and tangent circles, parallel lines, and connected arcs which share one or two endpoints.  These primitives are drawn with random opacity variation, random width and colors on the backgrounds. Circles can be filled or hollow. 
Finally, we add Gaussian blur, structured noise though randomly generated shapes, random resizing, and we chip small regions of the final diagram in order to simulate historical document degradation.

\section{Experiments}
\label{sec:expe}

\begin{table}[t]
    \centering
    \caption{Model comparison and ablation. The terms DA, QS and CDN denote Deformable Attention, Query Selection and Contrastive DeNoising.}
    
    \begin{tabular}{l@{\hspace{2em}}llll@{\hspace{2em}}llll}
        \hline
        Model &  Scales & DA & QS & CDN & $\text{AP}_\text{line}$  & $\text{AP}_\text{circle}$  & $\text{AP}_\text{arc}$ & \textit{m}AP \\
        \hline
        LETR~\cite{xu2021line} & 2 & & & & $0.263$ & - & - & - \\
        LETR-line (our data) & 2 & & & & 0.371 & - & - & - \\
        LETR-circle (our data) & 2 & & & & - & 0.754 & - & - \\

        Ours & 4 & \checkmark & \checkmark & \checkmark & \textbf{0.764}  & \textbf{0.917} & 0.483 & \textbf{0.722} \\

        \hline 
        \hspace{1em} w/o primitive refinement & 1 & $\checkmark$ &  &  & 0.490	 & 0.841 & 0.262	 &  0.531\\
        \hspace{1em} w/o multi-scale features & 1  & \checkmark &  & \checkmark & 0.599 & 0.887 & 0.216 & 0.567  \\
        \hspace{1em} w/o denoising & 4 & \checkmark & \checkmark &  & 0.739 & 0.904 & 0.452 & 0.698 \\
        \hspace{1em} w/o label flipping & 4 & \checkmark & \checkmark & \checkmark & 0.759 & 0.903 & \textbf{0.494}  & 0.719 \\
        \hspace{1em} w/ pure query selection & 4 & \checkmark & \checkmark & \checkmark & 0.756 & 0.917 & 0.476 & 0.717 \\
        \hspace{1em} w/o  query selection & 4 & \checkmark &  & \checkmark &  0.729 &	0.904	  & 0.390 & 0.674 \\
        \hline
    \end{tabular}
    \label{tab:results}
\end{table}

\begin{figure}[h]
    \centering
    \begin{tabular}{cccccc}
        \includegraphics[width=0.15\linewidth]{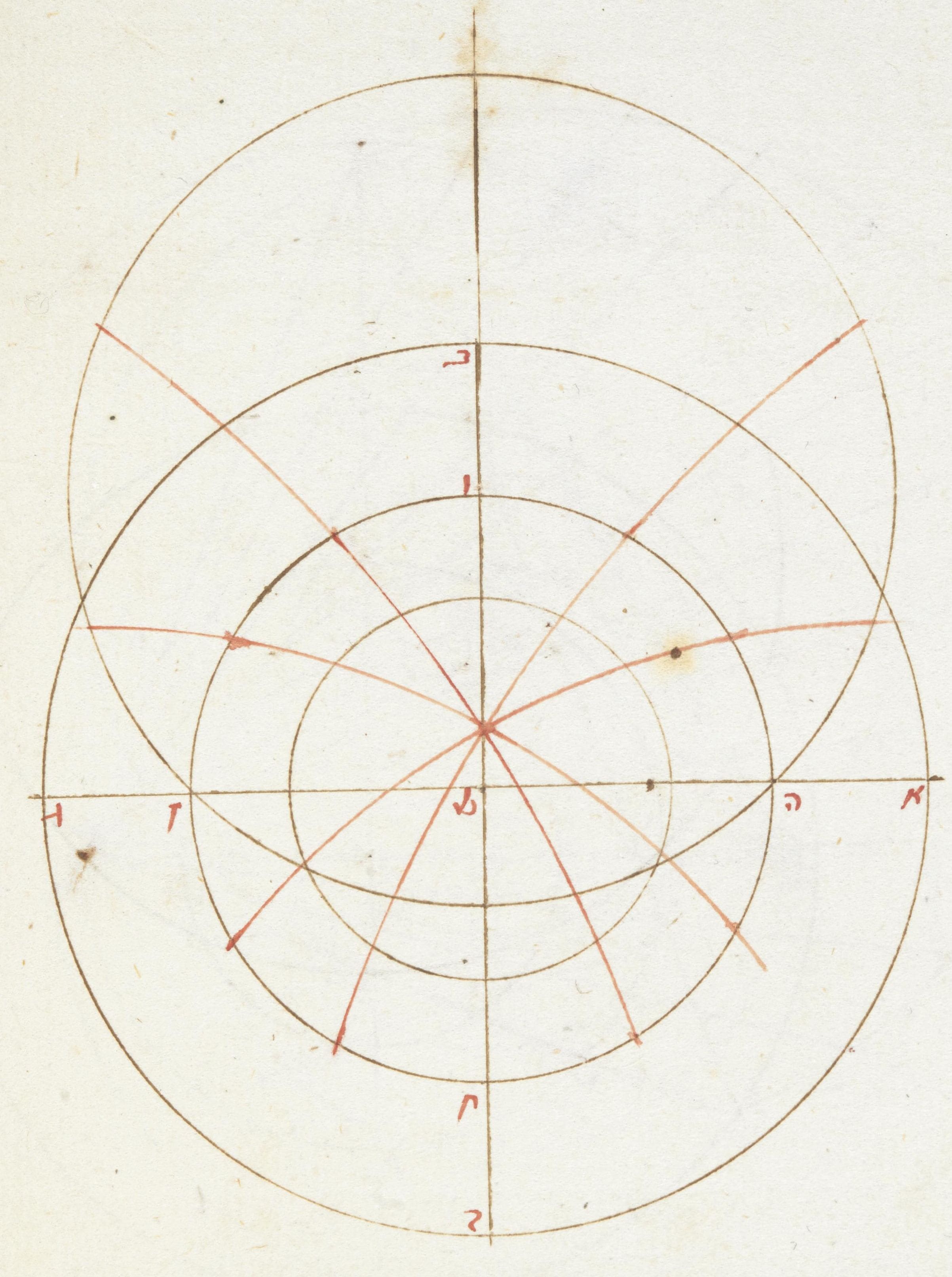} &
        \includegraphics[width=0.15\linewidth]{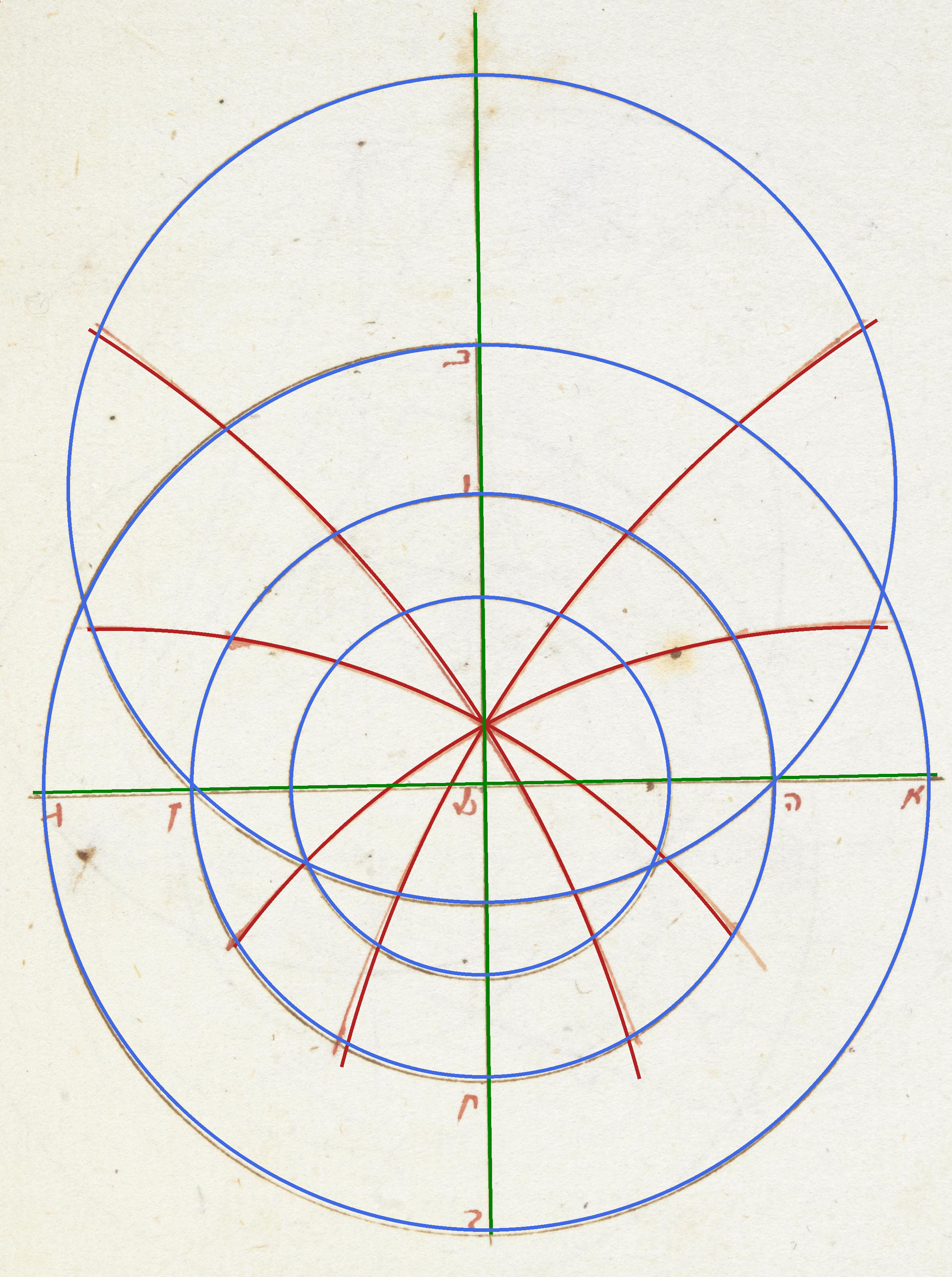} &
        \includegraphics[width=0.15\linewidth]{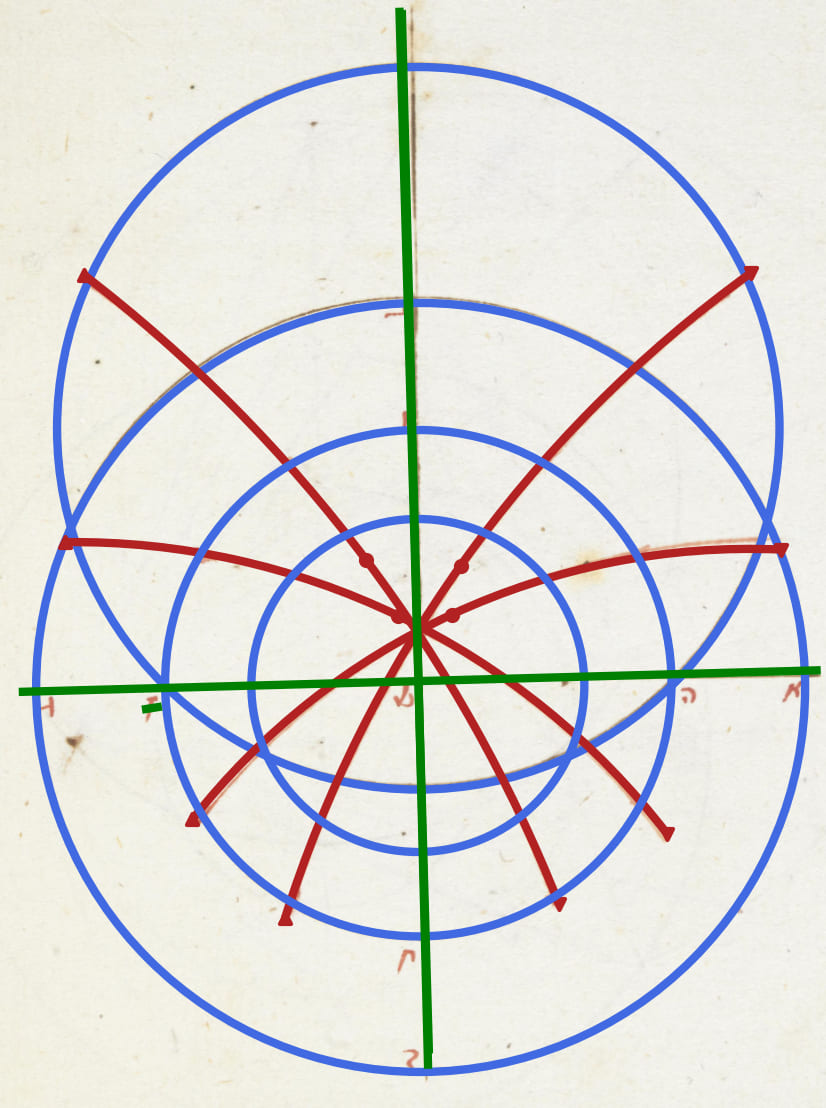} &
        \includegraphics[width=0.15\linewidth]{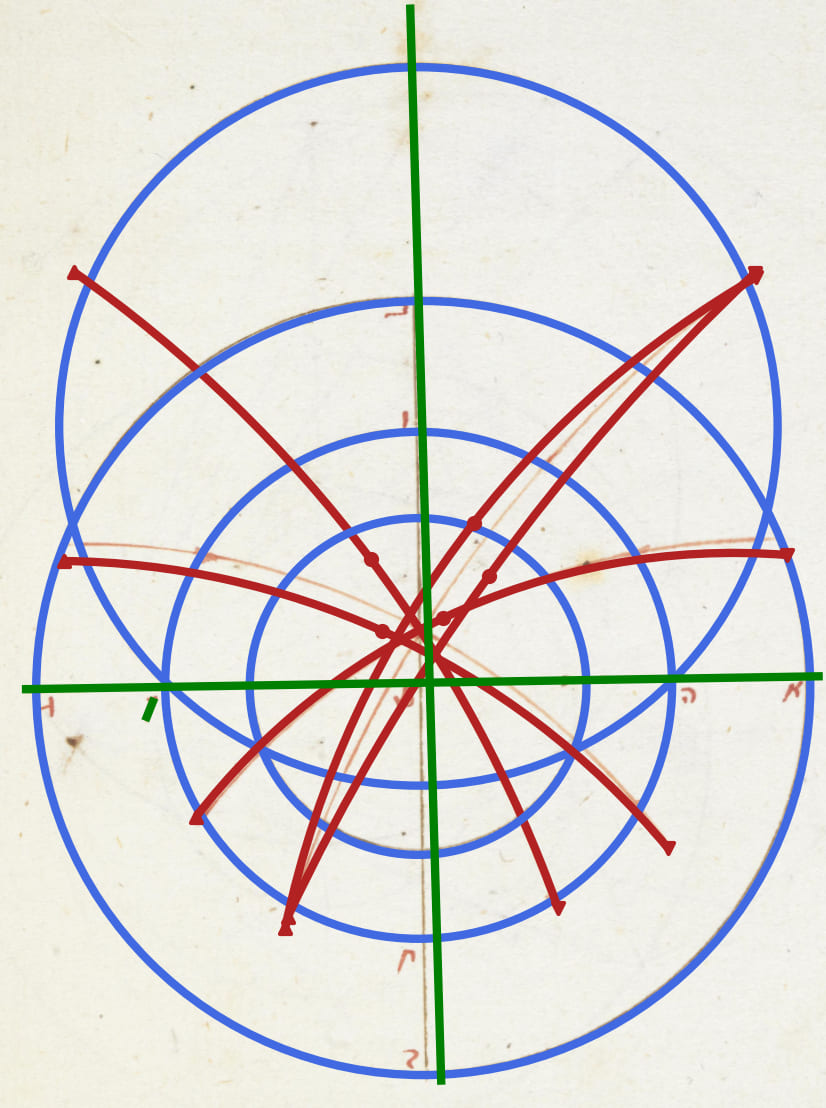} &
        \includegraphics[width=0.15\linewidth]{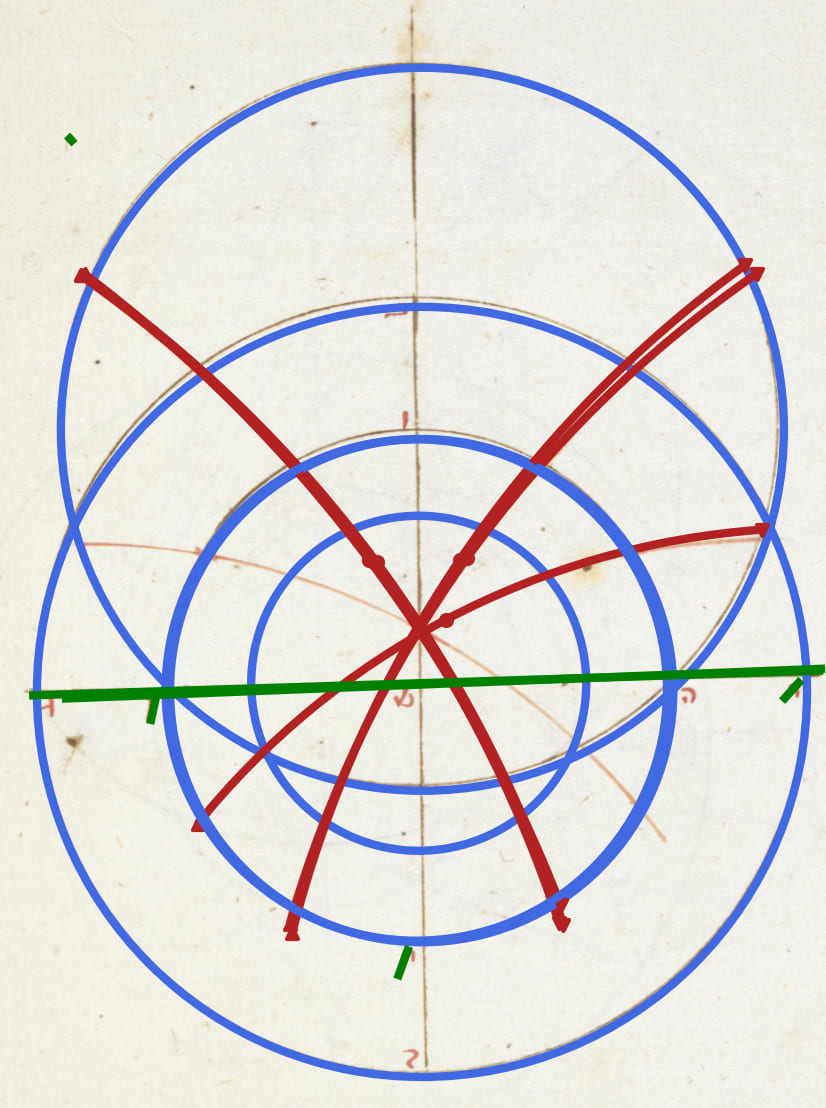} &
        \includegraphics[width=0.15\linewidth]{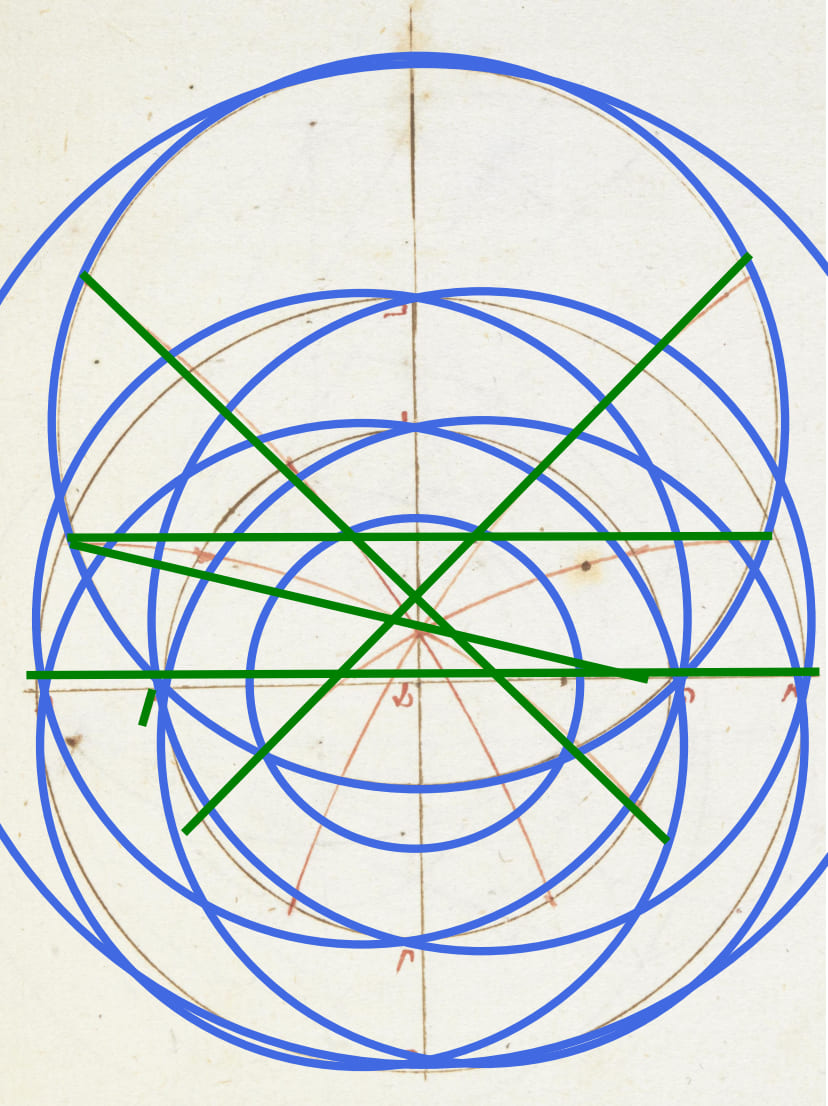} \\

        \includegraphics[width=0.15\linewidth]{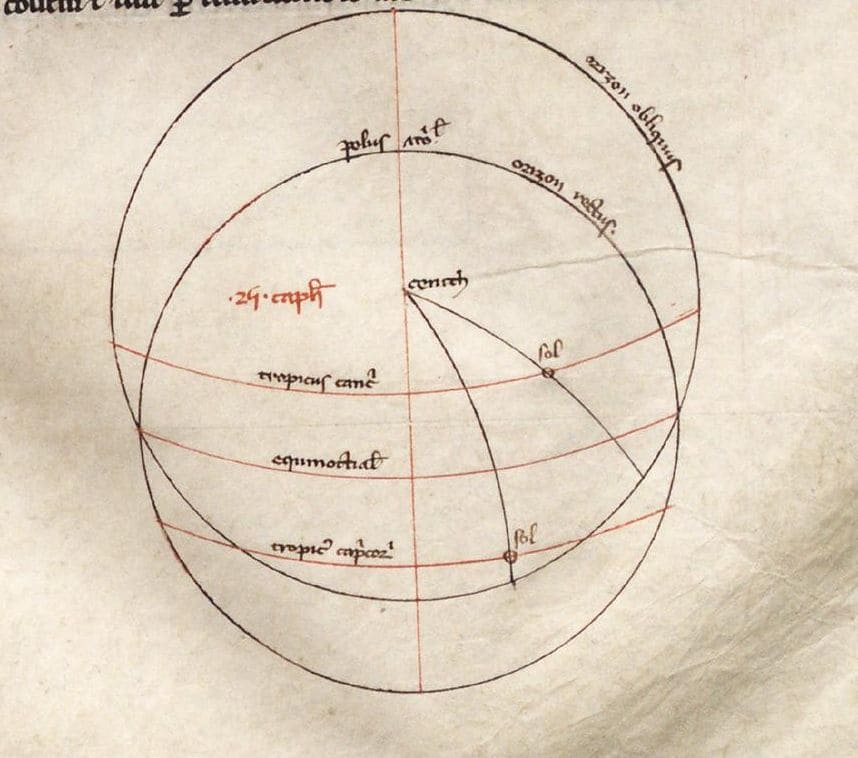} &
        \includegraphics[width=0.15\linewidth]{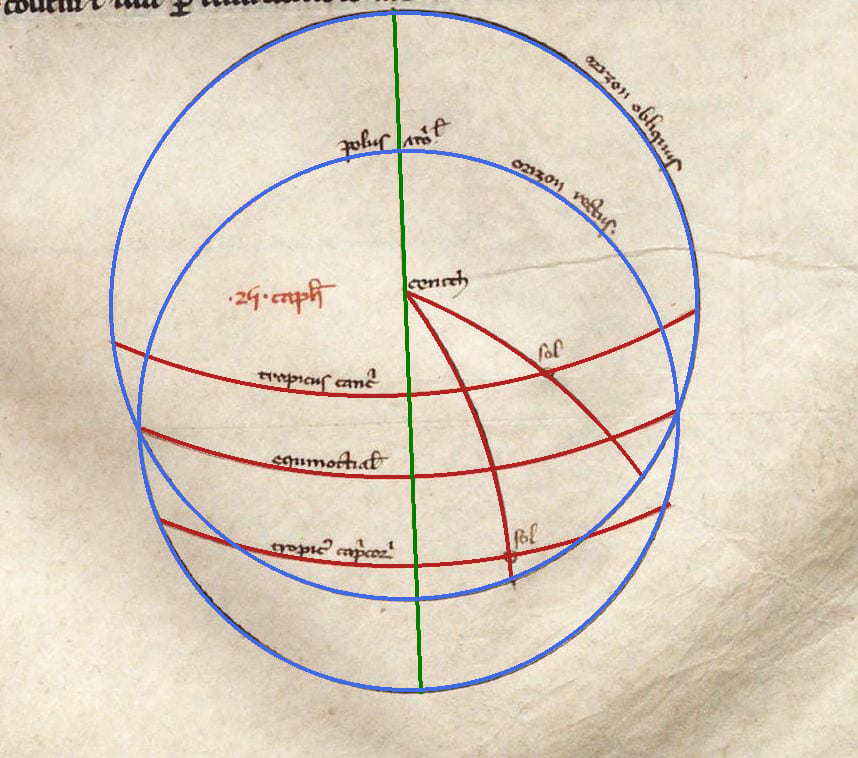} &
        \includegraphics[width=0.15\linewidth]{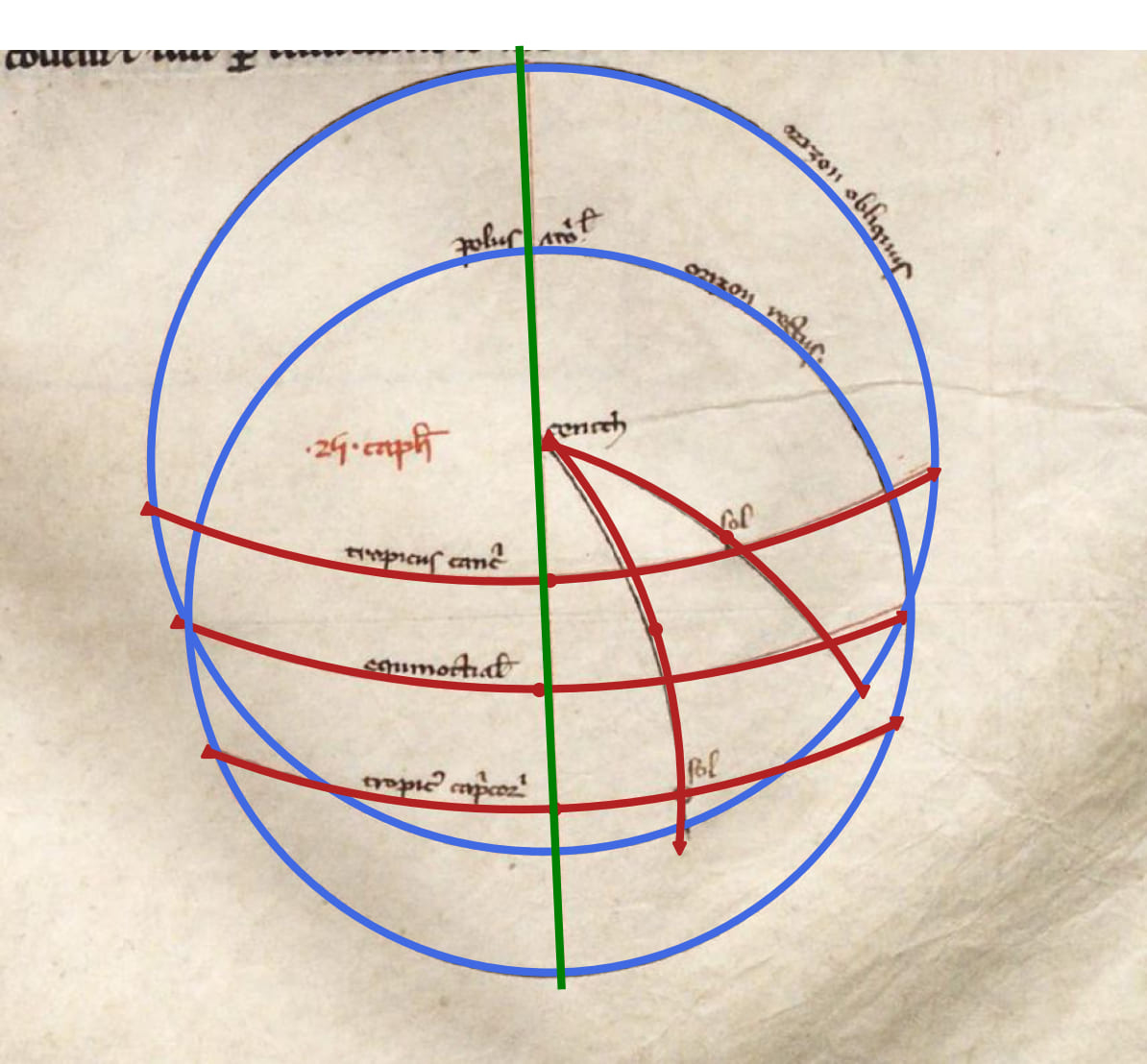} &
        \includegraphics[width=0.15\linewidth]{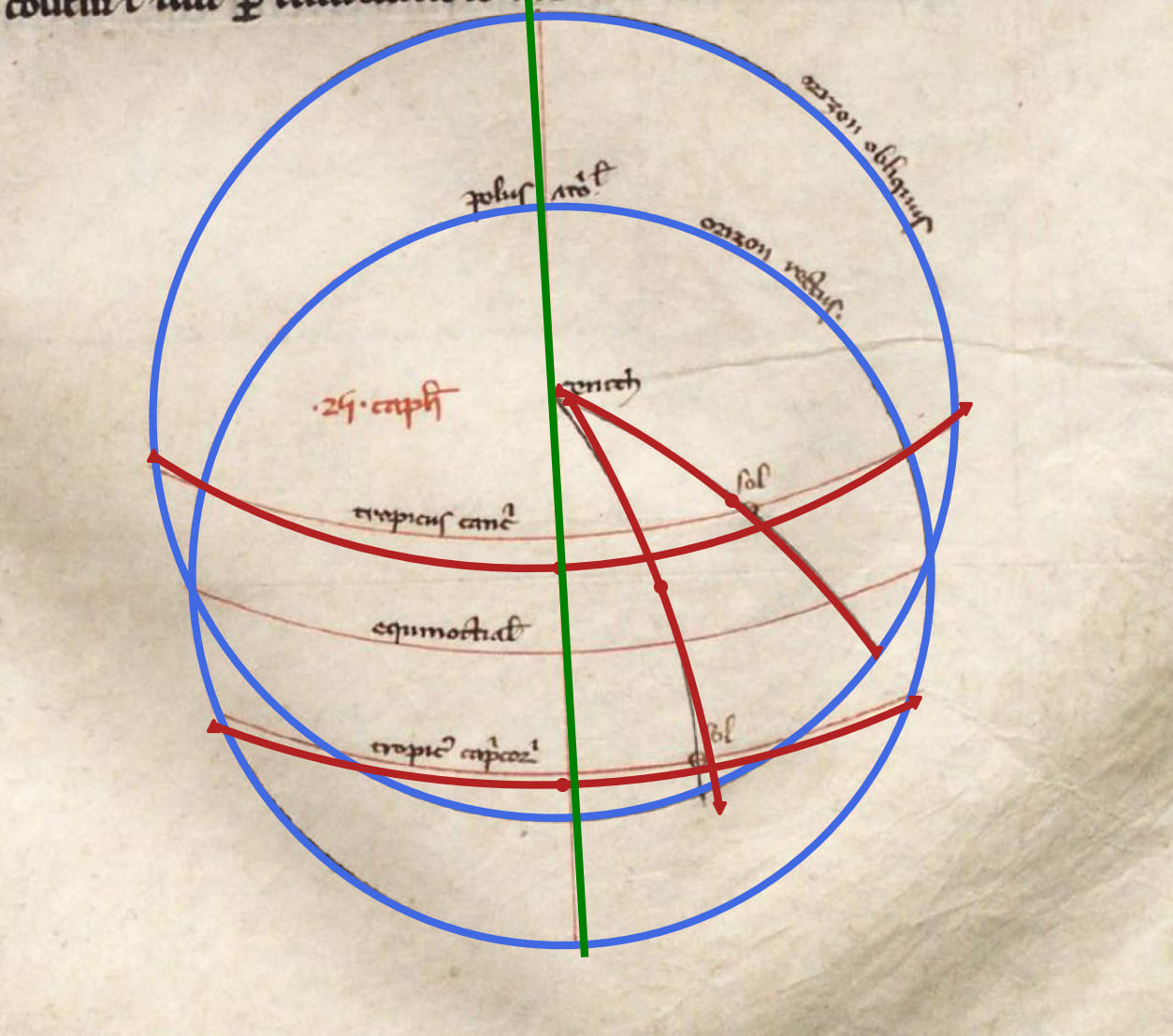} &
        \includegraphics[width=0.15\linewidth]{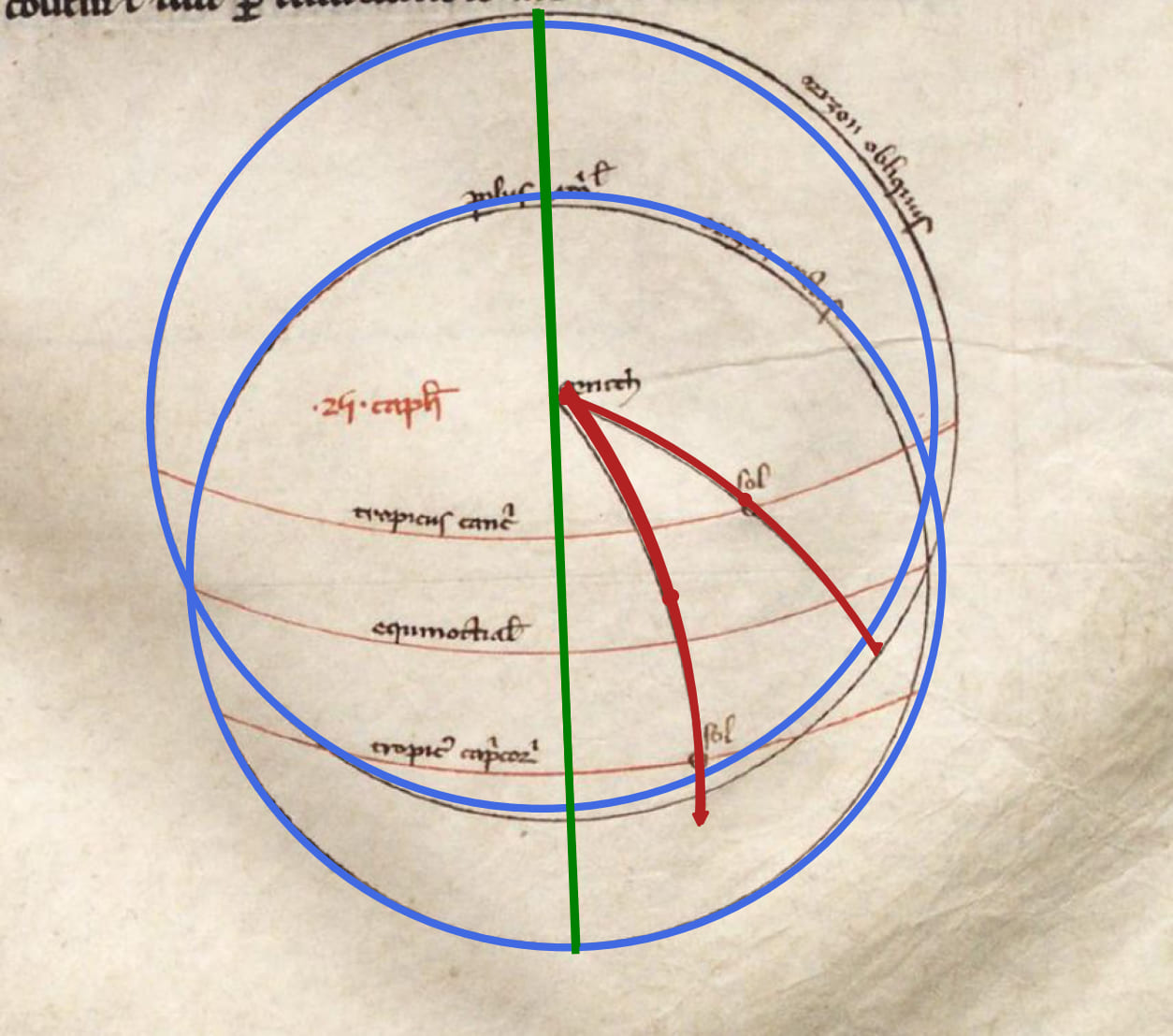} &
        \includegraphics[width=0.15\linewidth]{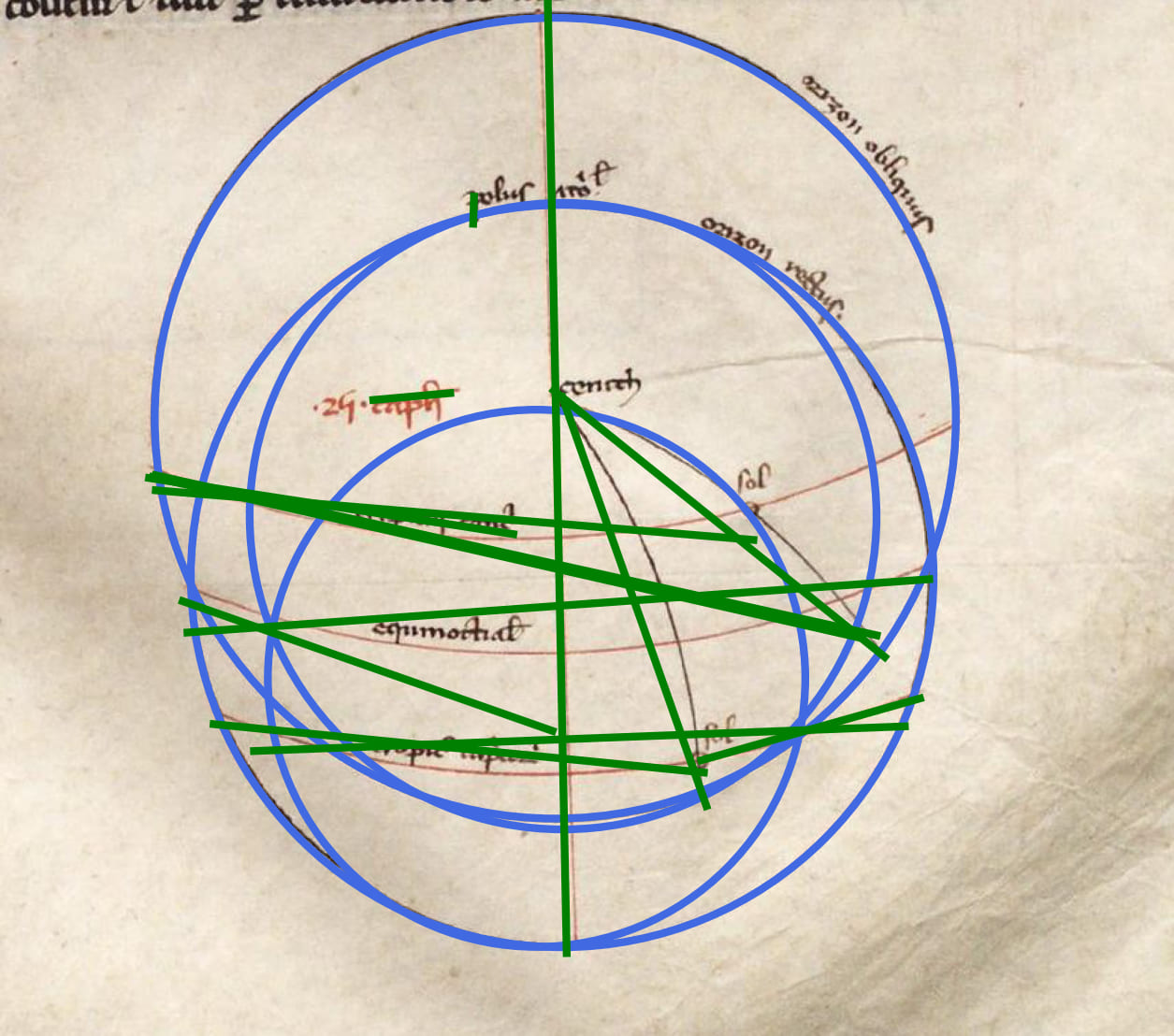} \\
        \rotatebox{90}{\includegraphics[height=0.15\linewidth, width=0.2\linewidth]{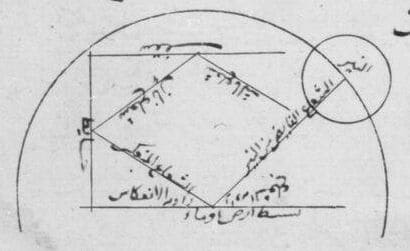}} &
        \rotatebox{90}{\includegraphics[height=0.15\linewidth, width=0.2\linewidth]{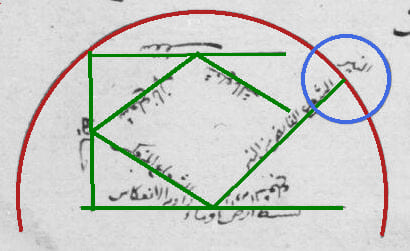}} &
        \rotatebox{90}{\includegraphics[height=0.15\linewidth, width=0.2\linewidth]{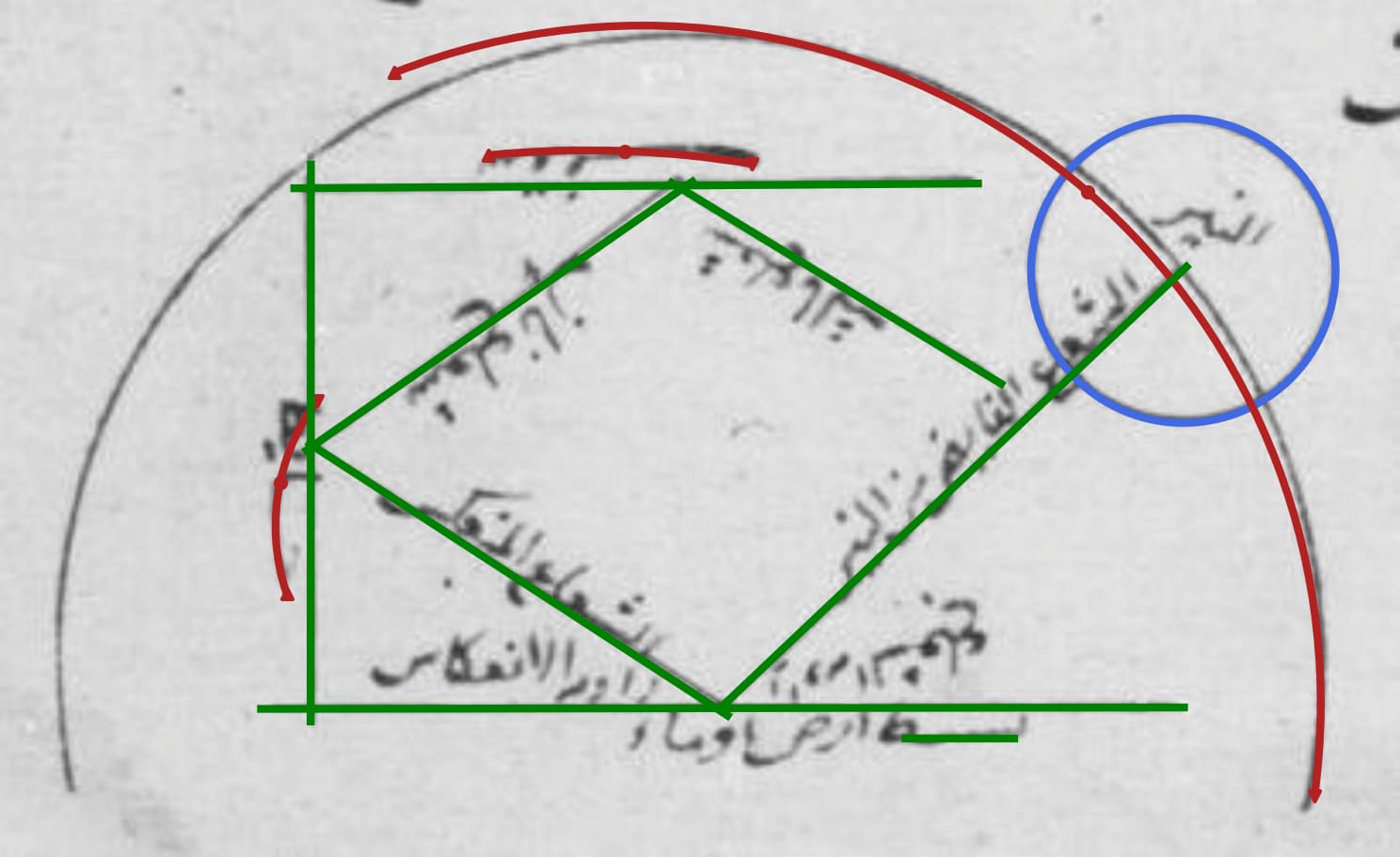}} &
        \rotatebox{90}{\includegraphics[height=0.15\linewidth, width=0.2\linewidth]{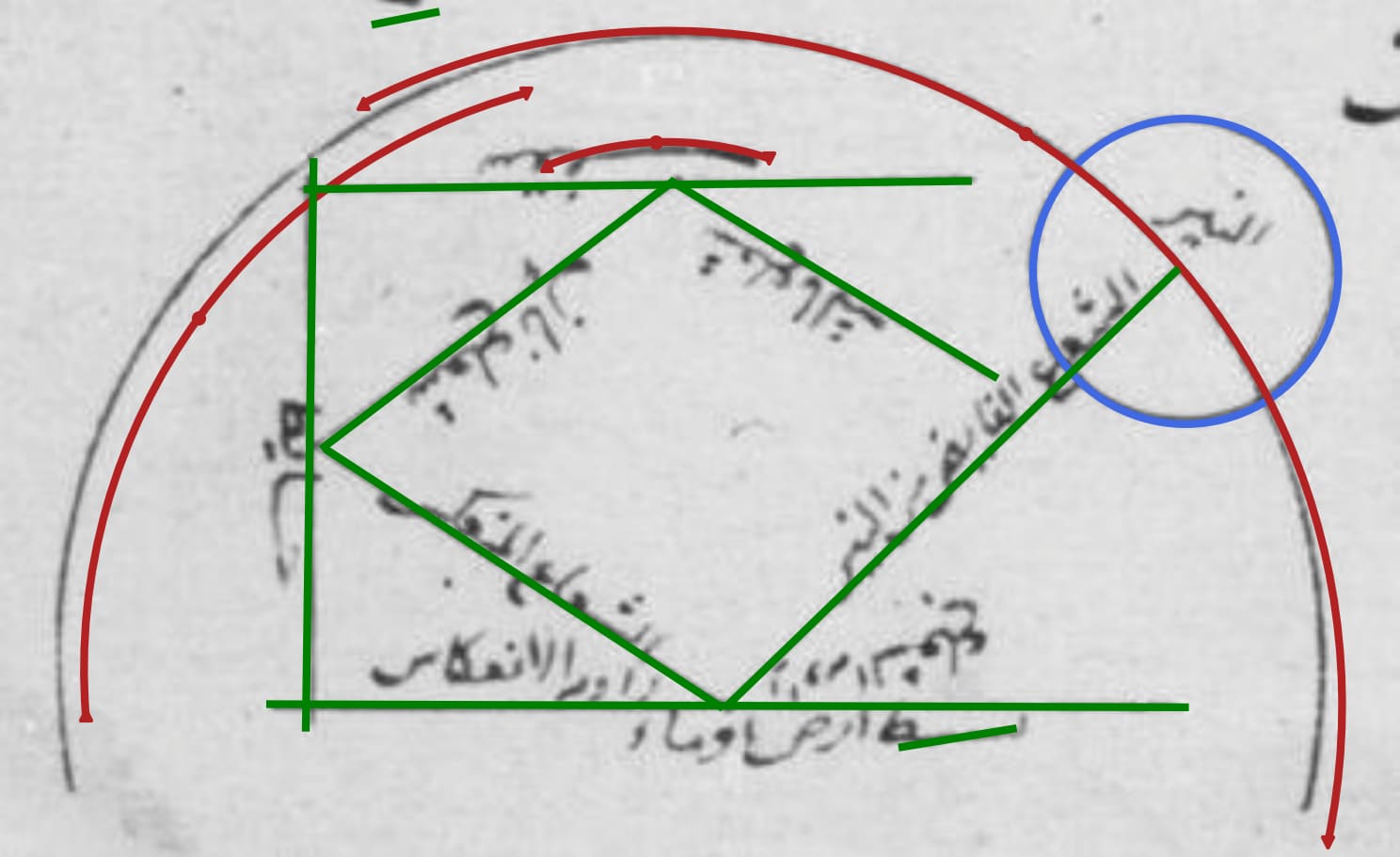}} &
        \rotatebox{90}{\includegraphics[height=0.15\linewidth, width=0.2\linewidth]{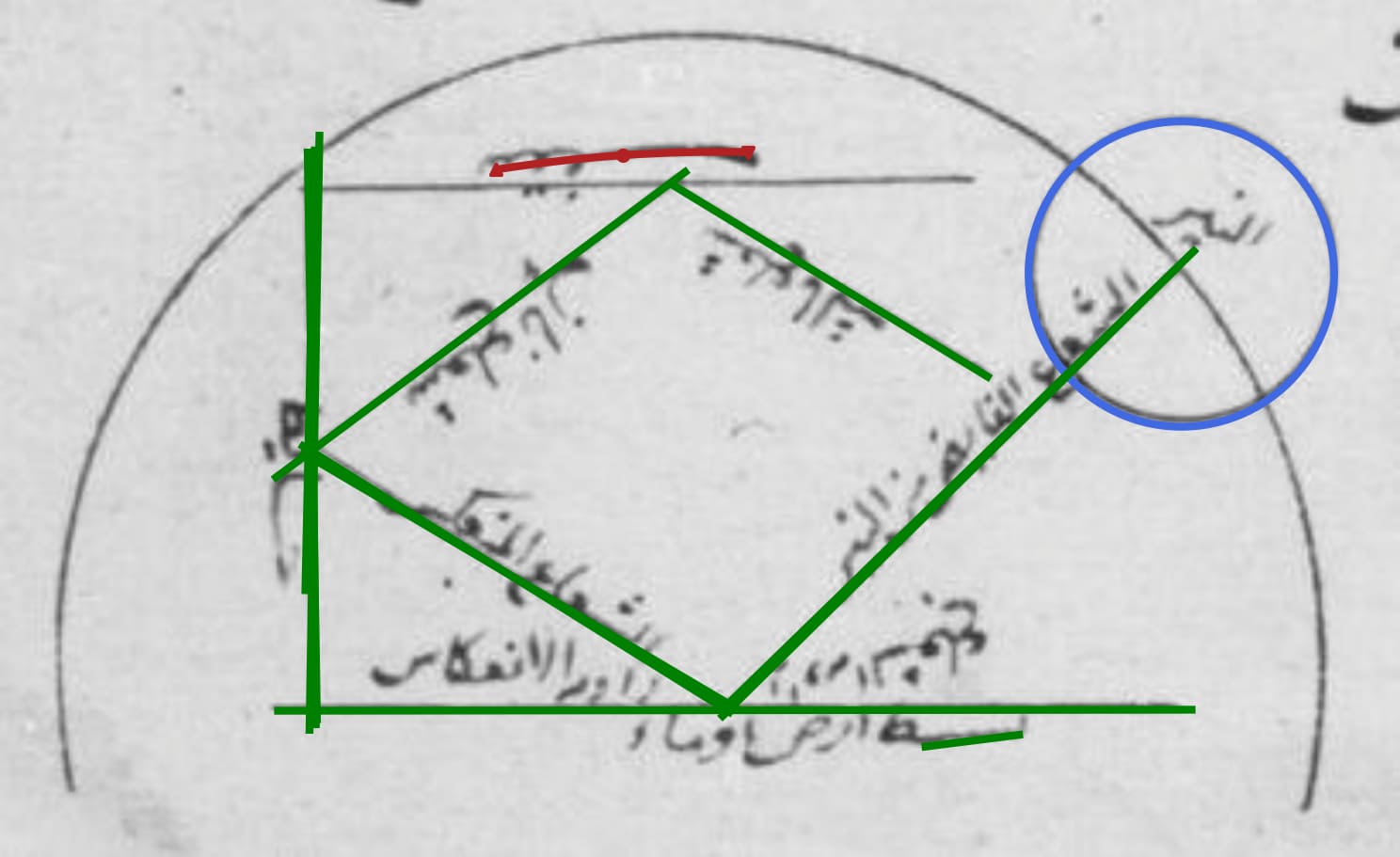}} &
        \rotatebox{90}{\includegraphics[height=0.15\linewidth, width=0.2\linewidth]{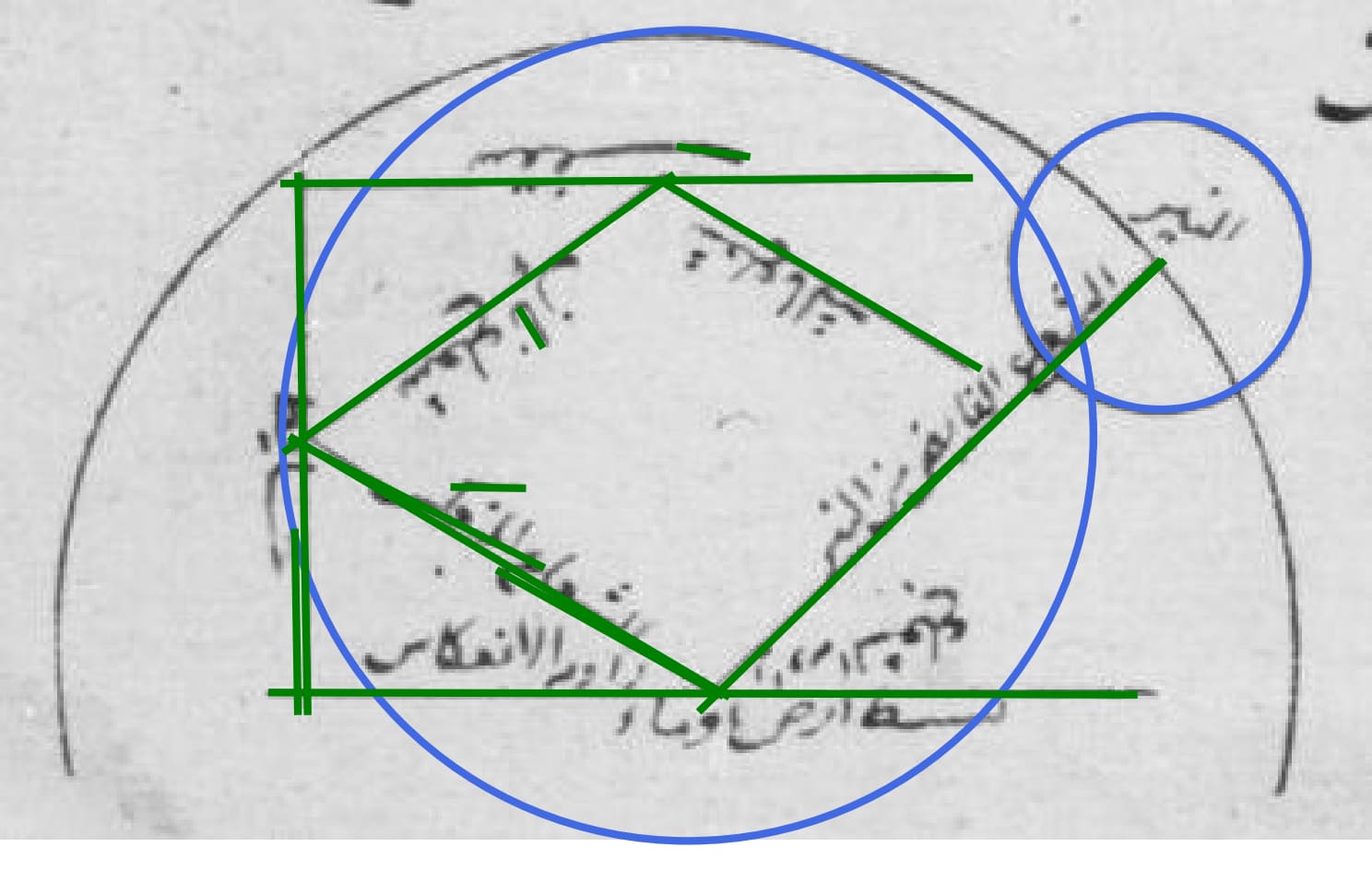}} \\

        \includegraphics[width=0.15\linewidth]{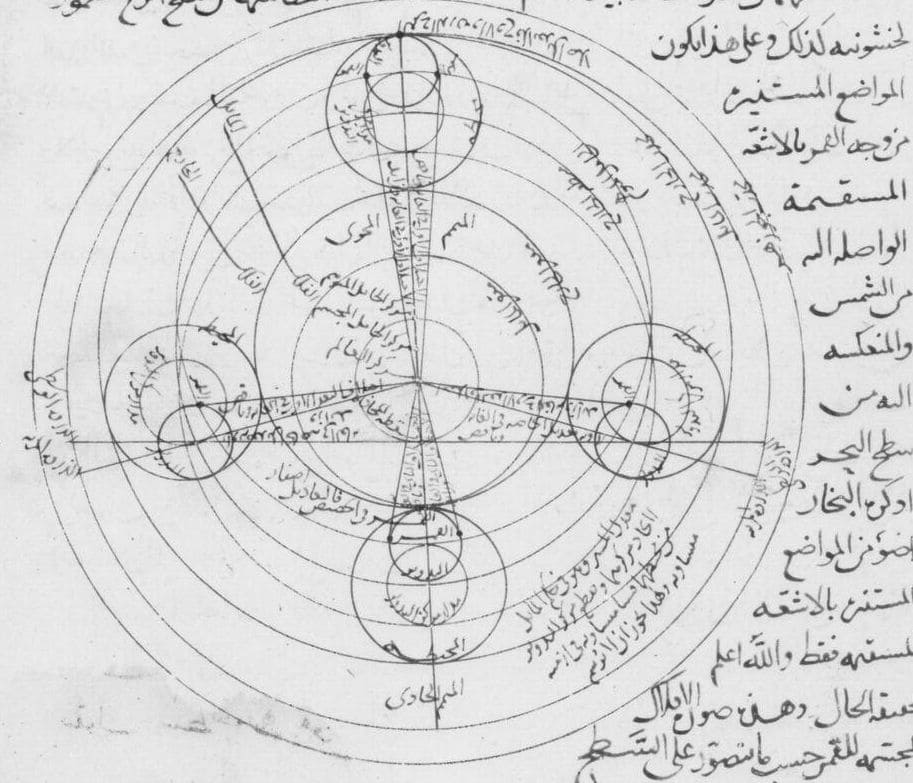} &
        \includegraphics[width=0.15\linewidth]{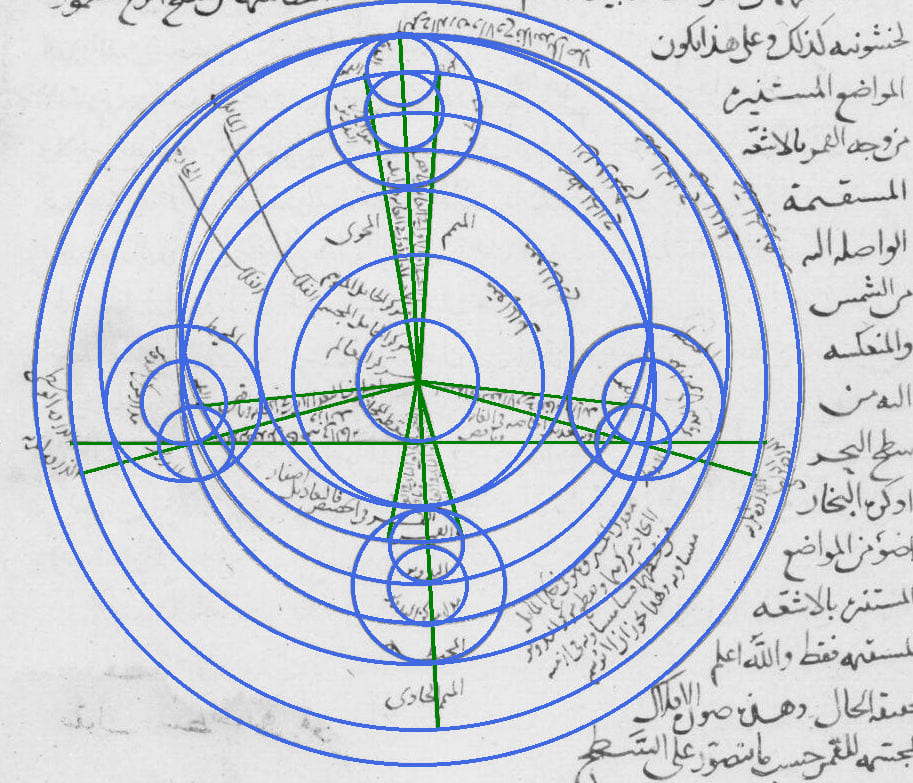} &
        \includegraphics[width=0.15\linewidth]{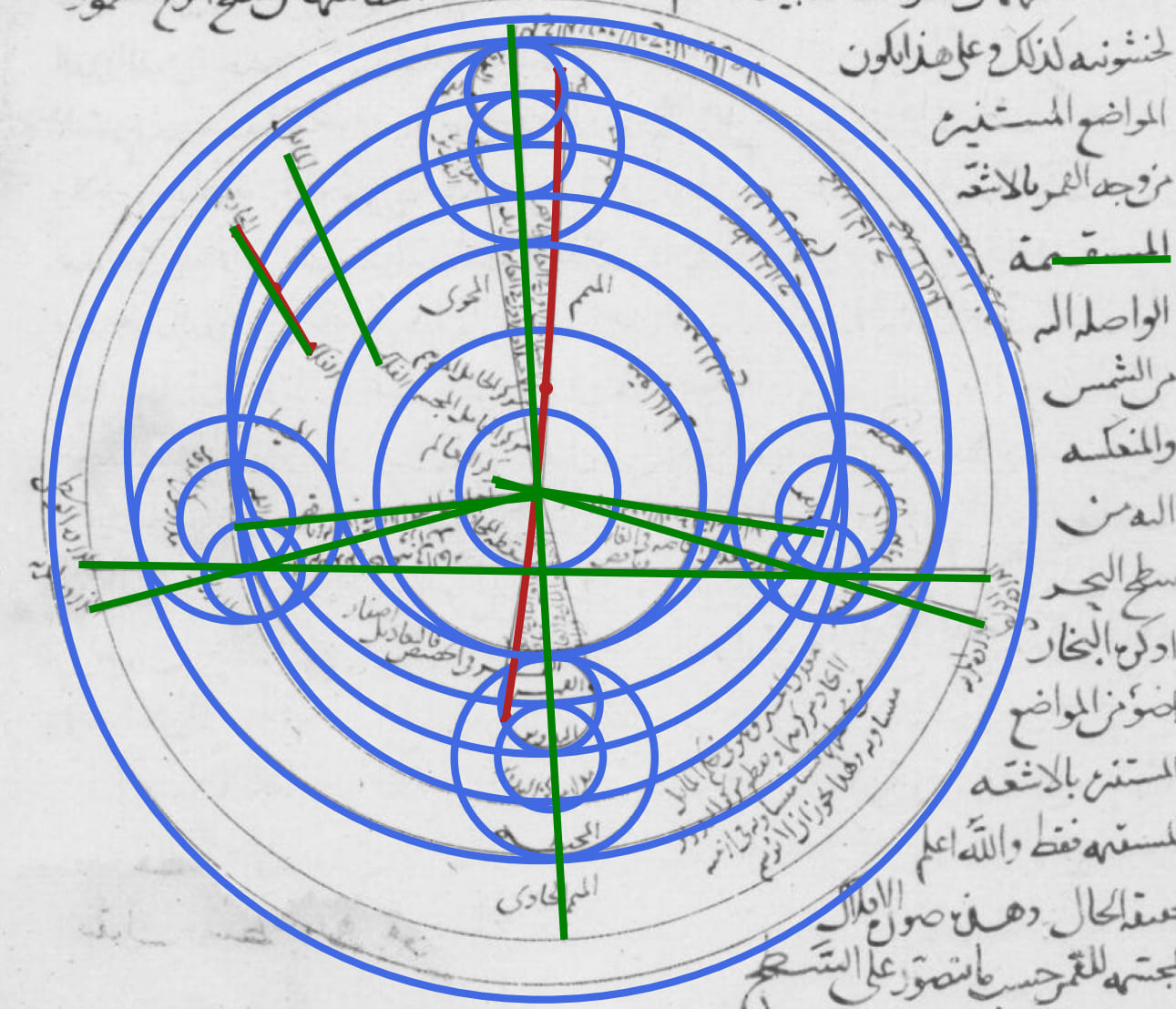} &
        \includegraphics[width=0.15\linewidth]{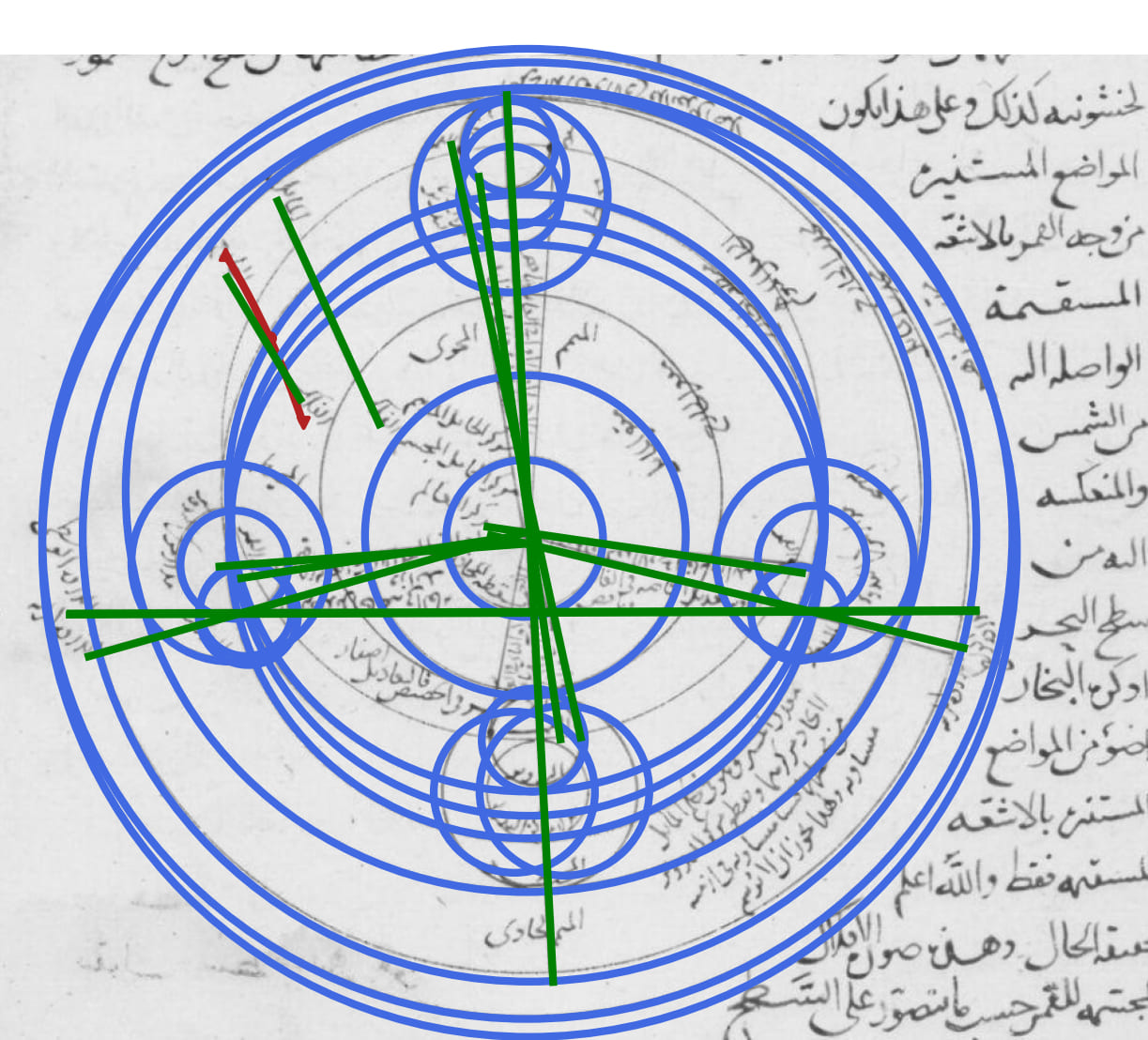} &
        \includegraphics[width=0.15\linewidth]{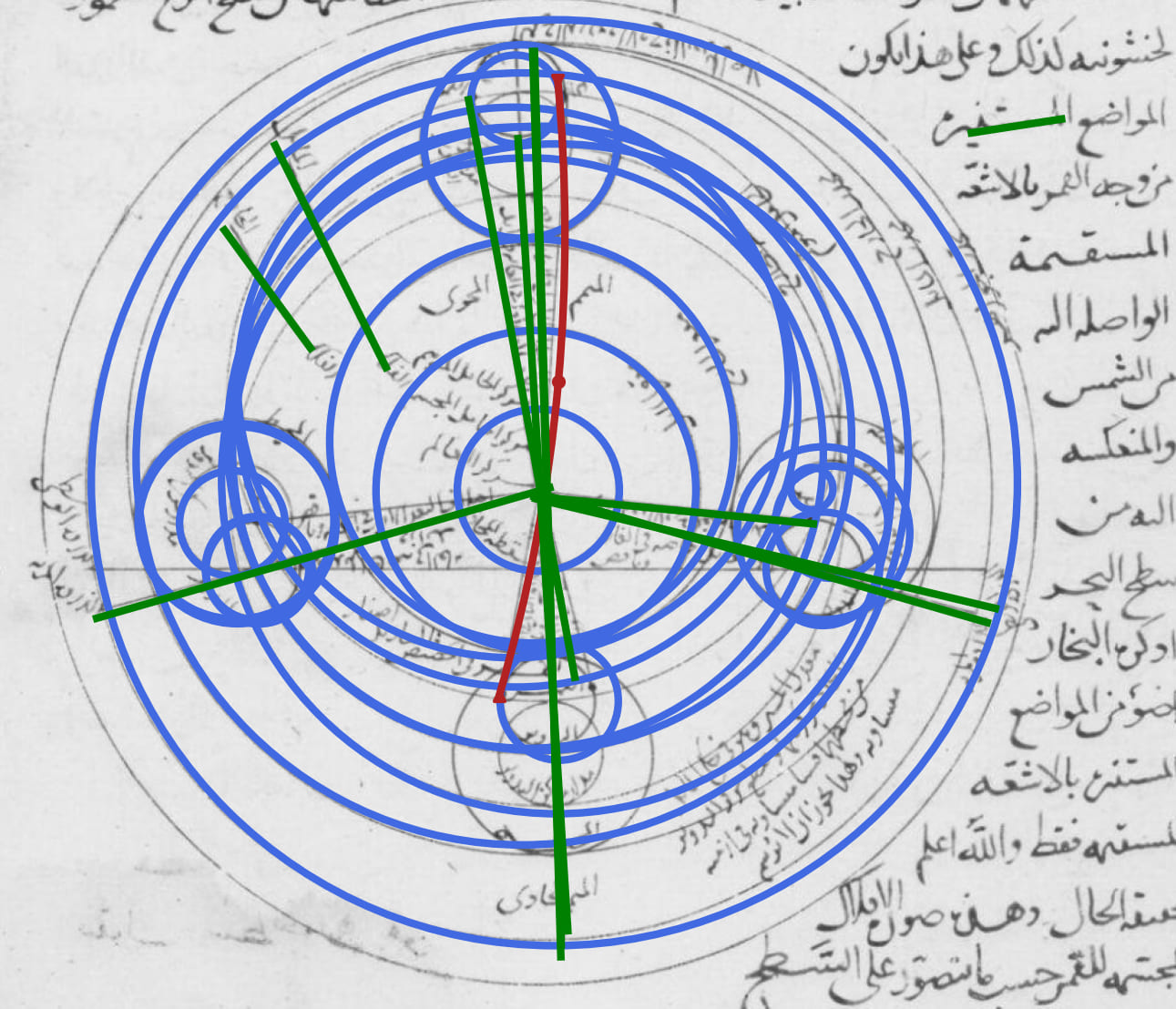} &
        \includegraphics[width=0.15\linewidth]{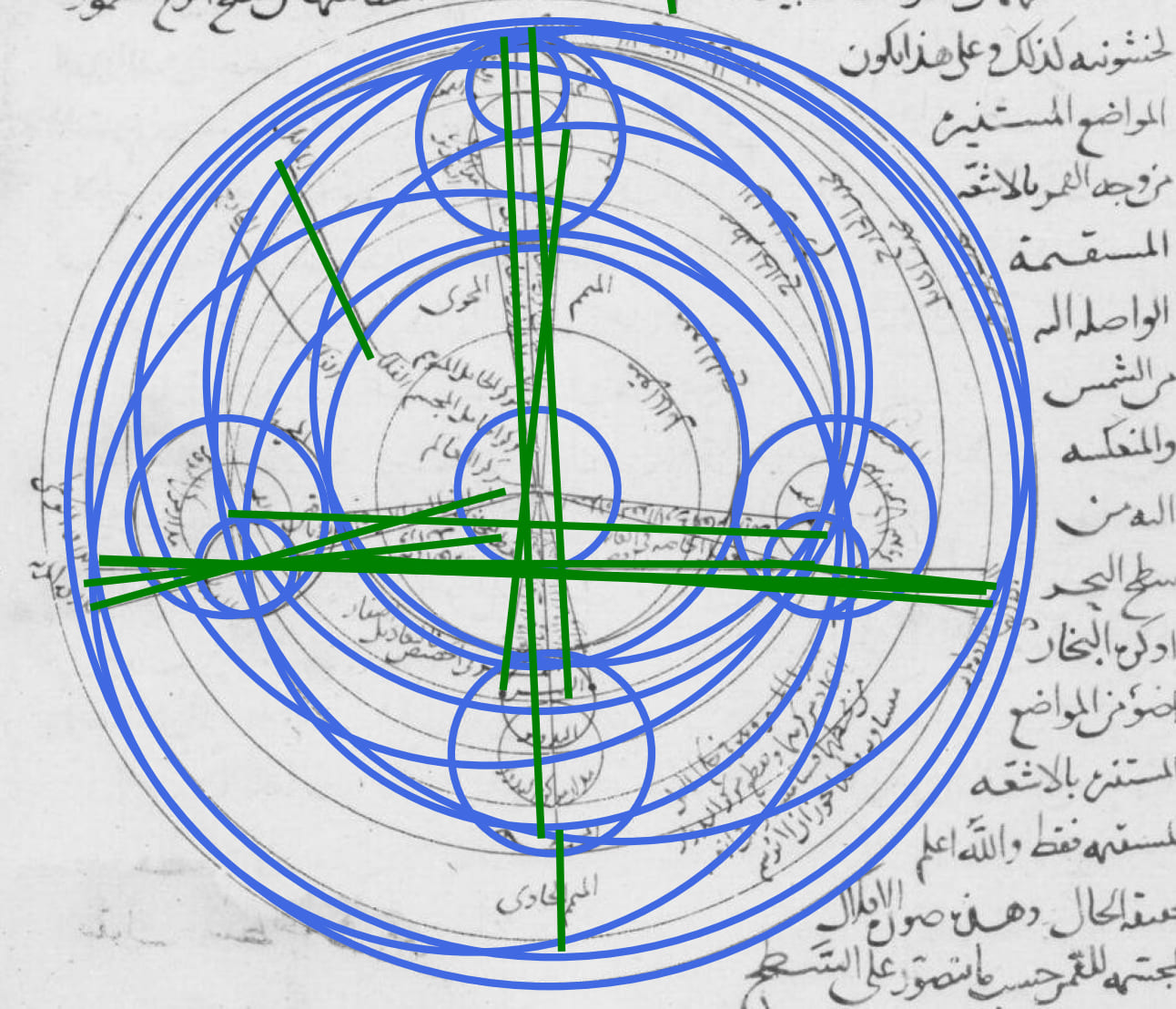} \\
        
        \includegraphics[width=0.15\linewidth]{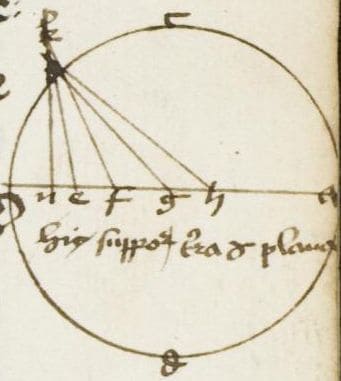} &
        \includegraphics[width=0.15\linewidth]{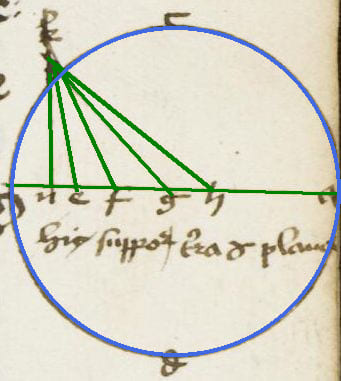} &
        \includegraphics[width=0.15\linewidth]{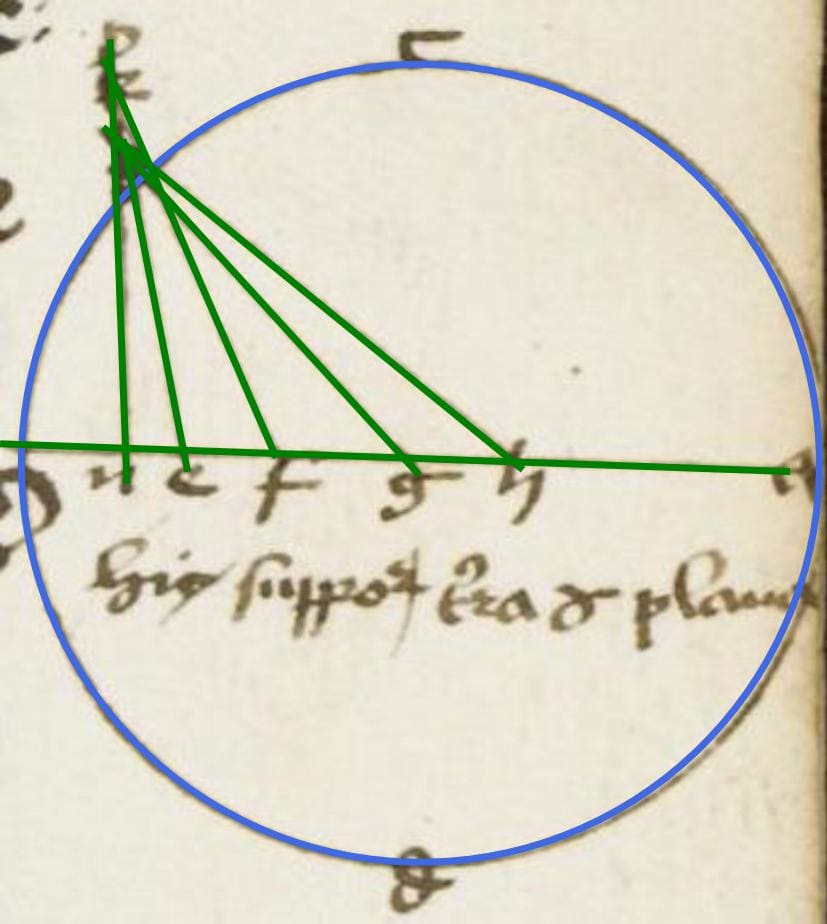} &
        \includegraphics[width=0.15\linewidth]{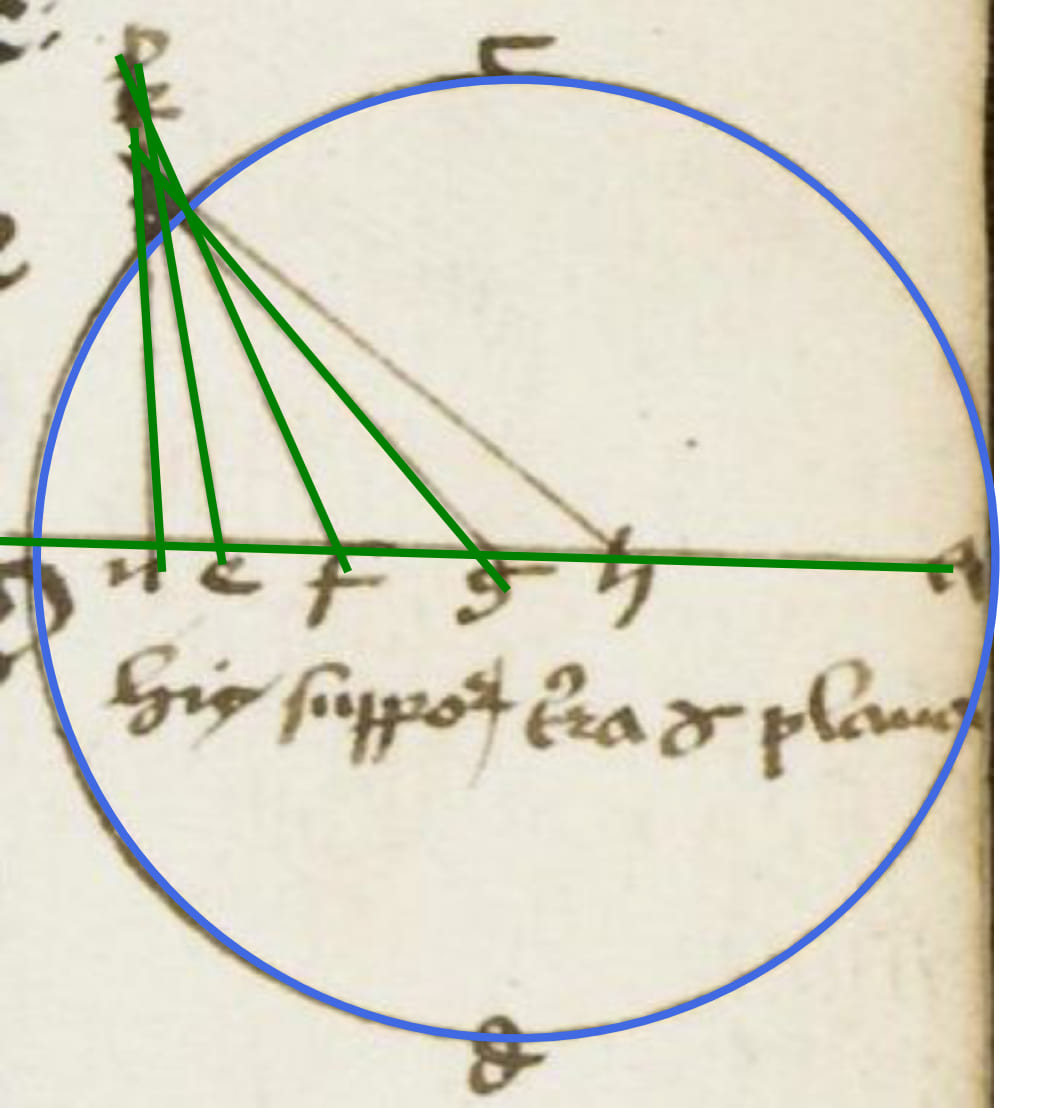} &
        \includegraphics[width=0.15\linewidth]{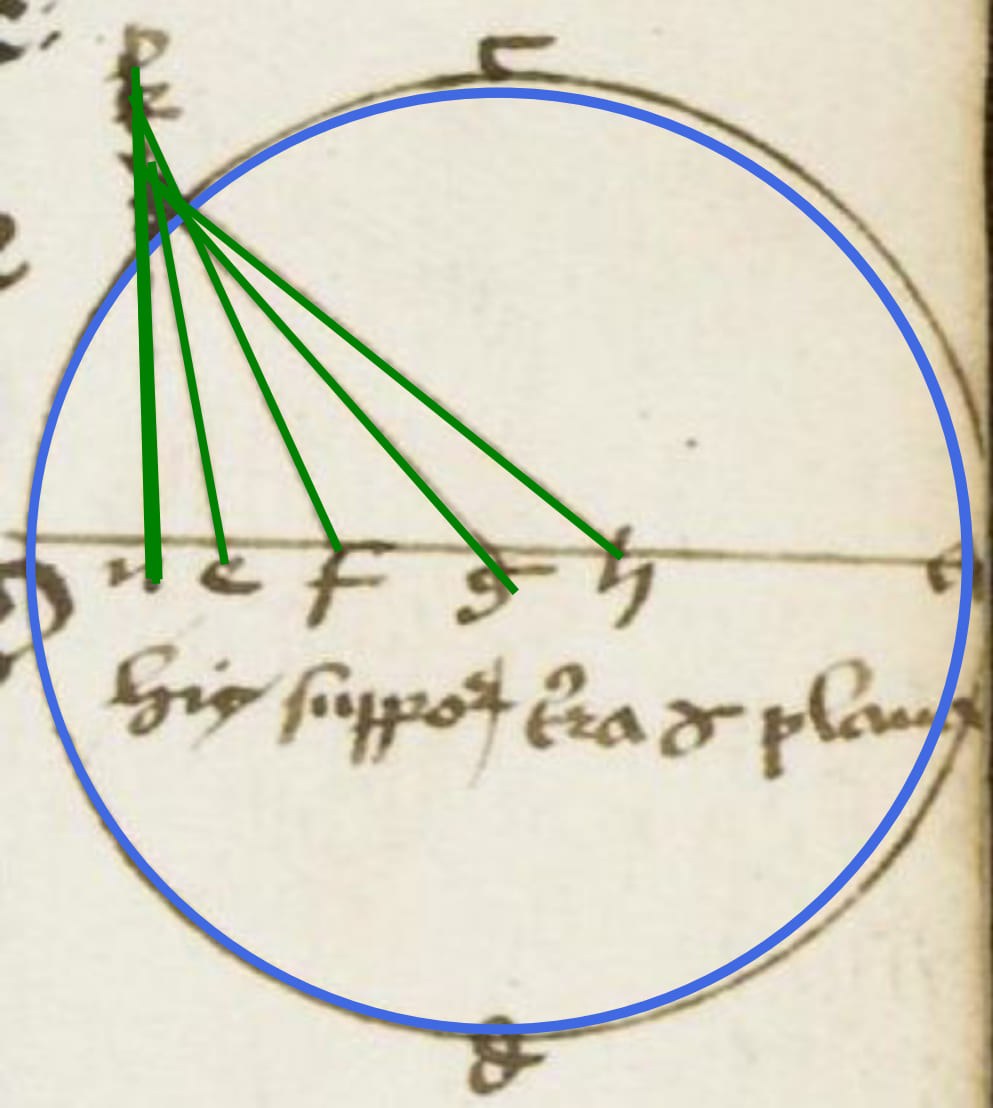} &
        \includegraphics[width=0.15\linewidth]{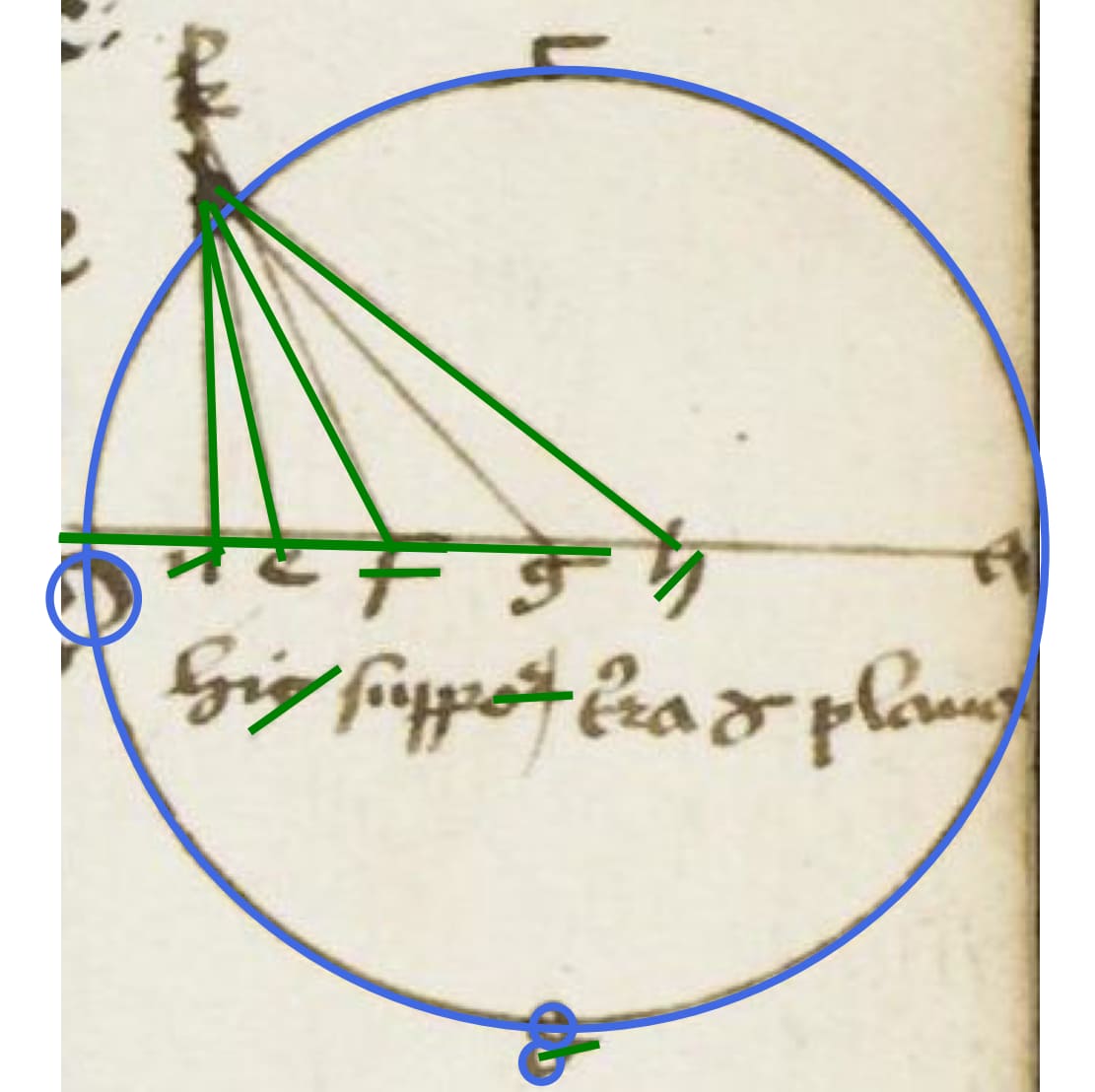} \\

        \includegraphics[width=0.15\linewidth]{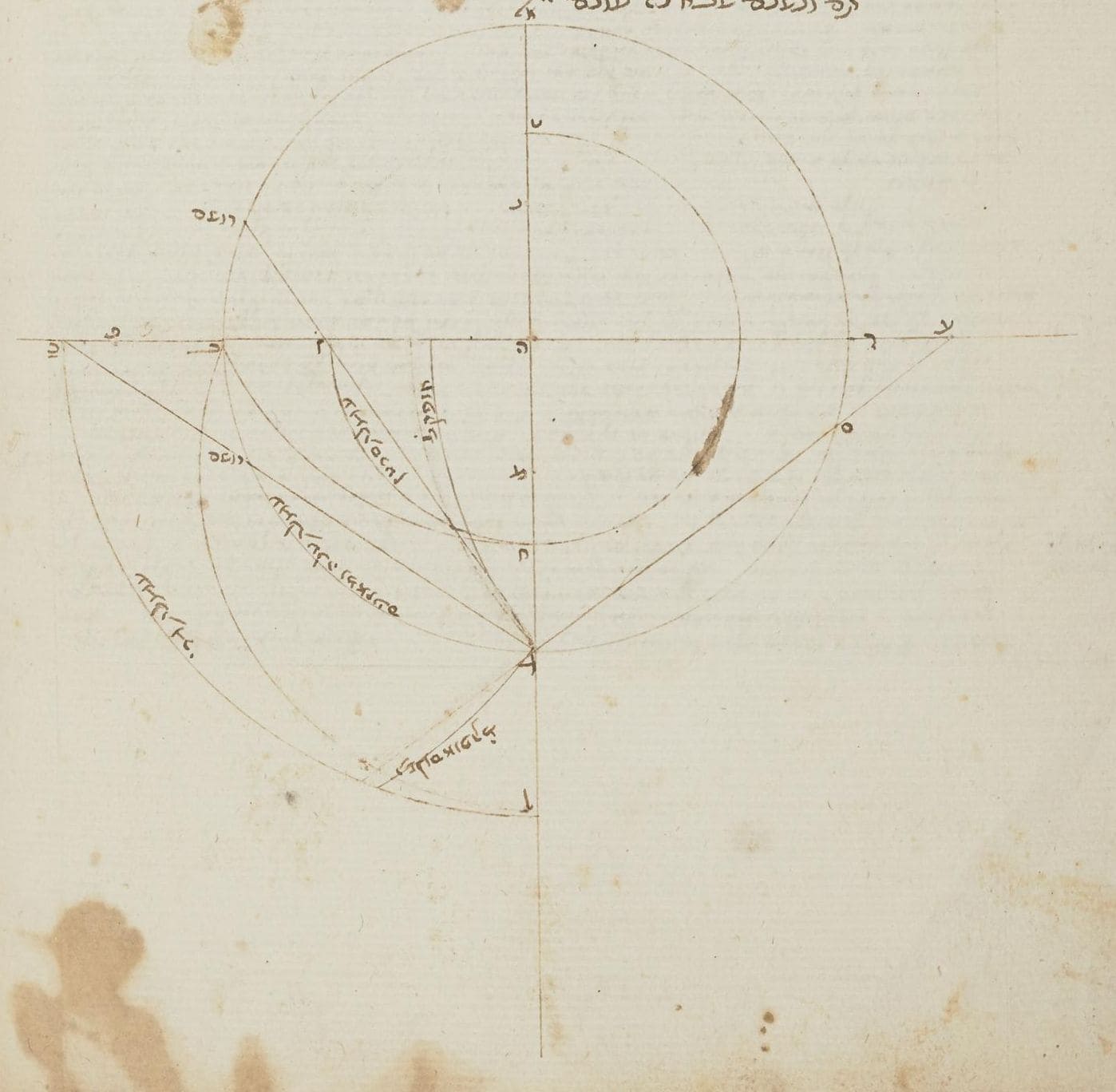} &
        \includegraphics[width=0.15\linewidth]{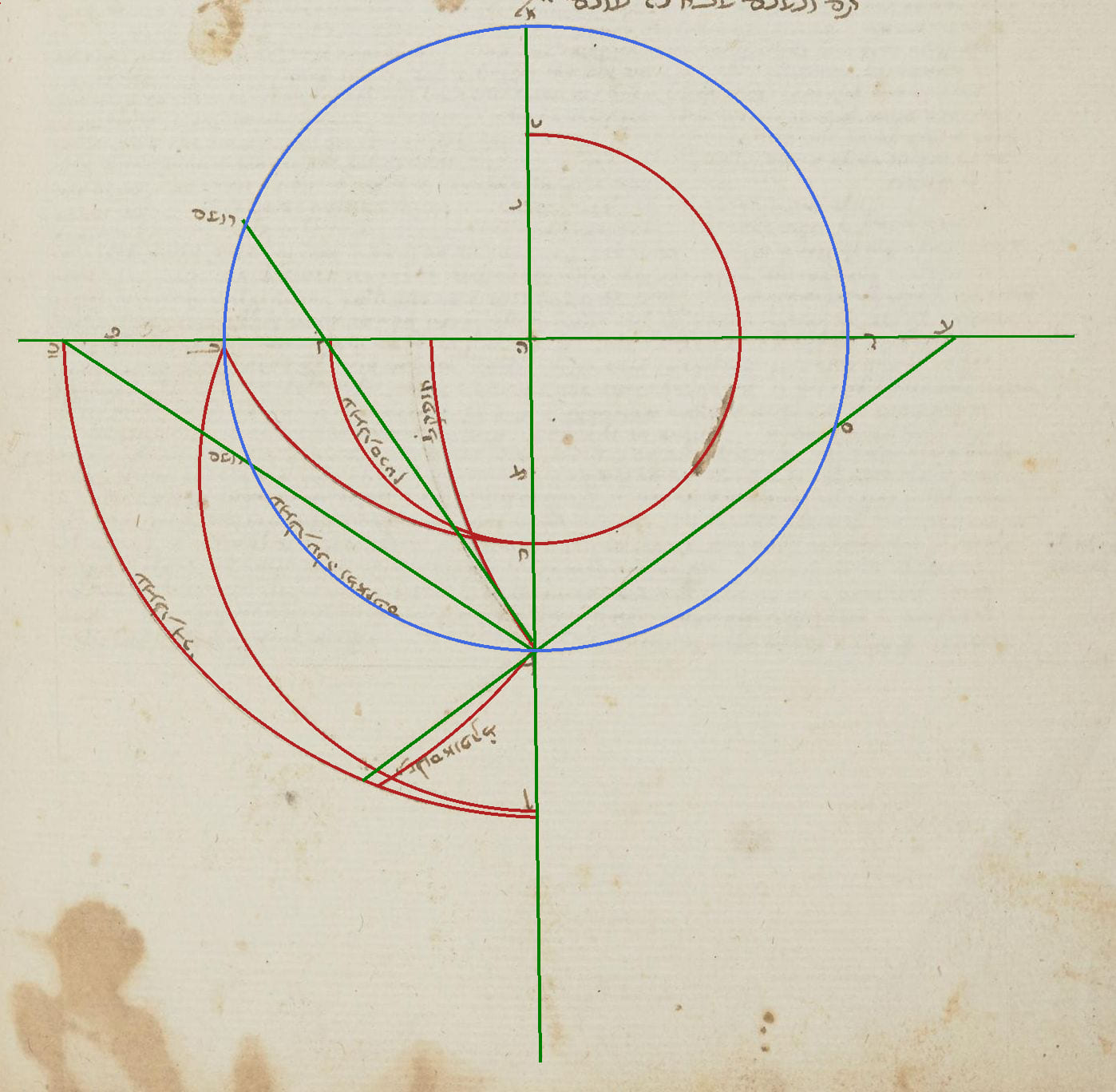} &
        \includegraphics[width=0.15\linewidth]{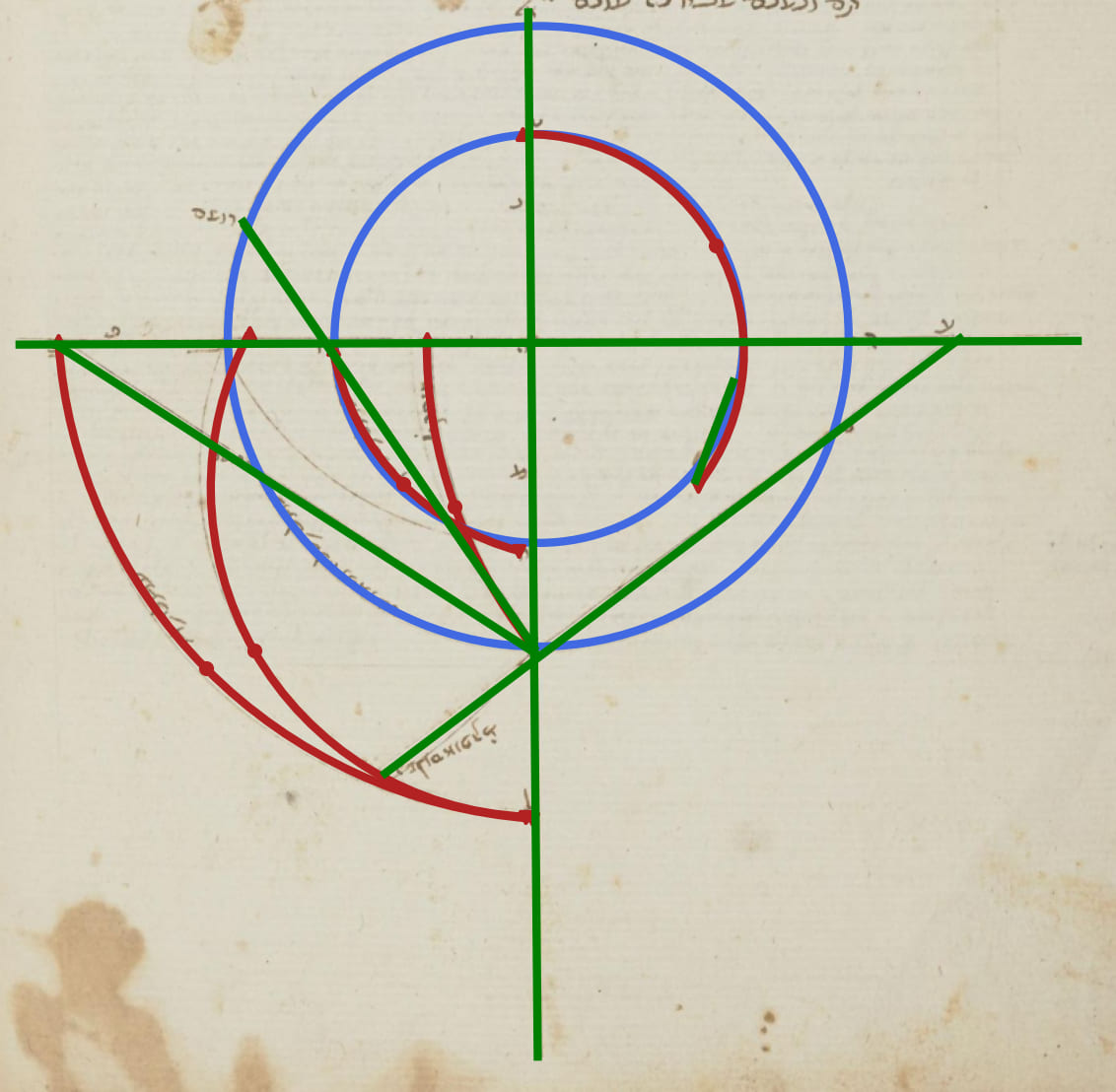} &
        \includegraphics[width=0.15\linewidth]{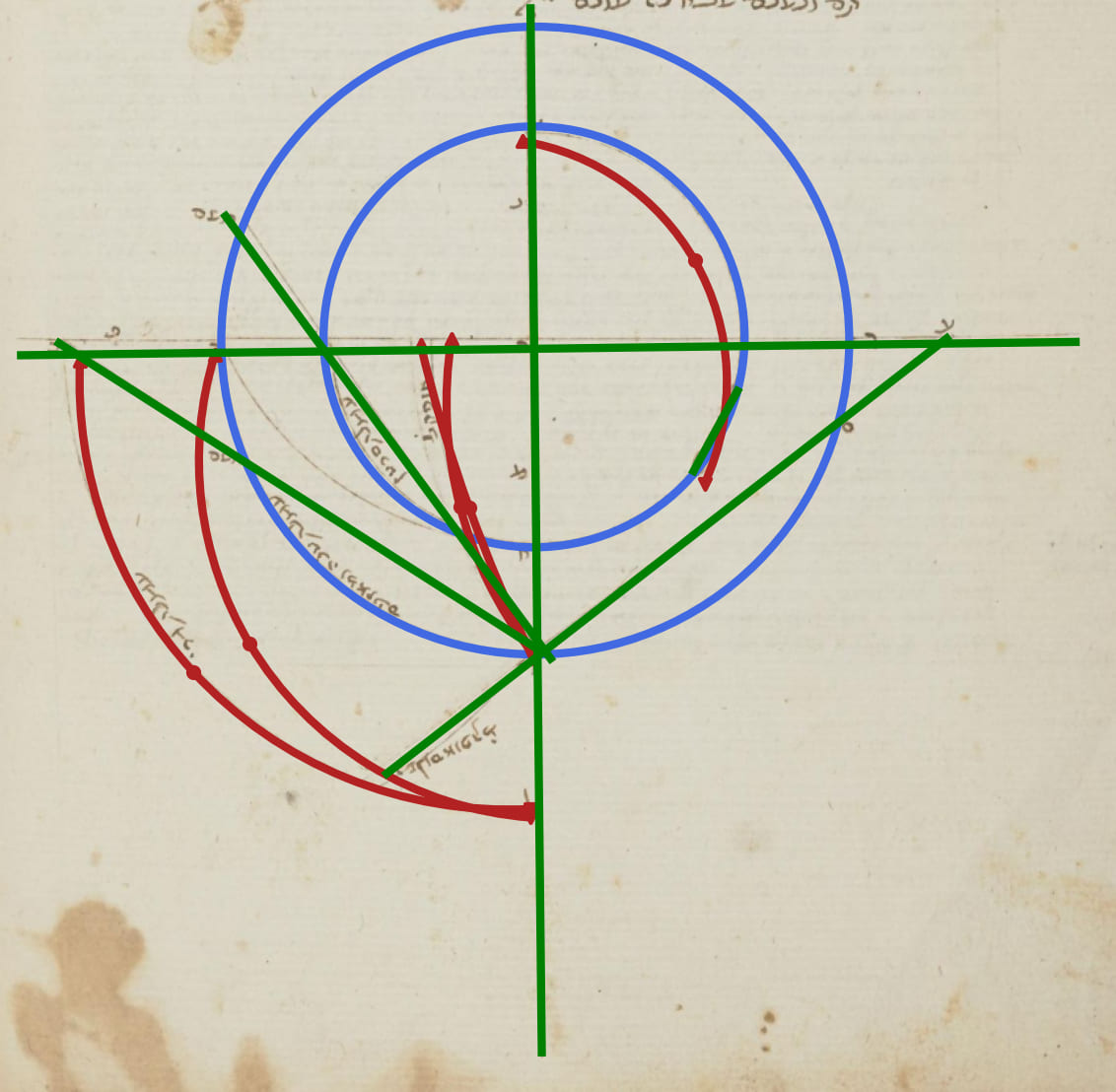} &
        \includegraphics[width=0.15\linewidth]{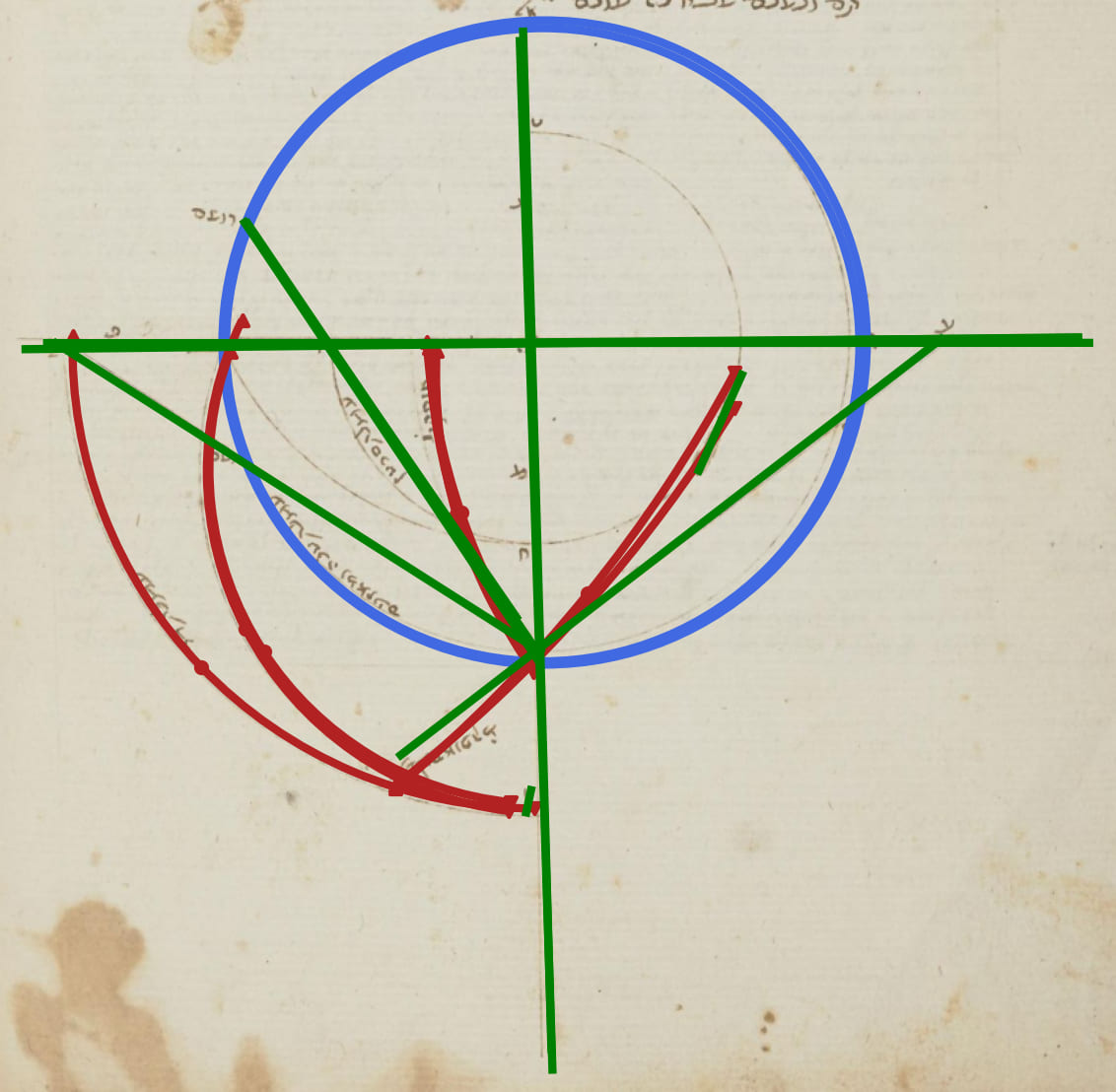} &
        \includegraphics[width=0.15\linewidth]{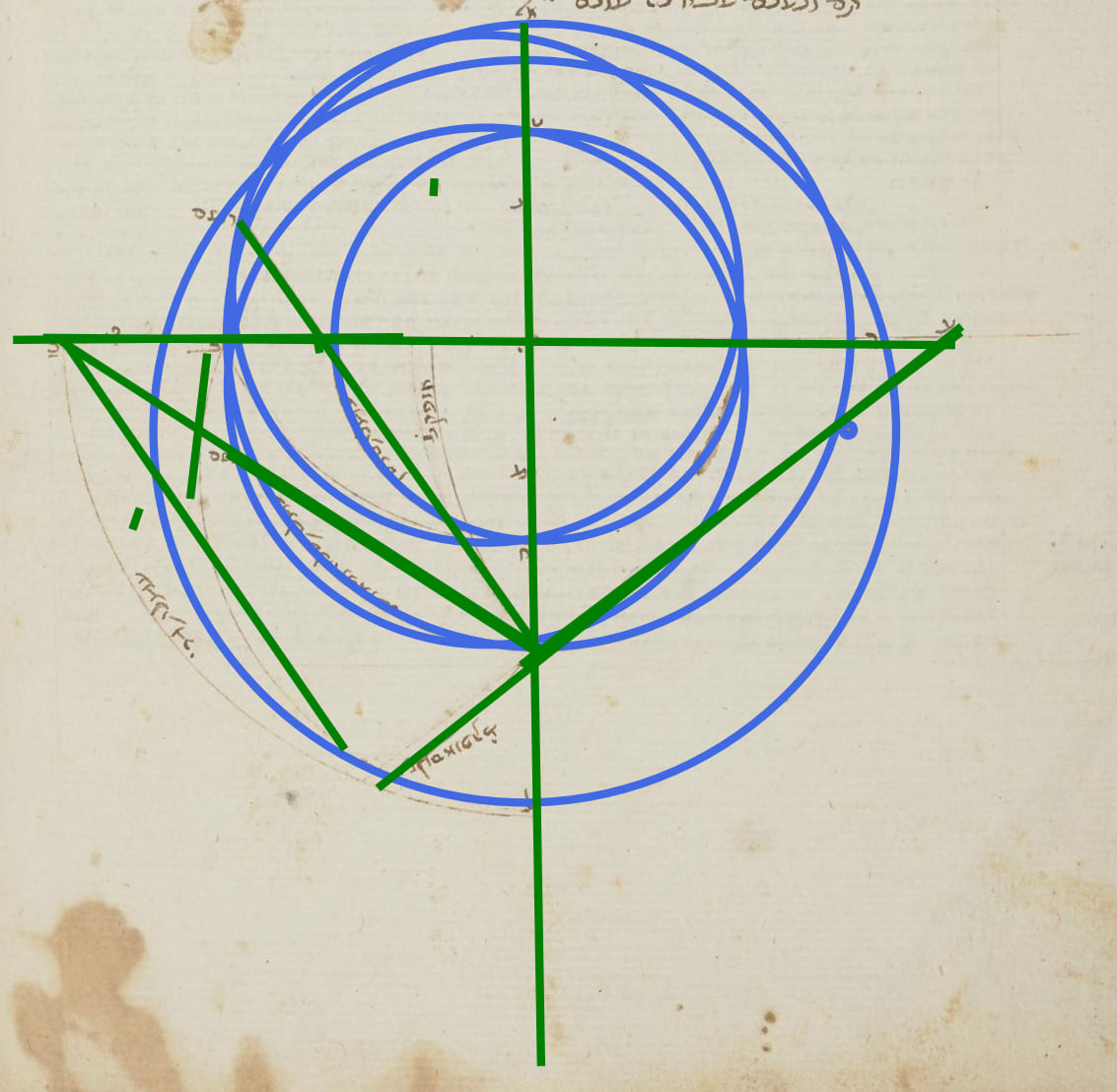} \\
        
        \includegraphics[width=0.15\linewidth]{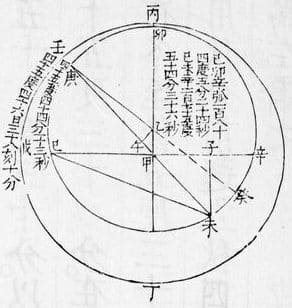} &
        \includegraphics[width=0.15\linewidth]{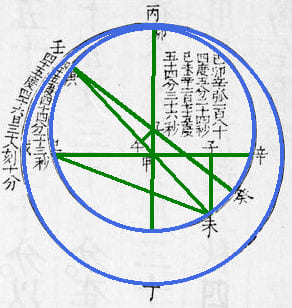} &
        \includegraphics[width=0.15\linewidth]{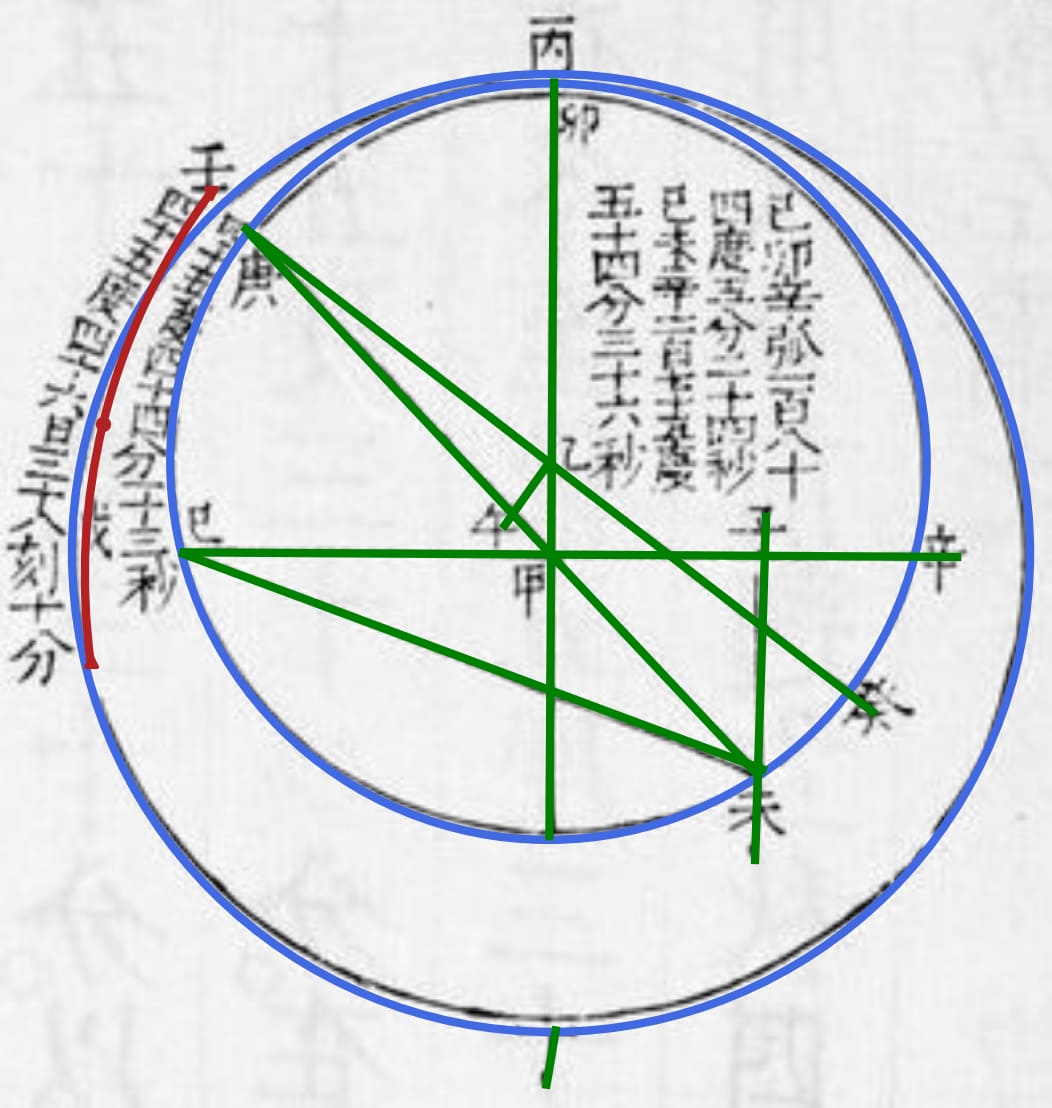} &
        \includegraphics[width=0.15\linewidth]{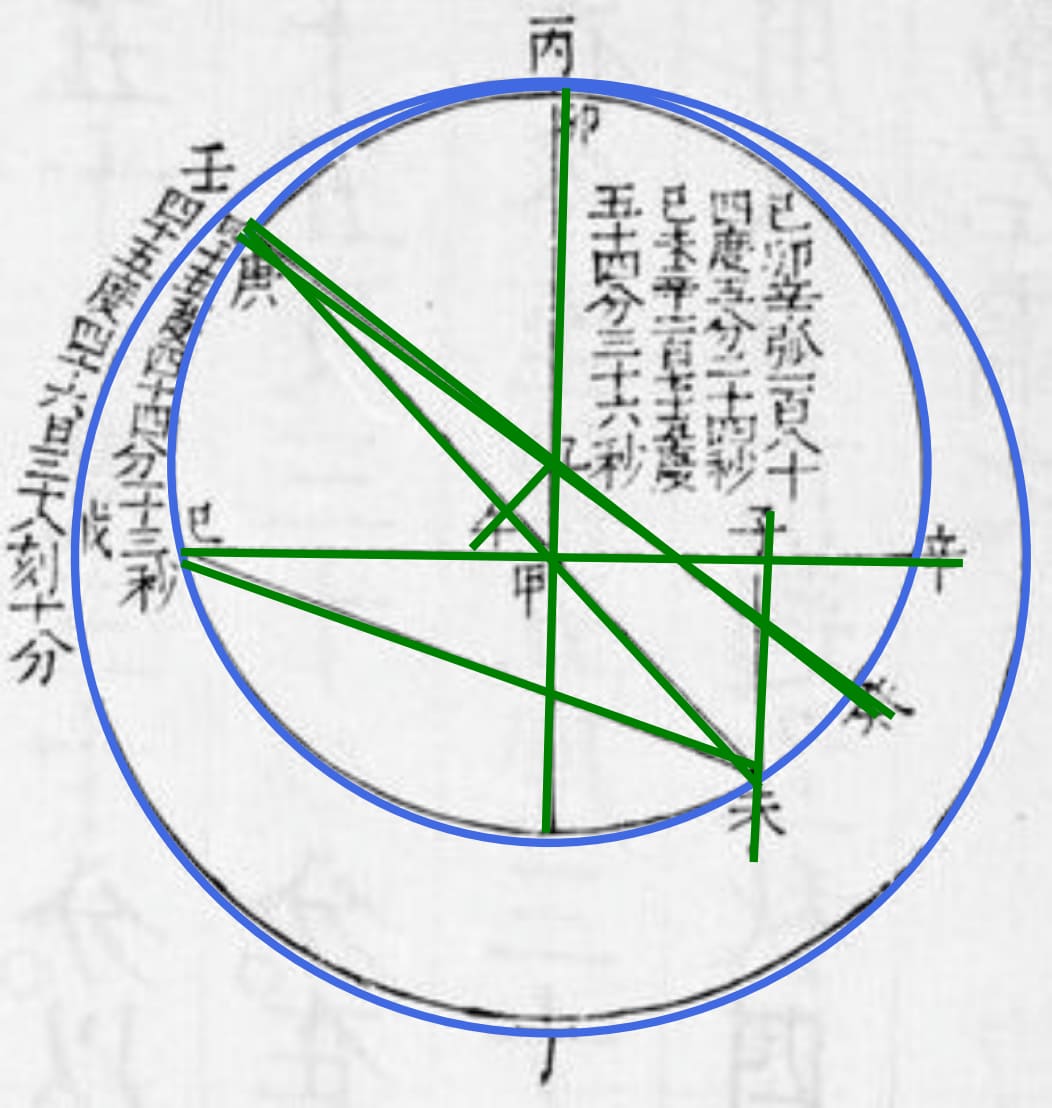} &
        \includegraphics[width=0.15\linewidth]{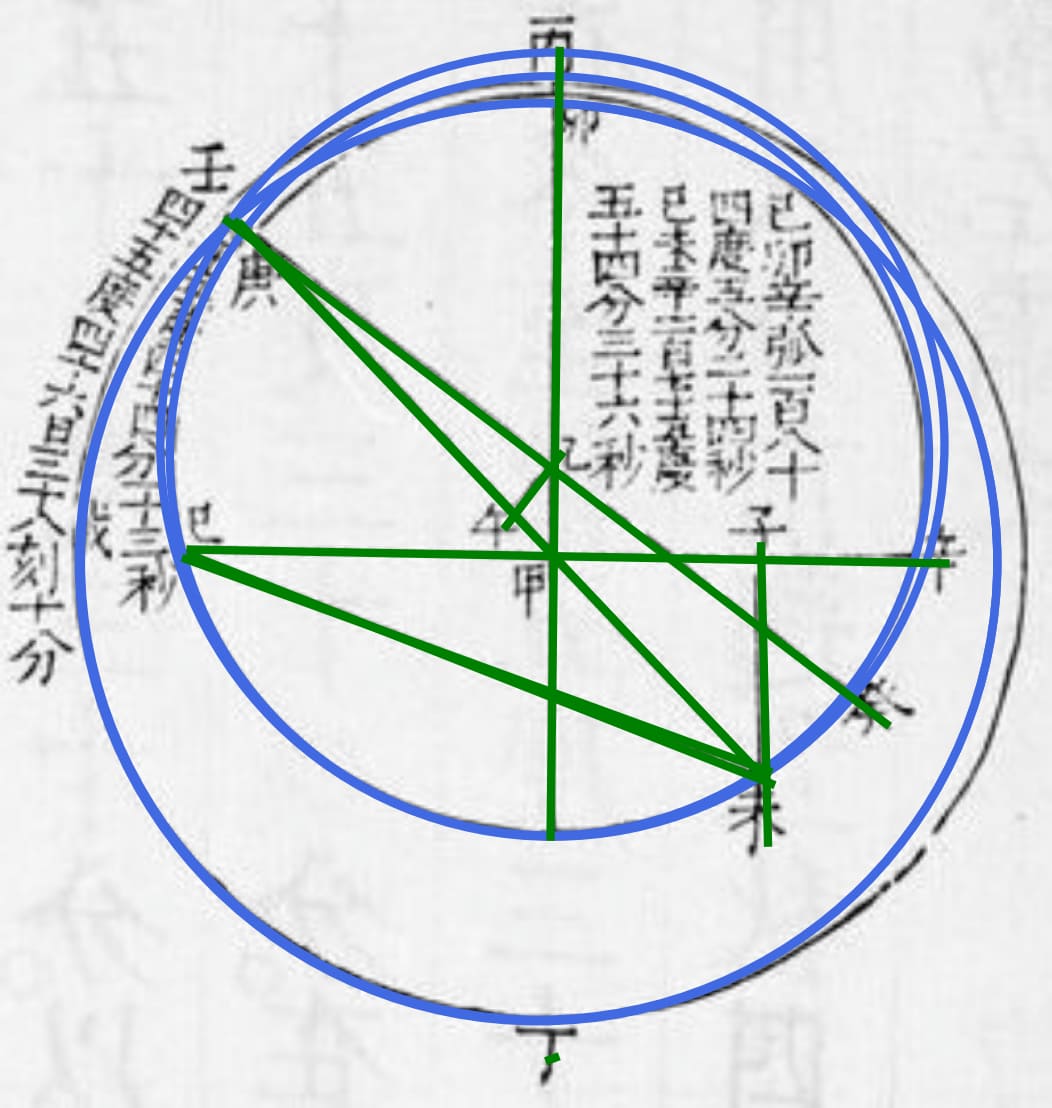} &
        \includegraphics[width=0.15\linewidth]{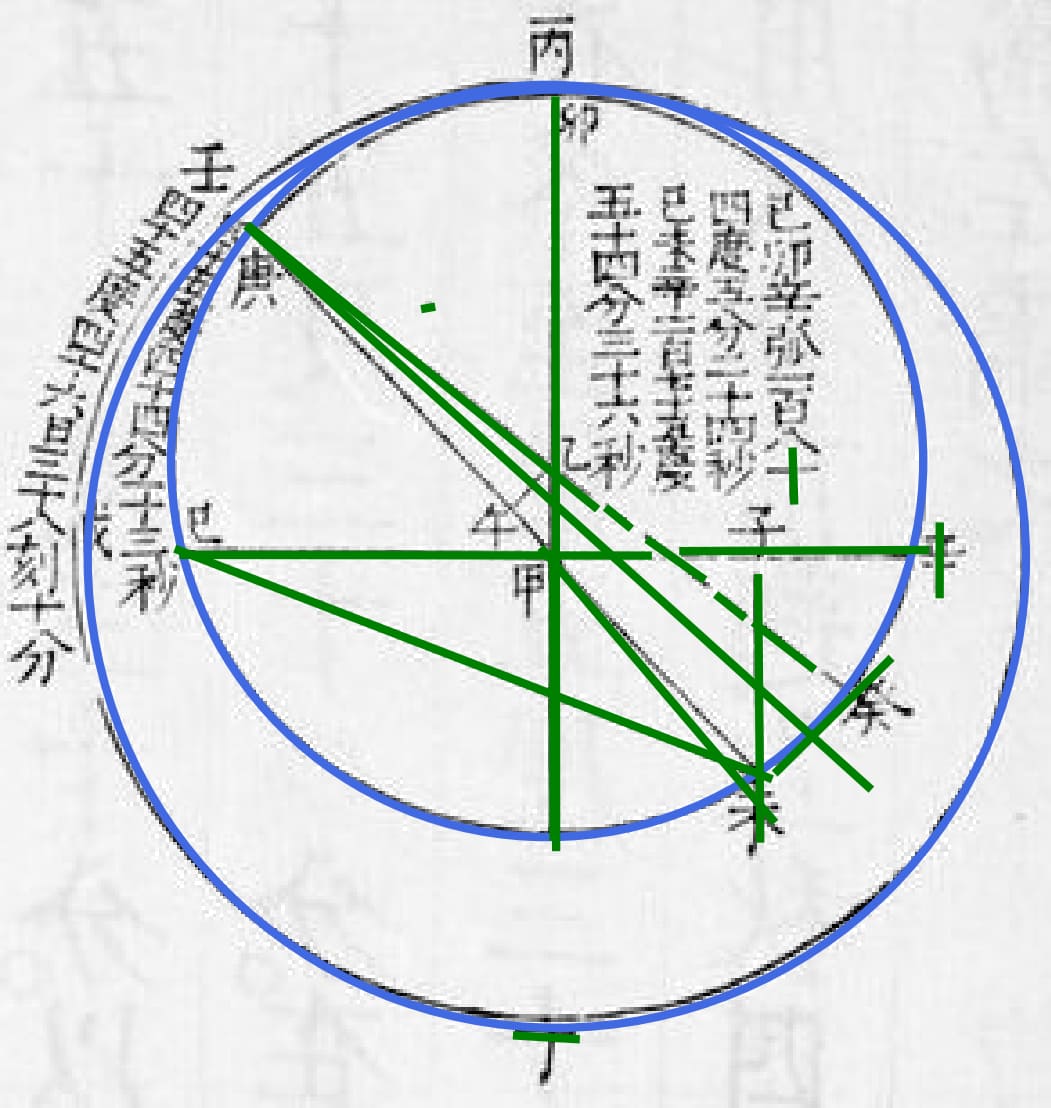} \\
        
        (a) Input & (b) GT & (c) Ours & (d) w/o QS & (e) w/o PR & (f) LETR  \\

          & & & & &   \\
    \end{tabular}
    \caption{Qualitative results on a random sample of $7$ diagrams from our dataset. The term PR denotes Primitive Refinement, while QS denotes Query Selection.}
    \label{fig:qual_results}
\end{figure}

\begin{figure}[t]
    \centering
    \includegraphics[width=\textwidth]{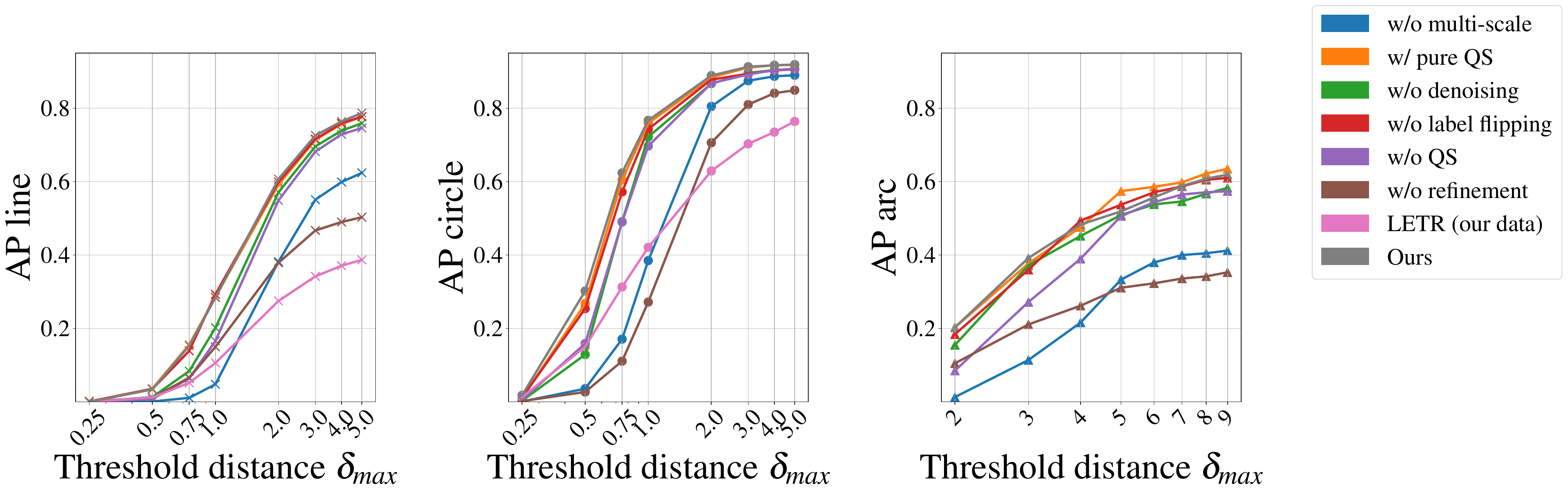}
    \caption{Average Precision for different distance thresholds. 
    }
    \label{fig:ap_plot}
\end{figure}

We report our main quantitative results in Table~\ref{tab:results}, using AP with threshold $\delta_{\text{max}}=4$ as explained in Section~\ref{sec:dataset}. 
We compare models for varying $\delta_{\text{max}}$ thresholds in Figure~\ref{fig:ap_plot}, which enables a finer analysis of the precision of each model. 
Figure~\ref{fig:qual_results} illustrates qualitative comparisons on $7$ randomly selected diagrams. 

\paragraph{Baselines}
We compare our model to the LETR~\cite{xu2021line} model, the closest model to our work, which expands DETR to perform line detection through a fine-to-coarse training strategy. We report both the performance of the model released by the authors, and the performance a LETR model re-trained with our dataset. We used pretrained DETR weights which, as emphasized in the original paper, we found necessary, and we train for 500 epochs in the coarse stage and for 200 epochs in the fine stage, which made training last approximately $150$ hours, significantly  longer than our model's training time. 
We also adapt LETR to perform circle detection by modifying the prediction head to output circle parameters, while keeping the same coarse-to-fine training strategy. Interestingly, we failed to adapt the LETR architecture for arcs in a similar way. Note that we could not compare to PPI-net~\cite{wang2023ppi}, which can predict circles, line segments and arcs, because the code is not available, but this architecture is based on DETR, required 72h training on 4 GPUs, and was only demonstrated on small black and white images of CAD designs~\cite{seff2020sketchgraphs}.

Unsurprisingly, Table~\ref{tab:results} shows that our synthetic dataset is more adapted for astronomical diagrams than the original LETR data. More interestingly, the boost in performance when using our network compared to the LETR approach is very significant, increasing the line AP from 0.371 to 0.764 and the circle AP from 0.754 to 0.917.

\clearpage

\paragraph{Ablation study} 
In our ablation study, we investigate the importance of several key components of our model:  (i) primitive refinement, which we ablate by directly predicting parameters with an encoder-only model; (ii) multi-scale feature maps, which we ablate by only using the last feature map of the backbone; (iii) contrastive denoising, which we ablate both completely (w/o denoising) or by only removing the label-flipping noise  (w/o label flipping); (iv) our primitive query initialization (mixed query selection~\cite{zhang2022dino}), which we also ablate in two ways, by either initializing both content and positional queries with the encoder (w/ pure query selection~\cite{zhu2020deformable}), or removing completely the query selection and directly learning the queries as parameters (w/o query selection). %

Table~\ref{tab:results} shows that using multi-scale features is critical to the success of our approach. This is clearly shown in the qualitative results in Figure~\ref{fig:qual_results}, with clearly inaccurate or some missing predictions.
Contrastive denoising and query selection also both have significant impacts on our result, which Figure~\ref{fig:ap_plot} enables to better analyze: they improve the precision of the predictions, which can be seen by looking at the AP using low thresholds $\delta$. Figure~\ref{fig:qual_results} shows that in addition to lower accuracy, not using query selection also translates in the qualitative results by the presence of duplicate or missing primitives. On the contrary, using label flipping or initializing the complete queries from the encoder does not impact the results significantly.

\begin{figure}[t]
    \centering
    \begin{subfigure}[t]{0.32\textwidth}
        \centering
        \includegraphics[width=0.48\textwidth]{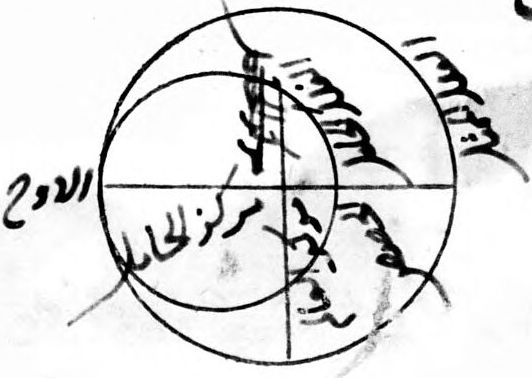}
        \includegraphics[width=0.48\textwidth]{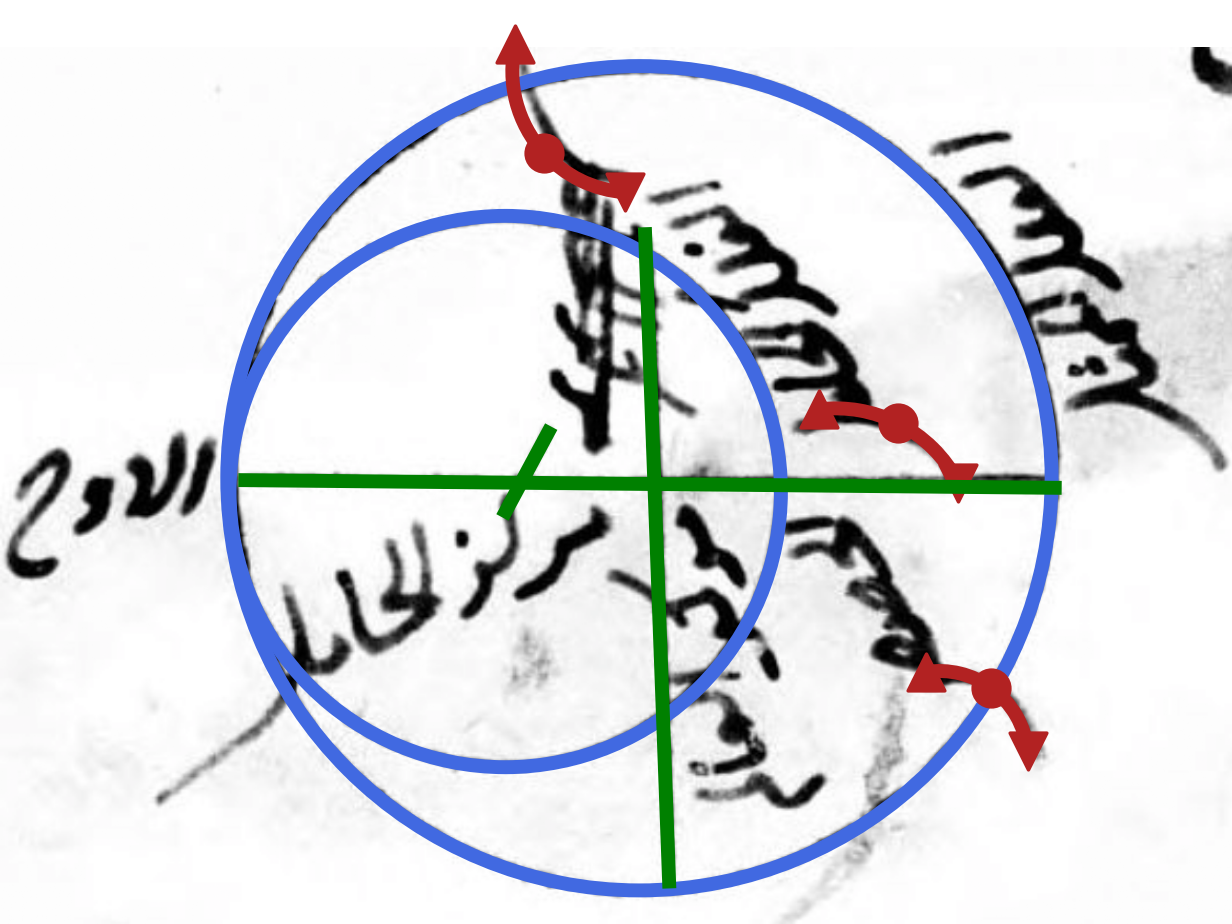}
        \includegraphics[width=0.48\textwidth, height = 1.8cm]{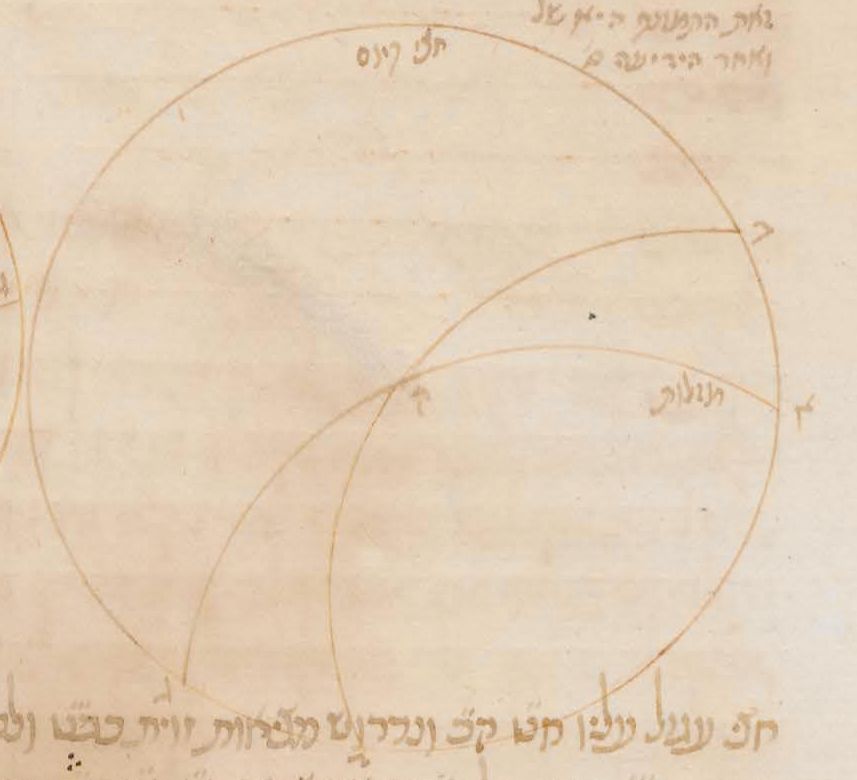}
        \includegraphics[width=0.48\textwidth, height = 1.8cm]{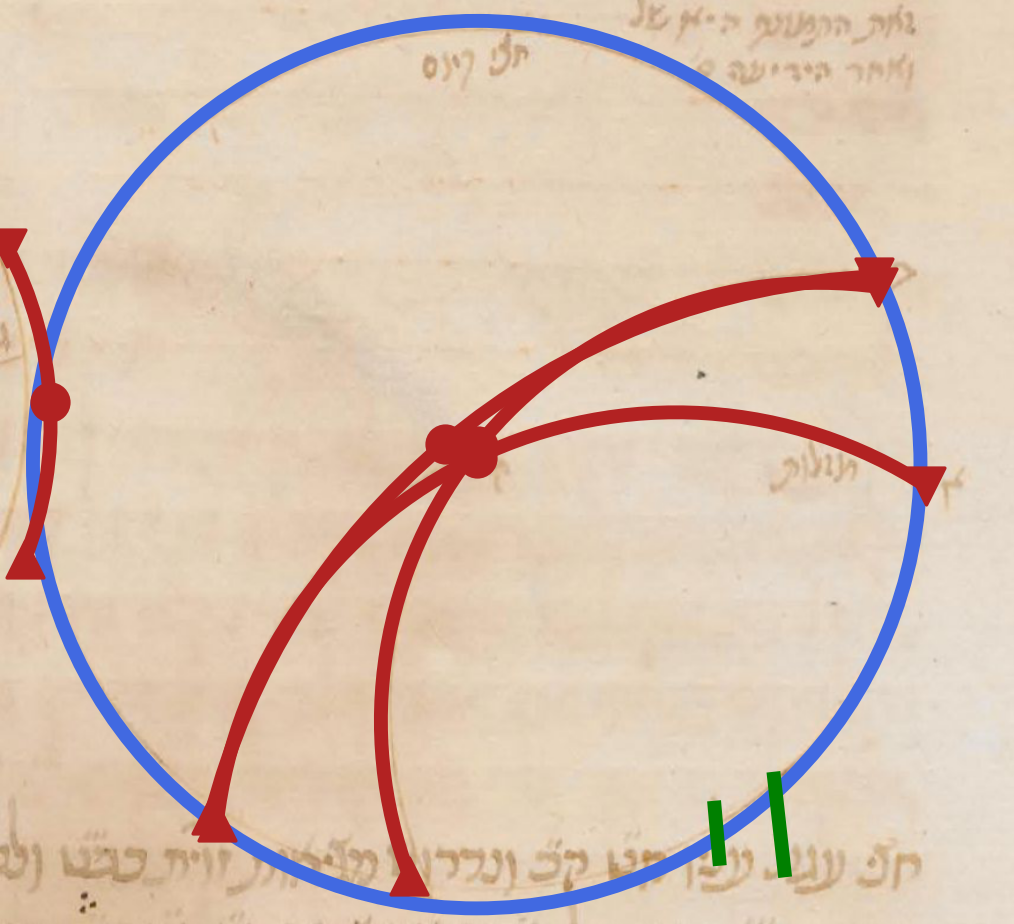}

        \caption{Letter detection}
        \label{fig:letter_fail}
    \end{subfigure}
    \begin{subfigure}[t]{0.325\textwidth}
        \centering
        \includegraphics[width=0.48\textwidth]{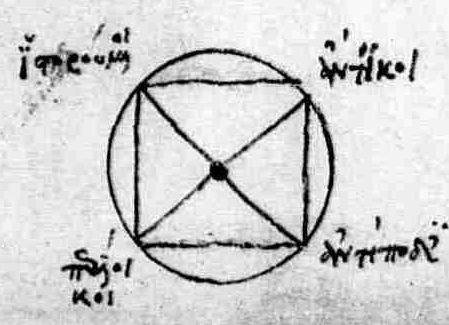}
        \includegraphics[width=0.48\textwidth]{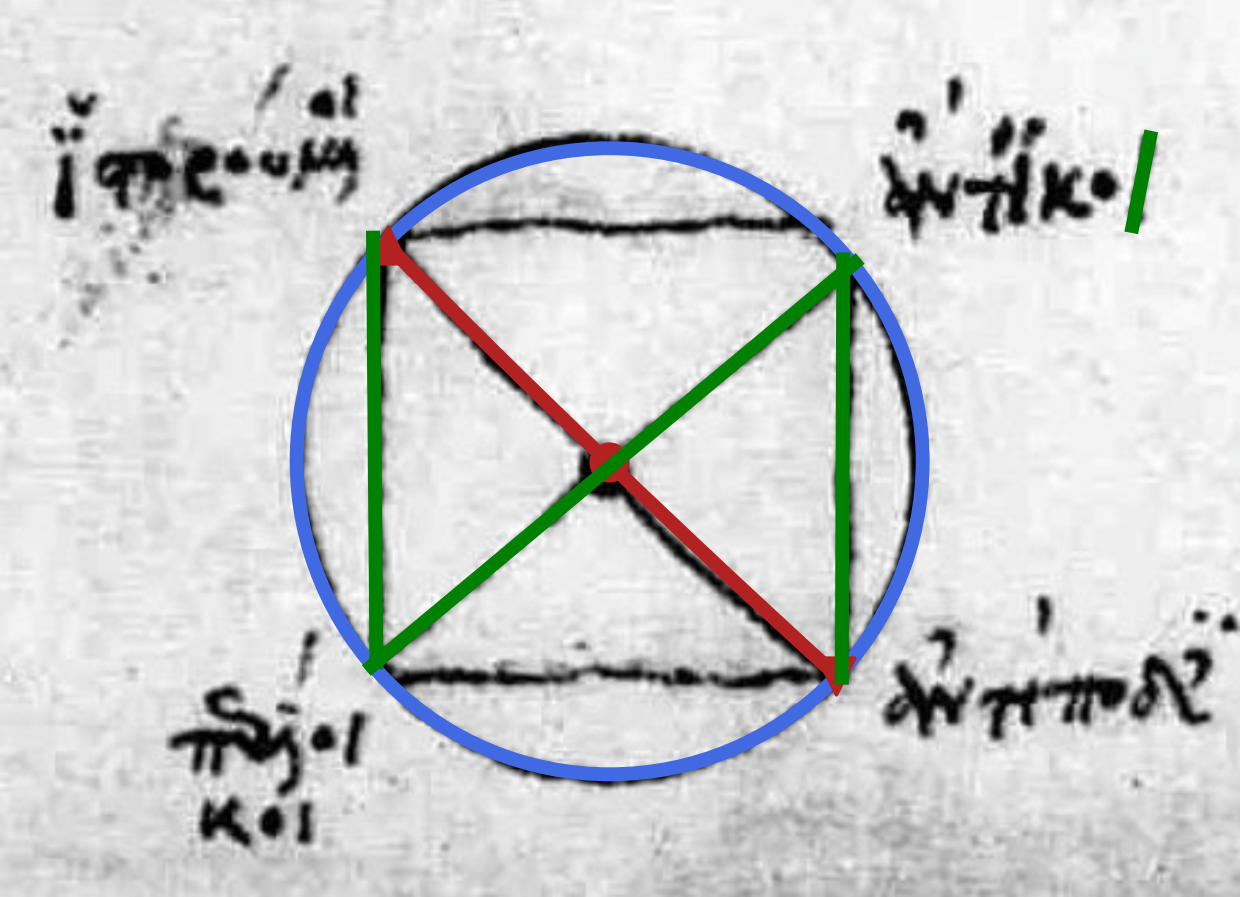}
        \includegraphics[width=0.475\textwidth, height = 1.8cm]{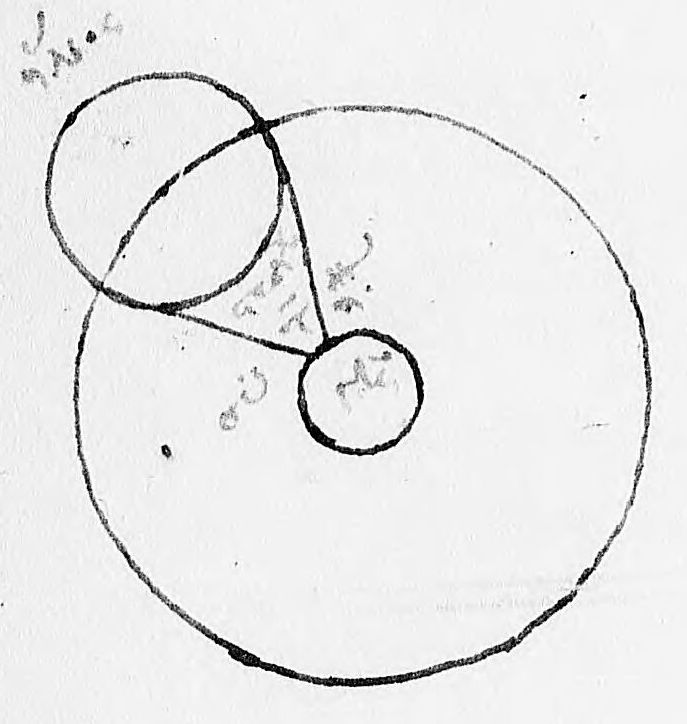}
        \includegraphics[width=0.475\textwidth, height = 1.8cm]{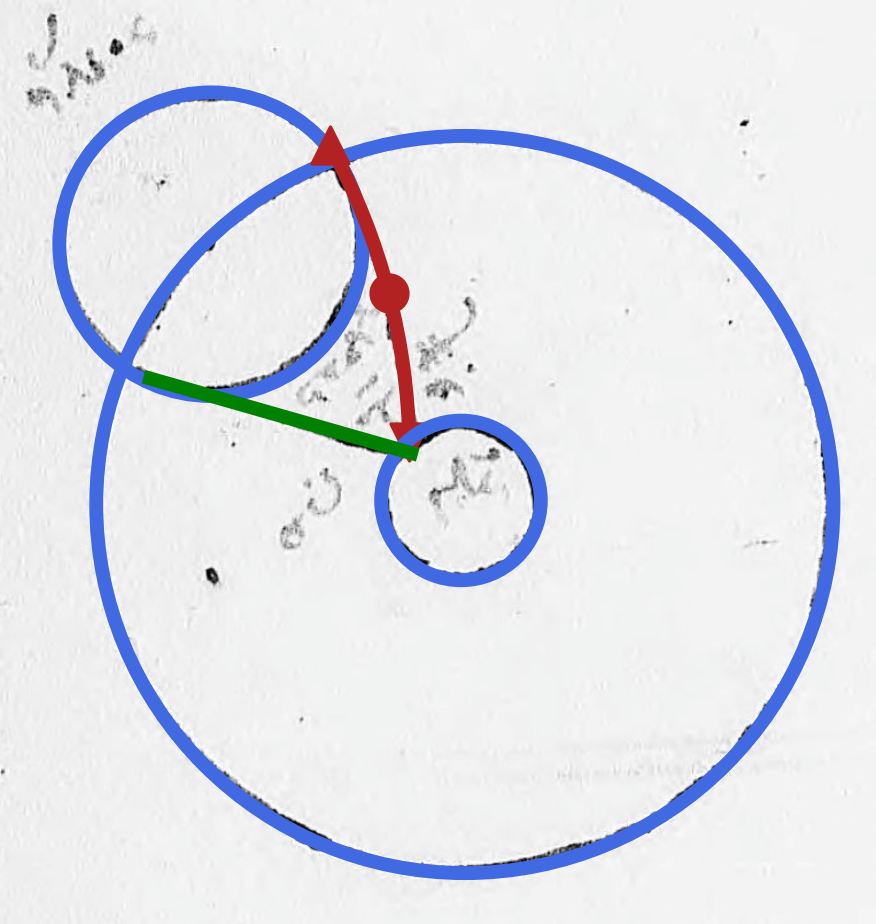}
        \caption{Curved lines}
        \label{fig:curve_fail}
    \end{subfigure}
    \hfill
    \begin{subfigure}[t]{0.325\textwidth}
        \centering
        \includegraphics[width=0.48\textwidth, height=1.4cm]{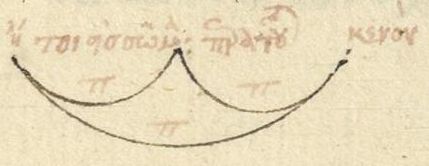}
        \includegraphics[width=0.48\textwidth, height=1.4cm]{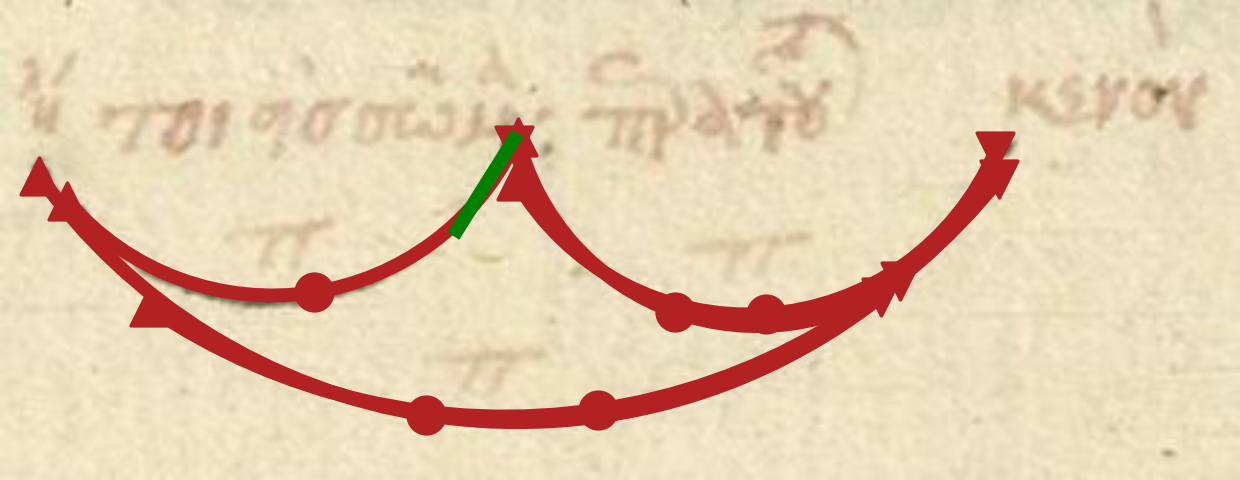}
        \includegraphics[width=0.475\textwidth, height = 1.8cm]{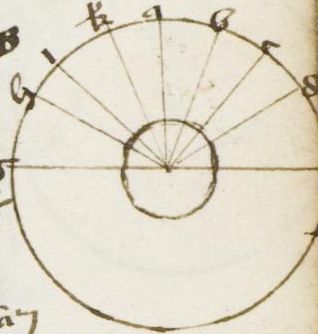}
        \includegraphics[width=0.475\textwidth, height = 1.8cm]{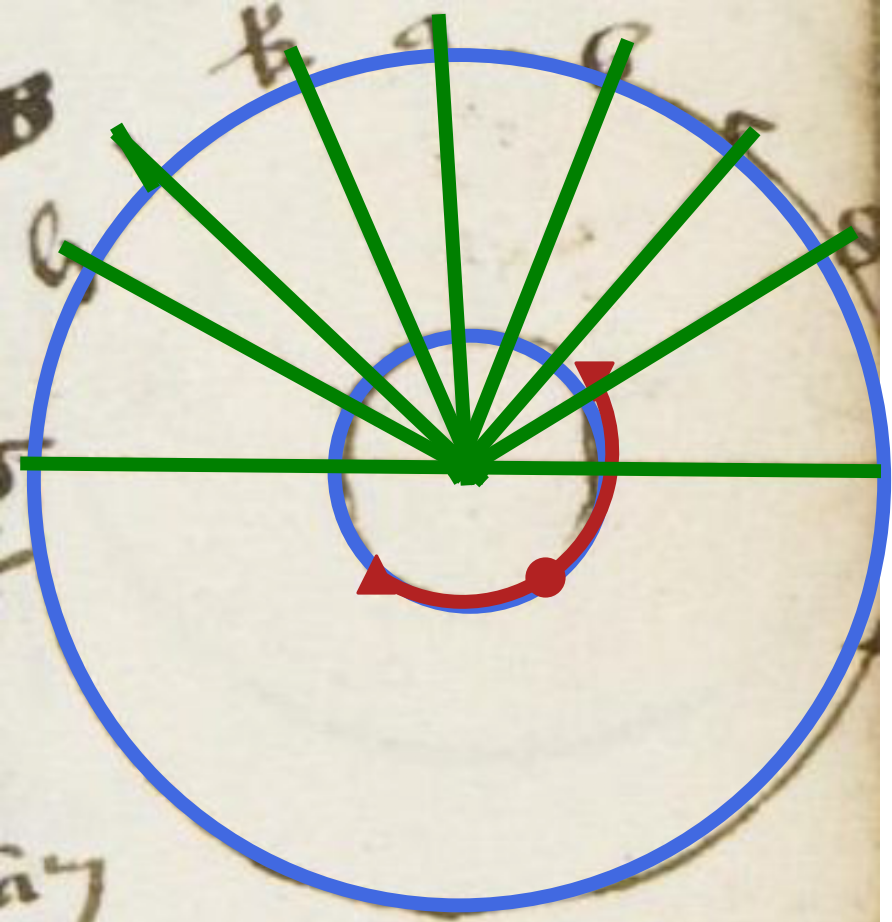}
        \caption{Duplicate arcs}
        \label{fig:duplicate_fail}
    \end{subfigure}
    \caption{Failure cases of our model.}
    \label{fig:failure_cases}
\end{figure}

\paragraph{Limitations}
Figure~\ref{fig:failure_cases} shows typical failure cases of our approach. We identified three main failure modes. First, our model is prone to detecting long curves or lines in some letters as primitives as arcs or line (Figure~\ref{fig:letter_fail}). Second, in freehand or less accurate drawings, our model fails to detect irregular lines or predicts low curvature arcs instead (Figure~\ref{fig:curve_fail}). Third, our model achieves lower performance on arcs than on other primitives, which is often showcased by the fact that it splits a single ground truth arc into several arcs or both a line and an arc, or overlays arcs with parts of irregular circles (Figure~\ref{fig:duplicate_fail}). While we believe that these failure cases could be mitigated by additional engineering efforts in the design of synthetic training data, or by fine-tuning our method on some real data, for example in an active learning setup, they are also related to the intrinsic difficulty of the data.

\begin{figure}
\begin{minipage}[t]{0.4\textwidth}
    \normalsize
    \vspace{-17mm} %
\paragraph{Qualitative results on S-VED}
To evaluate the generality of our model, we test it on samples from the S-VED dataset~\cite{buttner2022cor} shown in Figure~\ref{fig:s-ved}. While results are generally satisfactory, our synthetic data could be better adapted to paper prints, where shapes can be well seen in transparency, leading to false positives.
    \end{minipage}
    \hfill
    \begin{minipage}[t]{0.58\textwidth}
    \centering
    \begin{tabular}{cccc}
        \includegraphics[width=0.23\linewidth, height=0.22\linewidth]{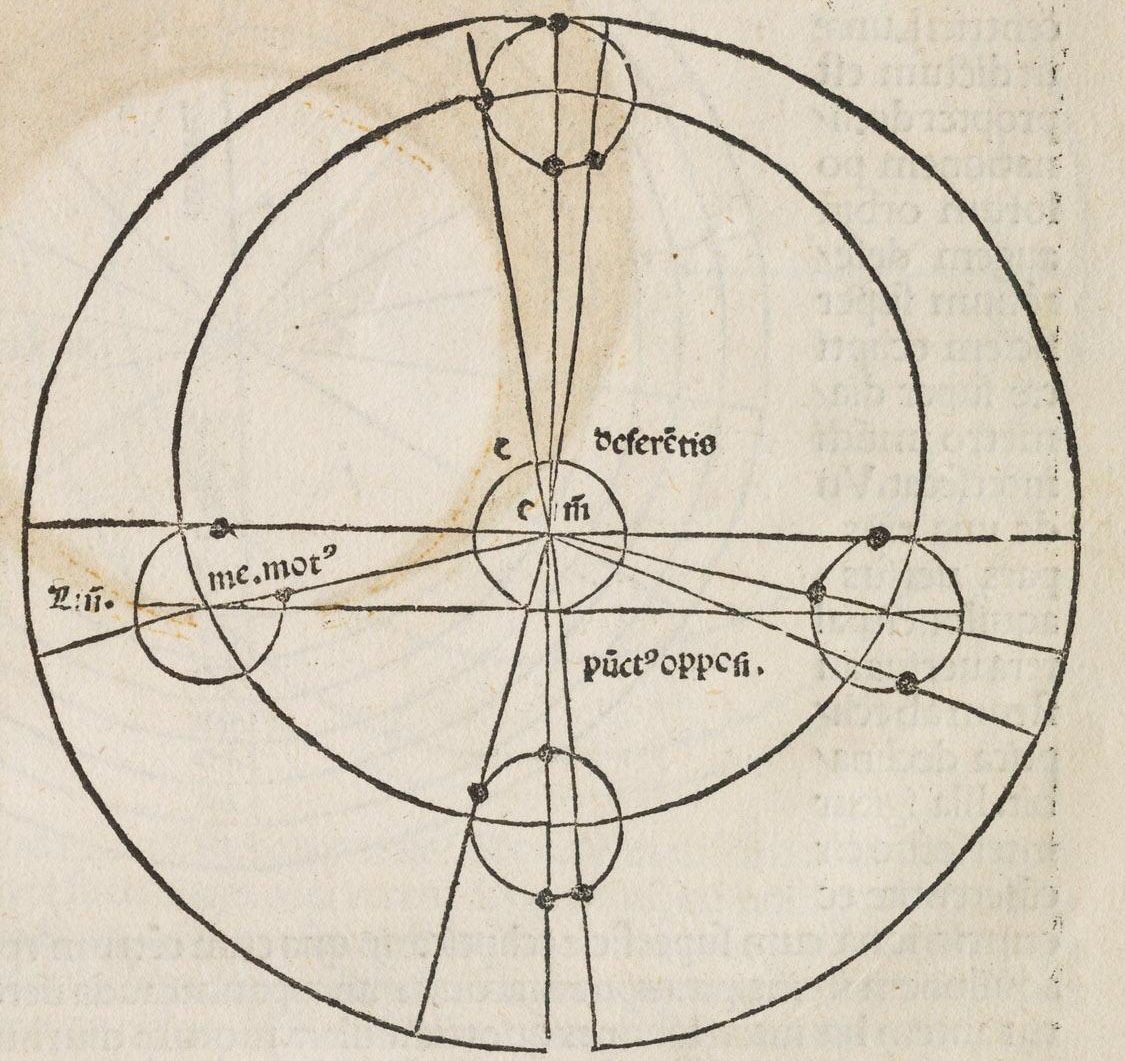} &
        \includegraphics[width=0.23\linewidth, height=0.22\linewidth]{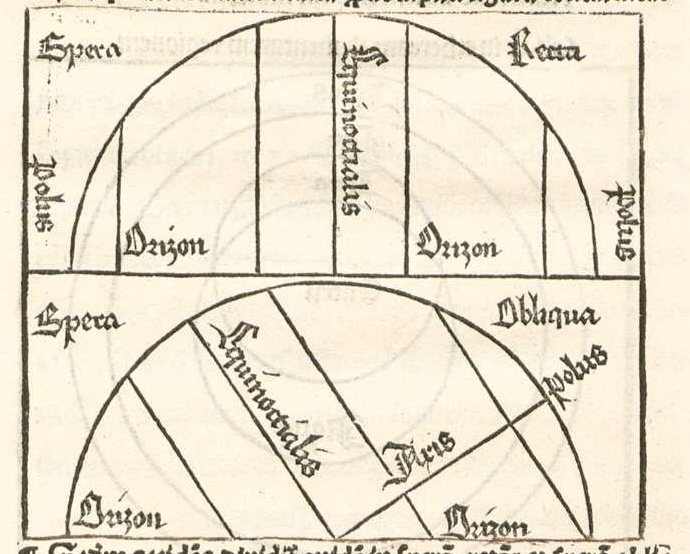} &
        \includegraphics[width=0.23\linewidth, height=0.22\linewidth]{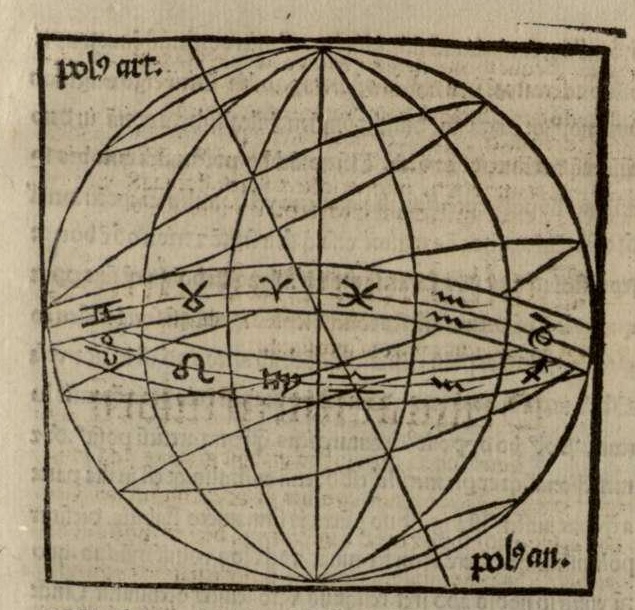}  &
        \includegraphics[width=0.23\linewidth, height=0.22\linewidth]{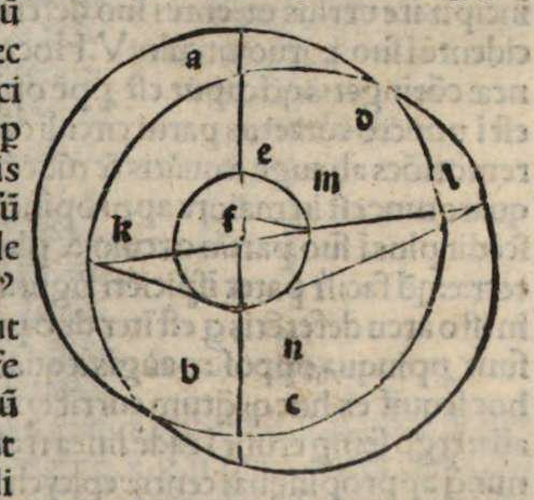}  
             \\
        \includegraphics[width=0.23\linewidth, height=0.22\linewidth]{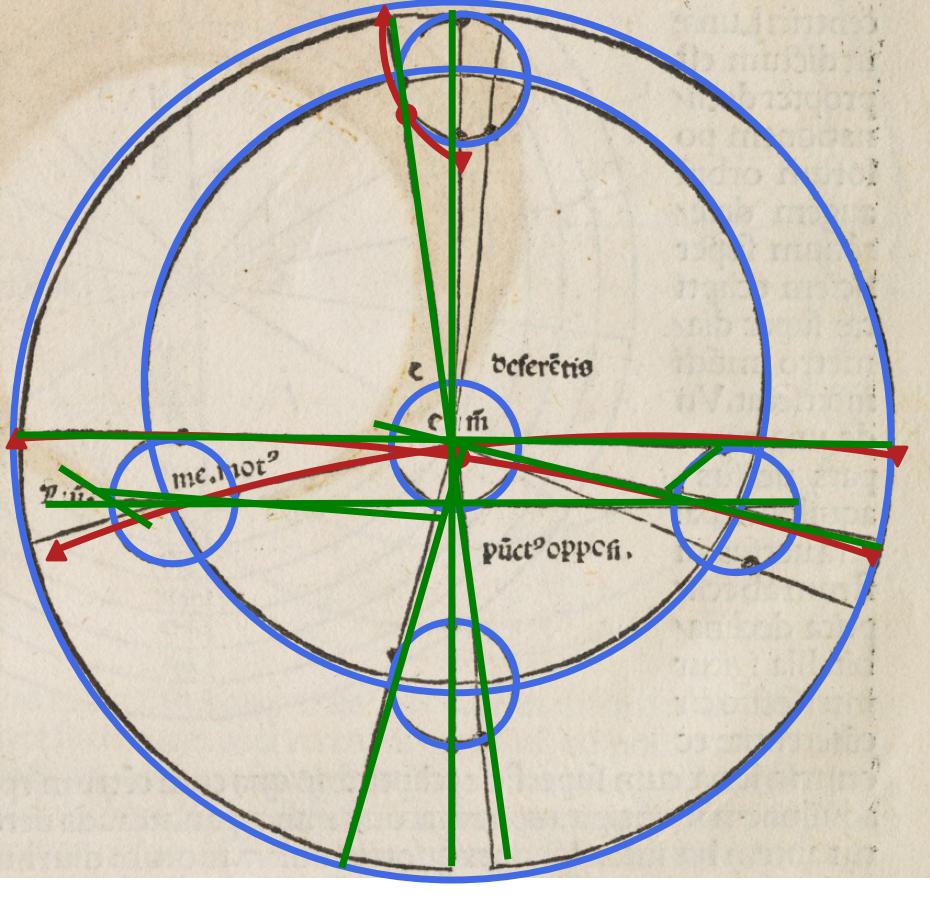} &
        \includegraphics[width=0.23\linewidth, height=0.22\linewidth]{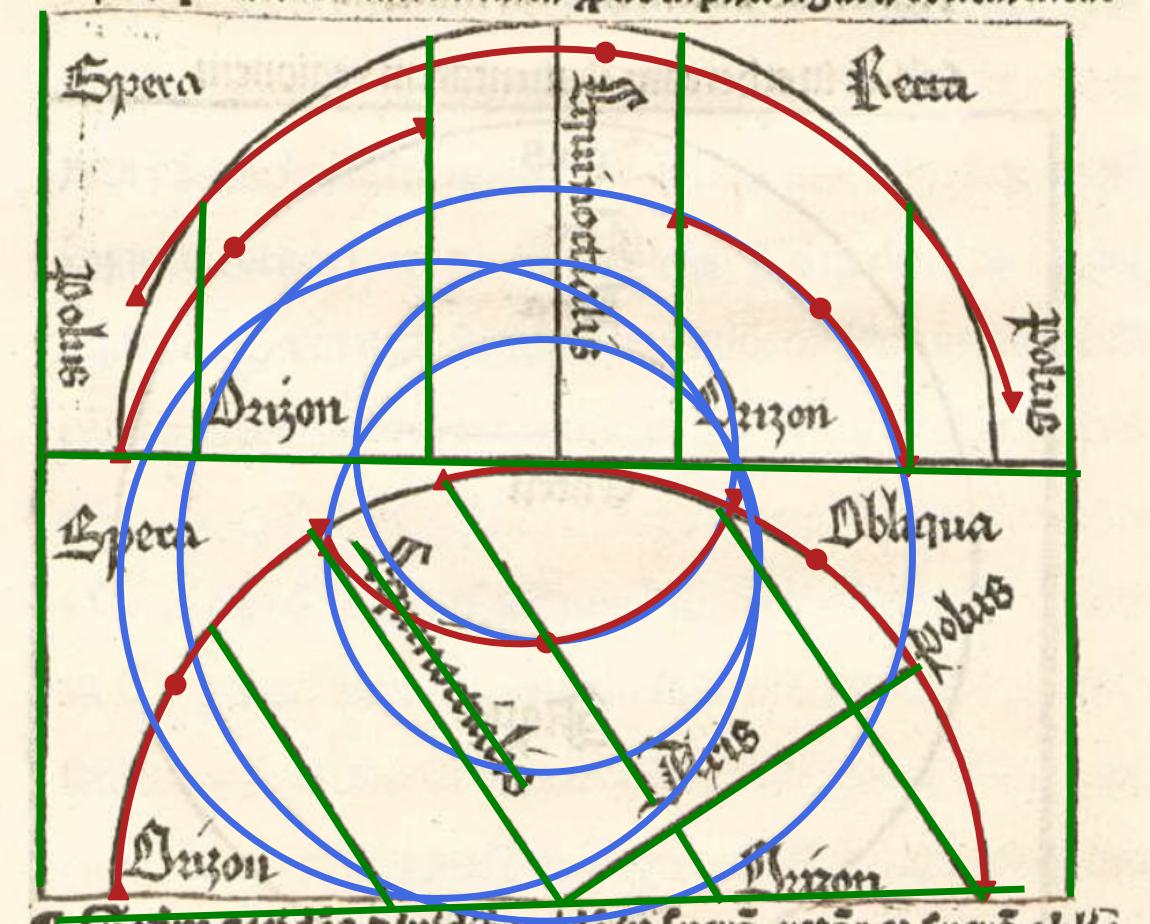} &
        \includegraphics[width=0.23\linewidth, height=0.22\linewidth]{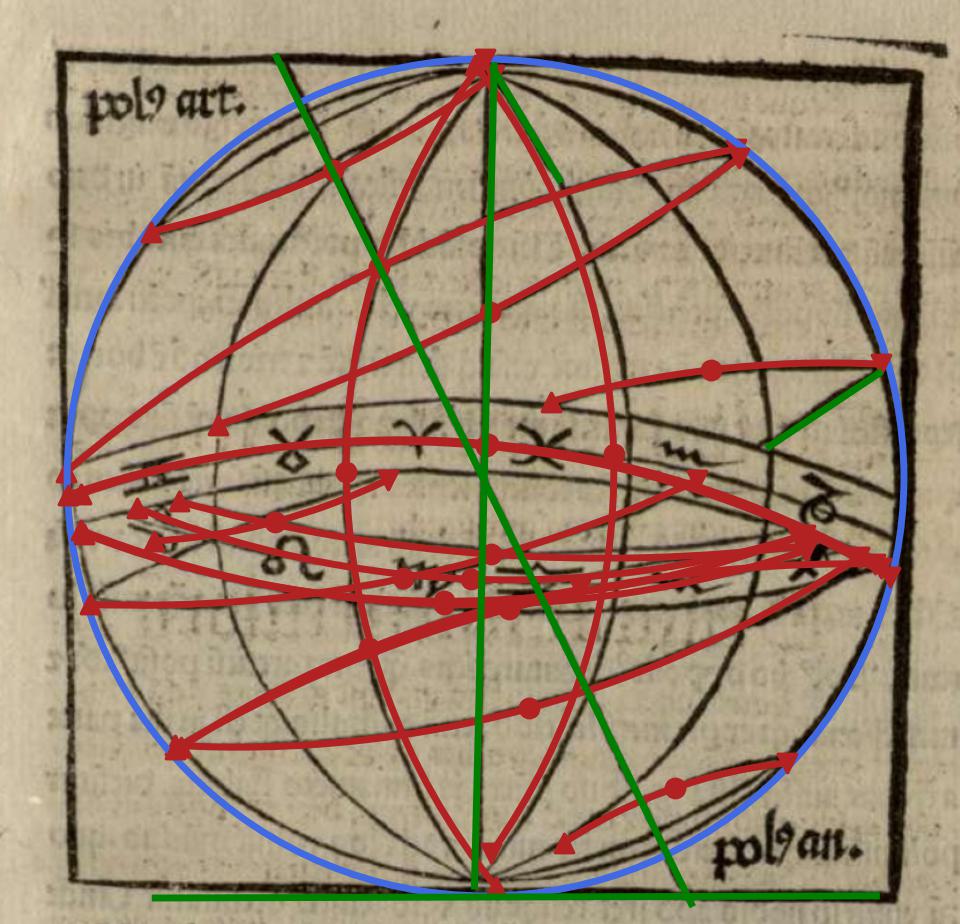} &
    \includegraphics[width=0.23\linewidth, height=0.22\linewidth]{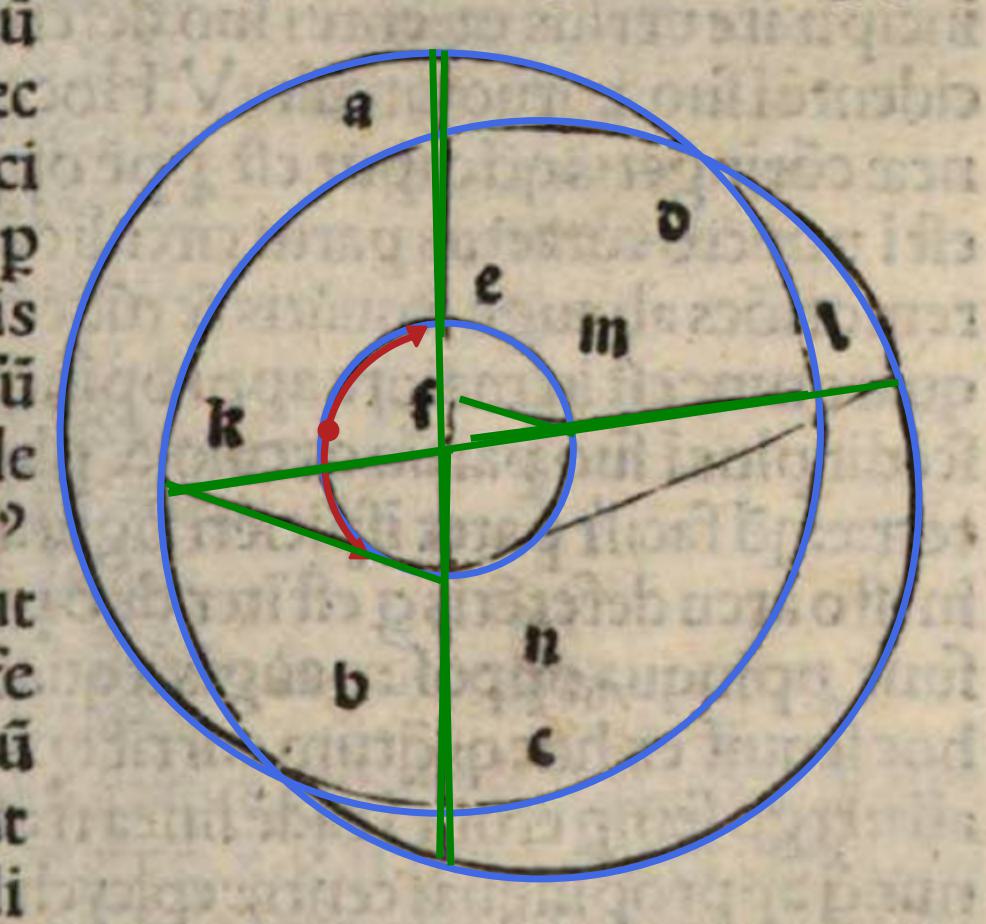}  

    \end{tabular}
    \caption{{\bf Qualitative results on printed S-VED~\cite{buttner2022cor} diagrams.} }
    \label{fig:s-ved}
    \end{minipage}%
\end{figure}

\section{Conclusion}

We presented a diverse and challenging dataset of 303 historical diagrams annotated with geometric primitives and a novel transformer-based approach to diagram vectorization. %
Our model refines the primitive parameters by relying on a novel parametrization of decoder queries. It is faster to train and more accurate than baselines, and it can naturally be extended to support other types of primitives. We show that this model can be trained solely on synthetic data, and generalize to real diagrams.
We believe that our method significantly improves over existing vectorization approaches, and our dataset opens up a novel and exciting application for History of Science. %

\section{Acknowledgments}
\small
This work was funded by ANR (project EIDA ANR-22-CE38-0014). The work of S. Trigg is supported by the European Research Council (ERC project NORIA, grant 724175). M. Aubry and S. Kalleli are supported by ERC project DISCOVER funded by the European Union’s Horizon Europe Research and Innovation program under grant agreement No. 101076028. We thank Ji Chen, Samuel Guessner, Divna Manolova, and Jade Norindr  for their help in collecting and annotating the dataset, %
and Sonat Baltaçi, Raphaël Benna, Yannis Siglidis,  Elliot Vincent, and  Malamatenia Vlachou for feedback and fruitful discussions. 

\bibliographystyle{splncs04}
\bibliography{biblio}

\end{document}